%% file: arxiv.tex
\documentclass[10pt,twocolumn,letterpaper]{article}

\usepackage{iccv}
\usepackage{times}
\usepackage{epsfig}
\usepackage{graphicx}
\usepackage{amsmath}
\usepackage{amssymb}
\usepackage{listings}
\usepackage{siunitx}
\usepackage{multirow}
\usepackage[table,xcdraw]{xcolor}
\usepackage{tabularx}
\usepackage{textcomp}
\usepackage{authblk}

\usepackage[pagebackref=true,breaklinks=true,letterpaper=true,colorlinks,bookmarks=false]{hyperref}

\iccvfinalcopy 

\newcommand{\PAR}[1]{\vskip3pt \noindent{\bf #1~}}

\newcommand{\kapture}[0]{kapture}
\newcommand{\Kapture}[0]{Kapture}

\newcolumntype{Y}{>{\centering\arraybackslash}X}

\begin{document}

\title{Robust Image Retrieval-based Visual Localization using Kapture}

\author{Martin Humenberger} 
\author{Yohann Cabon}
\author{Nicolas Guerin}
\author{Julien Morat}
\author{Vincent Leroy}
\author{J\'er\^ome Revaud}
\author{Philippe Rerole}
\author{No\'e Pion}
\author{Cesar de Souza}
\author{Gabriela Csurka}
\affil{NAVER LABS Europe, France \\
\url{https://europe.naverlabs.com} \\
{\tt\small firstname.lastname@naverlabs.com}
}

\maketitle


\begin{abstract}
Visual localization tackles the challenge of estimating the camera pose from images by using correspondence analysis between query images and a map. This task is computation and data intensive which poses challenges on thorough evaluation of methods on various datasets. However, in order to further advance in the field, we claim that robust visual localization algorithms should be evaluated on multiple datasets covering a broad domain variety. To facilitate this, we introduce \kapture, a new, flexible, unified data format and toolbox for visual localization and structure-from-motion (SFM). It enables easy usage of different datasets as well as efficient and reusable data processing. To demonstrate this, we present a versatile pipeline for visual localization that facilitates the use of different local and global features, 3D data (\eg depth maps), non-vision sensor data (\eg IMU, GPS, WiFi), and various processing algorithms. Using multiple configurations of the pipeline, we show the great versatility of \kapture~in our experiments. Furthermore, we evaluate our methods on eight public datasets where they rank top on all and first on many of them. To foster future research, we release code, models, and all datasets used in this paper in the \kapture~format open source under a permissive BSD license. \\ \url{github.com/naver/kapture} \\ \url{github.com/naver/kapture-localization}
\end{abstract}

\section{Introduction}

\begin{figure}[t]
    \centering
    \includegraphics[width=\linewidth]{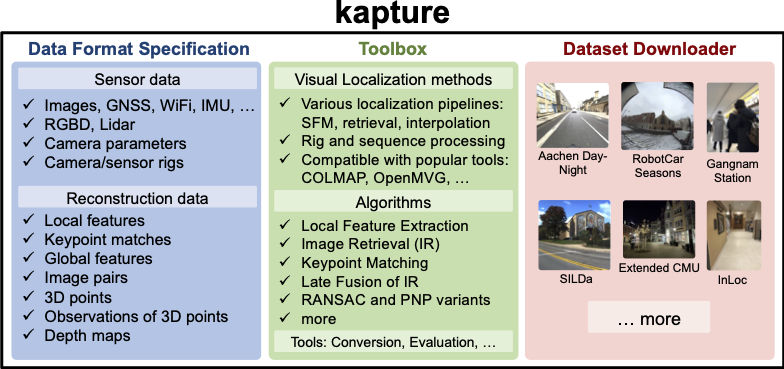}
    \caption{\Kapture~consists of a detailed sensor and reconstruction data specification, I/O and conversion tools, visual localization methods, and a dataset downloader that directly provides datasets in the \kapture~format.}
    \label{fig:banner}
    \vspace{-4mm}
\end{figure}

Visual localization is an important component of many location-based systems such as self-driving cars~\cite{HengICRA19ProjectAutoVisionLocalization3DAutonomousVehicle,da4adECCV20}, autonomous robots~\cite{LimIJRR15RealTimeMonocularImageBased6DoFLocalization}, or augmented, mixed, and virtual reality~\cite{ArthISMAR09WideAreaLocalizationMobilePhones,MiddelbergECCV14Scalable6DOFLocalization,LynenRSSC15GetOutVisualInertialLocalization}. 
The goal is to estimate the accurate position and orientation of a camera from captured images.
In more detail, correspondences between a representation of the environment (map) and a query image are utilized to estimate the camera pose in 6 degrees of freedom (DOF).
Environmental changes caused by time of day or season of the year, but also structural changes on house facades or store fronts, large view point changes between mapping and query images, as well as people, cars, or other dynamic objects that occlude parts of the scene, pose critical challenges on visual localization methods.

\PAR{Methods.}
Visual localization is an active research field and many approaches have been proposed.
Structure-based methods~\cite{Sattler2011ICCV,MoulinICCV13GlobalFusionRelativeMotions,SattlerPAMI17EfficientPrioritizedMatching,LiuICCV17Efficient2D3DMatchingLocalization,Taira2019TPAMI,SchonbergerCVPR16StructureFromMotionRevisited} use local features to establish correspondences between 2D query images and 3D reconstructions. 
These correspondences are then used to compute the camera pose using perspective-n-point (PNP) solvers~\cite{KneipCVPR11ANovelParametrizationAbsoluteCamPose} within a RANSAC loop~\cite{Fischler81CACM,Chum08PAMI,Lebeda2012BMVC,makingaffinecorresworkECCV20}. 
To reduce the search range in large 3D reconstructions, image retrieval methods~\cite{Pion3DV20BenchmarkingImageRetrievalVisualLocalization,SarlinCVPR19FromCoarsetoFineHierarchicalLocalization} can be used to first retrieve most relevant images from the SFM model. 
Second, local correspondences are established in the area defined by those images.

Scene point regression methods~\cite{ShottonCVPR13SceneCoordinateRegression,BrachmannCVPR18LearningLessIsMore6DLocalization,Brachmann2019ICCVa,hieraSceneCoordCVPR20,brachmann2020visual} establish the 2D-3D correspondences using a deep neural network (DNN) and absolute pose regression methods~\cite{KendallICCV15PoseNetCameraRelocalization,LaskarICCVWS17CameraRelocalizationRelativePosesCNN,BalntasECCV18RelocNetMetricLearningRelocalisation,SattlerCVPR19UnderstandingLimitationsPoseRegression,kendall2017geometric,6drelocambiguousECCV20} directly estimate the camera pose with a DNN.
Another strategy is to estimate the pose of an image by directly aligning deep images features with a reference 3D model~\cite{sarlin2021feature}.

Image retrieval can also be used for visual localization when no 3D map is available. 
The camera pose of a query image can be computed from the poses of top retrieved database images by interpolating retrieved image poses~\cite{ToriiICCVWS11VisualLocalizationByLinearCombination,ZamirECCV10AccurateImageLocalization,Torii2019TPAMI,SattlerCVPR19UnderstandingLimitationsPoseRegression}, estimating the relative pose between query and retrieved images~\cite{ZhouICRA20ToLearnLocalizationFromEssentialMatrices,ZhangS3DPVT06ImageBasedLocUrbanEnvironments}, estimating absolute pose from 2D-2D matches~\cite{Zheng2015ICCV}, via relative pose regression~\cite{BalntasECCV18RelocNetMetricLearningRelocalisation,DingICCV19CamNetRetrievalForReLocalization} or by building local 3D models on the fly~\cite{Torii2019TPAMI,Pion3DV20BenchmarkingImageRetrievalVisualLocalization}.

Furthermore, objects~\cite{WeinzaepfelCVPR19VisualLocObjectsOfInterestDenseMatchRegression,salas2013slam++,cohen2016indoor,ardeshir2014gis} or semantic information~\cite{fan2020visual} can also be used for visual localization.

\PAR{Processing.}
For many visual localization algorithms, large neural networks have to be trained and executed, and large-scale SFM reconstructions have to be computed. 
Considering the fact that even medium sized datasets for visual localization often consist of more than 1000 images, this typically makes it a computationally intensive task.
Furthermore, since the structure of the map is often domain, application, and method specific, it needs to be separately constructed for each dataset and sometimes even for each experiment.
Given the algorithmic challenges mentioned above, we claim that visual localization or SFM algorithms should be evaluated on multiple datasets covering broad in-domain and/or cross-domain variety.
Even if these datasets exist (see Section~\ref{sec:eval:datasets}), evaluation comes with the burden that datasets are often provided in different formats including different coordinate systems and camera parameter representations.

To facilitate data processing for visual localization, SFM, sensor fusion, and related fields such as visual simultaneous localization and mapping (VSLAM), in this paper we first introduce a new toolbox called \kapture~(see Figure~\ref{fig:banner}).
We then use it to implement a versatile processing pipeline which we evaluate on eight popular datasets with various parameter settings and pipeline configurations.

Our pipeline is inspired by structure-based methods that use image retrieval to select the image pairs used for local feature matching.
This approach has the following advantages that strengthen our choice: (i) it allows convenient integration of data-driven methods, most notably for image retrieval and local feature matching, (ii) while still providing the benefits of accurate pose computation using a geometric pipeline. 
Finally, (iii) as will be shown in Section~\ref{sec:eval:summary}, such methods are applicable on a variety of datasets and applications scenarios.

\PAR{Image features.}
As a consequence, most relevant to our work are data-driven local~\cite{NohICCV17LargeScaleAttentiveDeepLocalFeatures,AndersonCVPR18SuperpointInterestPoint,DusmanuCVPR19D2NetDeepLocalFeatures,RevaudNIPS19R2D2ReliableRepeatableDetectorsDescriptors,CsurkaX18FromHandcraftedToDeepLocalFeatures,aslfeatCVPR20,learningfeaturesfromcameraposeECCV20} and global~\cite{ArandjelovicCVPR16NetVLADPlaceRecognition,RadenovicPAMI19FineTuningCNNImRet,RevaudICCV19LearningwithAPTrainingImgRetrievalListwiseLoss} image features. 
Using those, recent advances in the field showed great results on tasks like image matching~\cite{RadenovicCVPR18RevisitingOxfordParisImRetBenchmarking,aggdeeplocalfeatECCV20} and visual localization~\cite{SarlinCVPR19FromCoarsetoFineHierarchicalLocalization,DusmanuCVPR19D2NetDeepLocalFeatures,RevaudX19R2D2ReliableRepeatableDetectorsDescriptors,imagetoregionECCV20,Pion3DV20BenchmarkingImageRetrievalVisualLocalization}.
Motivated by the good performance reported for the local feature R2D2~\cite{RevaudNIPS19R2D2ReliableRepeatableDetectorsDescriptors} on visual localization and the global feature APGeM~\cite{RevaudICCV19LearningwithAPTrainingImgRetrievalListwiseLoss} on landmark retrieval, in this paper, using \kapture~and our proposed pipeline, we further evaluate the performance of such methods on multiple visual localization datasets.

\PAR{Contributions.}
Summarizing, this paper makes the following contributions:
\textbf{First}, we propose a toolbox called \kapture~that consists of a unified data format as well as processing tools for visual localization and structure from motion.
\textbf{Second}, using \kapture, we propose a versatile localization pipeline encompassing several variants of image retrieval, SFM, and RGBD-based localization techniques relying on robust local and global image features.
\textbf{Third}, we propose extensive experimental validation of \kapture~as well as the pipeline and its variants on eight challenging visual localization datasets ranking top on all and first on many of them.
The \kapture~toolbox, including code and datasets used in this paper (in \kapture~format), is publicly available.

\section{Related Work}
\label{sec:related}

\PAR{Local features.}
Local features play an important role in visual localization as shown in recent benchmarks comparing local features for matching~\cite{BalntasCVPR17HPatchesBenchmark,JinIJCV19ImageMatchingAcrossWideBaselinesPaperToPractice}, SFM~\cite{Schoenberger2017CVPR}, and localization~\cite{LTVL,JinIJCV19ImageMatchingAcrossWideBaselinesPaperToPractice}. 
Early methods used handcrafted local feature extractors, notably the popular SIFT descriptor~\cite{LoweIJCV04DistinctiveImageFeaturesScaleInvariantKeypoints}. 
However, such keypoint extractors and descriptors have several limitations, including the fact that they are not necessary tailored to the challenges of the target task, in particular to changing conditions such as day-night and seasons of the year.
Therefore, several data-driven learned representations were proposed recently including learning local features with end-to-end deep architectures (see the evolution of local features in \cite{CsurkaX18FromHandcraftedToDeepLocalFeatures,Schoenberger2017CVPR}). 
In particular, Key.Net~\cite{BarrosoLagunaICCV19KeyNetKeypointDetHandcraftedAndCNN} has shown state-of-the art keypoint detection performance, SuperGlue~\cite{SarlinCVPR20SuperGlueLearningFeatureMatchingGNN} and AdaLAM~\cite{CavalliX20AdaLAMRevisitingHandcraftedOutlierDetection} improve feature matching, while the joint detector+descriptors Superpoint~\cite{AndersonCVPR18SuperpointInterestPoint},
D2-Net~\cite{DusmanuCVPR19D2NetDeepLocalFeatures}, R2D2~\cite{RevaudX19R2D2ReliableRepeatableDetectorsDescriptors}, and ASLFeat~\cite{aslfeatCVPR20} report state-of-the art performance on several local feature and visual localization benchmarks~\cite{BalntasCVPR17HPatchesBenchmark,JinIJCV19ImageMatchingAcrossWideBaselinesPaperToPractice,LTVL}.

In this paper we use R2D2, D2-Net, ASLFeat, and SIFT for local feature matching. 
We also propose a modification of the R2D2 network architecture that reduces its processing time by a factor three and preserves or even improves the localization accuracy.

\PAR{Global image representations.}
Another key element are global image representations used for image retrieval. 
Image retrieval-based localization methods~\cite{Torii2019TPAMI,Taira2019TPAMI,SarlinCVPR19FromCoarsetoFineHierarchicalLocalization,DusmanuCVPR19D2NetDeepLocalFeatures,Germain20193DV,ZhouICRA20ToLearnLocalizationFromEssentialMatrices,imagetoregionECCV20} mostly use the handcrafted DenseVLAD~\cite{ToriiCVPR15PlaceRecognitionByViewSynthesis}, which aggregates local RootSIFT~\cite{ArandjelovicCVPR12ThreeThingsEveryoneObjRet,LoweIJCV04DistinctiveImageFeaturesScaleInvariantKeypoints} descriptors extracted on a multi-scale, densely sampled grid into an intra-normalized~VLAD representation~\cite{JegouCVPR10AggregatingLocalDescriptors} or NetVLAD~\cite{ToriiCVPR15PlaceRecognitionByViewSynthesis}, which aggregates mid-level convolutional features extracted from the entire image into a compact single vector representation for efficient indexing in a similar way.
\cite{Pion3DV20BenchmarkingImageRetrievalVisualLocalization} compares these features with state-of-the art deep image representations APGeM~\cite{RevaudICCV19LearningwithAPTrainingImgRetrievalListwiseLoss} and DELG~\cite{CaoX20UnifyingDeepLocalGlobalFeatures} trained on landmark recognition tasks, where the aim is to retrieve images of the same building or place, independently of the pose or viewing conditions.
APGeM, similarly to \cite{RadenovicPAMI19FineTuningCNNImRet} uses a generalized-mean pooling layer (GeM) to aggregate CNN-based descriptors of several image regions at different scales but instead of contrastive loss, it directly optimizes the Average Precision (AP) approximated by histogram binning to make it differentiable. 
DELG uses a CNN that is trained to jointly extract local and global features. After a common backbone, the model is split into two parts (heads), one to detect relevant local feature and one which describes the global content of the image as a compact descriptor. 
The two networks are jointly trained in an end-to-end manner using the ArcFace~\cite{ArcFaceCVPR2019Deng} loss for the global descriptor. 
The method was originally designed for image search, where the local features enable geometric verification and re-ranking.

In this paper, we use DenseVLAD, NetVLAD, APGeM, and DELG for image retrieval.
We also propose a late fusion technique~\cite{csurka2012empirical} to combine multiple global image representations in order to outperform each representation individually.

\PAR{Datasets and processing.}
In order to evaluate visual localization methods under various conditions and scenarios, several datasets were introduced. 
Some relevant examples are Cambridge Landmarks~\cite{KendallICCV15PoseNetCameraRelocalization}, an outdoor dataset consisting of 6 landmarks of the city of Cambridge, UK and 7-scenes~\cite{ShottonCVPR13SceneCoordinateRegression}, which consists of seven small indoor areas captured using an RGBD camera.
Furthermore, \cite{SattlerCVPR18Benchmarking6DoFOutdoorLoc} provides an online benchmark which includes several datasets covering a variety of challenges for visual localization in different application scenarios such as autonomous driving and indoor as well as outdoor handheld camera localization (see Section~\ref{sec:eval:datasets} for more details).
More examples include TUM-LSI~\cite{walch17spatiallstms}, ETH3D~\cite{schoeps2017cvpr}, KITTI~\cite{kitti}, Middlebury~\cite{middlebury}, and 1DSfM~\cite{wilson_eccv2014_1dsfm}.

Next to various datasets, a rich set of tools for visual localization and SFM was released in the last two decades which greatly boosted the research and applications in these fields. 
While the variety of tools and datasets is important and provides many opportunities, the different data formats they come with impose operational challenges using them in practice.
For example, well established tools like Bundler~\cite{Bundler}, COLMAP~\cite{SchonbergerCVPR16StructureFromMotionRevisited}, OpenMVG~\cite{openMVG}, OpenSfM~\cite{OpenSfM}, VisualSFM~\cite{VisualSFM}, as well as the datasets mentioned above, all come with their own data format.
To better understand the differences, Table~\ref{tab:sfmdataformat} compares popular data formats.
We mark a data type as supported if it is inherently part of the format and not only a product of a processing step.
For example, COLMAP can produce depth maps from reconstructions but it is not possible to import depth data from an RGBD camera.
As can be seen, none of them fulfills all requirements of modern visual and sensor-based localization (see Section~\ref{sec:kapture} for details).
Notably, none of the existing data formats supports global features, 3D sensor data (\eg~LiDAR or RGBD), and integration of other sensors such as WiFi or IMU (inertial measurement unit).
Extending an existing format would be a valid option (and is often done in practice), however this would disentangle it from its original purpose which we think will not have great usability and impact.
Thus, inspired by existing formats and practical needs, we decided to design a new data format from scratch that combines all required features, is extendable and independent of existing tools but easily convertible into existing data formats.

The remainder of the paper is structured as follows.
First, in Section~\ref{sec:kapture}, we introduce and describe our new toolbox \kapture. 
Second, in Section~\ref{sec:method}, we describe our mapping and localization pipeline and its variants.
Finally, in Section~\ref{sec:eval}, we show extensive experimental evaluations and comparisons that highlight the versatility of \kapture~and provide interesting insights on visual localization using different datasets.

\begin{table}[]
\center
\resizebox{\linewidth}{!}{%
\begin{tabular}{|c|c|c|c|c|c|c|}
\hline
\rowcolor[HTML]{F5E0C7} 
\textbf{Data} & \textbf{Bundler} & \textbf{COLMAP} & \textbf{OpenMVG} & \textbf{OpenSfM} & \textbf{VisualSFM} & \textbf{\kapture} \\ \hline \hline
\rowcolor[HTML]{EFEFEF} 
\multicolumn{7}{|c|}{Sensor data} \\ \hline
Camera intrinsics & X & X & X & X & X & X \\ \hline
Extrinsics & X & X & X & X & X & X \\ \hline
Camera rig & - & (X) & - & - & - & X \\ \hline
GNSS (GPS) & - & (X) EXIF & (X) EXIF & X & (X) EXIF & X \\ \hline
LiDAR point cloud & - & - & - & - & - & X \\ \hline
Depth map & - & (X) & - & (X) & - & X \\ \hline
Other sensors & - & - & - & - & - & X \\ \hline \hline
\rowcolor[HTML]{EFEFEF} 
\multicolumn{7}{|c|}{Reconstruction data} \\ \hline
Local features & X & X & X & X & (X) & X \\ \hline
Global features & - & - & - & - & - & X \\ \hline
Matches & X & X & X & X & X & X \\ \hline
Observations & X & X & X & X & X & X \\ \hline
3D points & X & X & X & X & X & X \\ \hline
\end{tabular}
}
\vspace{0.1cm}
\caption{Comparison of various SFM tools and data formats. X:~supported, (X):~partly~supported, -:~not~supported}
\label{tab:sfmdataformat}
\vspace{-5mm}
\end{table}

\section{\Kapture~description}
\label{sec:kapture}

As briefly summarized in Figure~\ref{fig:banner}, \kapture~consists of the following core features we think are essential for modern visual localization\footnote{A full format specification and API is provided on the \kapture~website.}.

\PAR{Sensor data.}
The minimum data needed for visual localization are images as well as intrinsic and extrinsic camera parameters (supporting various camera and image distortion models).
If multiple cameras are used, their data should be associated with timestamps and the relative transformations between these cameras should be provided.
This allows to define static, ideally time-synchronized, camera rigs and the usage of image sequences (Section~\ref{sec:eval:datasets} presents examples for both).
Furthermore, modern location-based services not only use images for localization but rather combine visual localization with data coming from other sensors such as WiFi~\cite{wifiloc}, GNSS~\cite{gnssloc}, or IMUs~\cite{imuloc}.
Finally, self-driving cars or mobile robot platforms are often equipped with 3D sensors such as Lidar scanners~\cite{lidarloc} or RGBD cameras~\cite{rgbdloc}. 
The \kapture~format supports all of this and provides a clear definition of cameras, sensors, and their respective parameters.

\PAR{Reconstruction data.}
Another important aspect is handling processed data such as local and global image features, keypoint matches, keypoint observations, 3D points, and image lists that contain image pairs to be matched.
The \kapture~format unifies the definition of such data.

\PAR{Processing tools.}
Since \kapture~is designed to facilitate the use of existing and future open source tools and datasets, it consists of a growing set of conversion tools between existing formats.
\kapture~also contains tools to perform the entire visual localization pipeline presented in this paper.
Finally, \kapture~provides tools to merge, split, comparing, and evaluate datasets.

\PAR{Datasets.}
\kapture~is not only a data format definition, we also provide all datasets used in this paper for download using a first of its kind dataset downloader (included in the toolbox). 

\PAR{Open source.}
In an effort to ease further research in the field, we publicly release \kapture. 
We believe that it will facilitate large-scale experiments with a multitude of datasets, it will provide new opportunities to combine existing tools, and it will unveil new directions for benchmarking, visual localization and SFM research, as well as sensor fusion.

\subsection{Mapping and localization pipeline}
\label{sec:method}

\begin{figure}[ttt]
    \centering
    \includegraphics[width=0.9\linewidth]{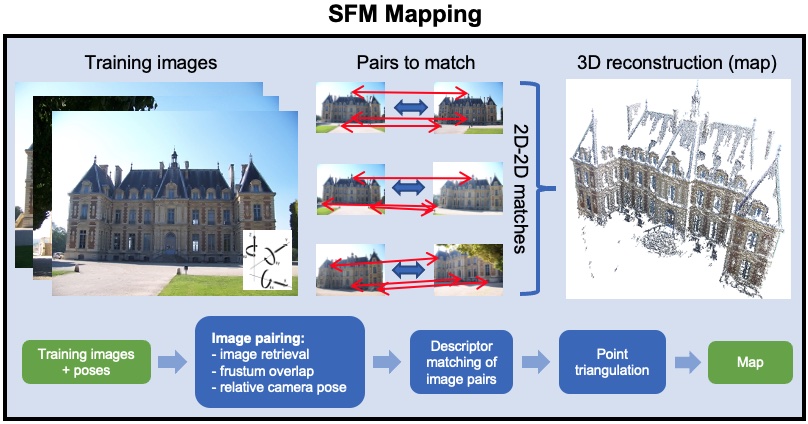}
    \includegraphics[width=0.9\linewidth]{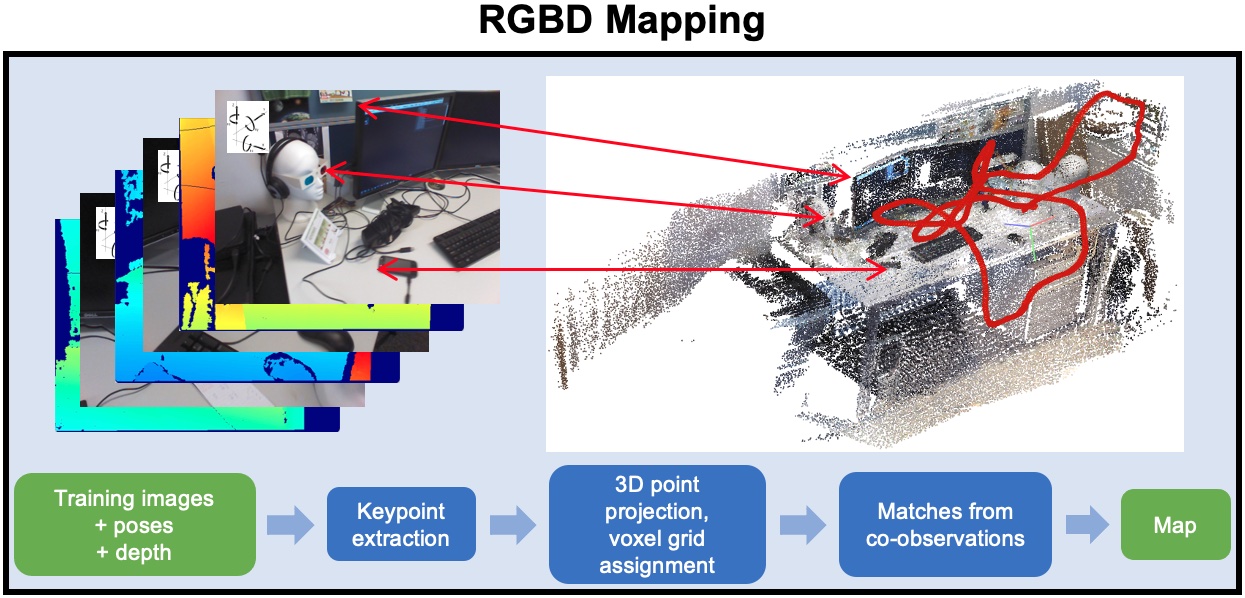}
     \caption{SFM and RGBD mapping pipelines. Image credit: 7-scenes~\cite{ShottonCVPR13SceneCoordinateRegression} and Sceaux Castle image dataset~\cite{openMVG}.}
    \label{fig:mapping}
    \vspace{-2mm}
\end{figure}

\begin{figure}[ttt]
    \centering
    \includegraphics[width=0.9\linewidth]{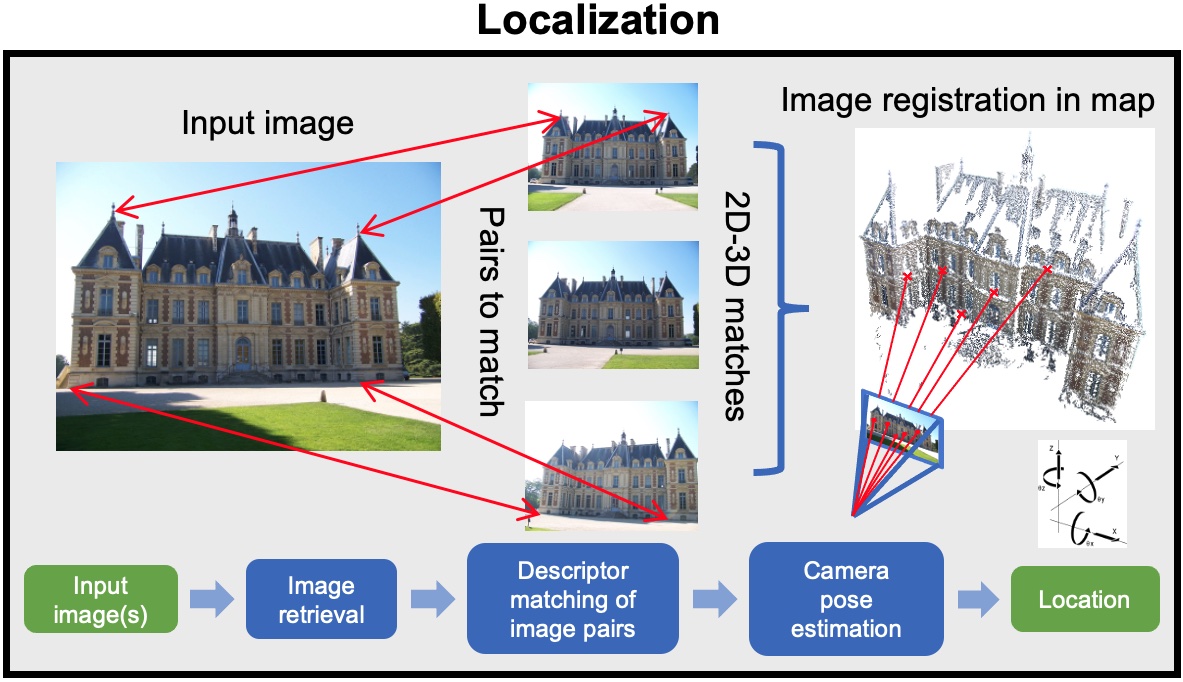}
     \caption{Localization pipeline. Image credit: Sceaux Castle image dataset~\cite{openMVG}.}
    \label{fig:localization}
    \vspace{-5mm}
\end{figure}

In order to demonstrate the versatility of \kapture, we implemented and evaluated a structure-based localization pipeline (Figure~\ref{fig:localization}) utilizing images, camera poses, and depth maps.
We follow the workflow of image retrieval as well as structure-based methods for mapping (Figure~\ref{fig:mapping}) and localization (Figure~\ref{fig:localization}).
For image retrieval, we use multiple global image representations (Section~\ref{sec:eval:global}), for local feature matching, we use state-of-the-art methods such as R2D2~\cite{RevaudNIPS19R2D2ReliableRepeatableDetectorsDescriptors}, SIFT~\cite{LoweIJCV04DistinctiveImageFeaturesScaleInvariantKeypoints}, D2-Net~\cite{DusmanuCVPR19D2NetDeepLocalFeatures}, and ASLFeat~\cite{aslfeatCVPR20} (Section~\ref{sec:eval:local}).
For 3D point triangulation (\eg during mapping), we use the COLMAP SFM library~\cite{SchonbergerCVPR16StructureFromMotionRevisited}. 
For image registration, we again use COLMAP, but also RansacLib~\cite{Sattler2019Github,Lebeda2012BMVC}, and pycolmap~\cite{Dusmanu2020pycolmap}. 
We compare these variants in Section~\ref{sec:eval:pipeline} and use COLMAP as default configuration for all other experiments.

\PAR{Mapping.} We use two strategies to generate 3D maps, SFM and RGBD.
SFM is one of the most popular strategies for reconstruction of a 3D scene from un-ordered photo collections~\cite{Snavely08IJCV,Heinly2015CVPR,SchonbergerCVPR16StructureFromMotionRevisited,sfm_survey_2017} and is often used to create 3D models used for visual localization. 
Figure~\ref{fig:mapping} (top) illustrates our approach where we first establish 2D-2D correspondences between local image features (keypoints) of image pairs, followed by geometric verification to remove outliers. 
By exploiting transitivity, observations of a keypoint found in several images allow to apply relative pose estimation for initialization of the reconstruction followed by 3D point triangulation~\cite{KangPR14RobustMultiViewL2Triangulation} and image registration for accurate 6DOF camera pose estimation.
RANSAC~\cite{Fischler81CACM,Chum08PAMI,Lebeda2012BMVC} can be used to increase robustness of several steps in this pipeline and bundle adjustment~\cite{triggs1999bundle} can be used for global (and local) optimization of the model (3D points and camera poses).
Note that the visual localization datasets we use in this paper already provide camera poses for the training images, thus, we can skip the camera pose estimation of SFM and directly triangulate the 3D points from the keypoint matches.
Since matching all possible training image pairs of a dataset becomes very processing time intensive, we use several strategies (see Figure~\ref{fig:mapping}) to create a short list of images to match. 
\emph{Image retrieval} uses global image features to pair visually similar images, \emph{frustum overlap} pairs images by the size of their overlapping camera frusta, and \emph{relative camera pose} uses the position and orientation difference between cameras to select images that potentially observe the same scene.
Note that \kapture~also provides a pairing scheme that uses co-observations of 3D points obtained from a pre-constructed SFM map.

When RGBD data (\eg depth maps) is available, as illustrated in Figure~\ref{fig:mapping} (bottom), instead of triangulating keypoint matches, we construct the 3D map by projecting the keypoints to 3D space using the provided depth.
If needed, co-observations of the same 3D point can be used to create keypoint correspondences across multiple images.

\PAR{Localization.} Our localization method (Figure~\ref{fig:localization}), similarly to mapping, establishes 2D-2D local feature correspondences between the query image and the database images used to generate the map. 
In order to only match relevant images, we use image retrieval to obtain the top $k$ most similar images from the database (we found that $k=20$ was in general a good choice). 
Since many keypoints from the database images correspond to 3D points of the map, 2D-3D correspondences between query image and map can be established. 
These 2D-3D matches are then used to compute the 6DOF camera pose by solving a PNP problem~\cite{KneipCVPR11ANovelParametrizationAbsoluteCamPose,Kukelova13ICCV,Larsson2017ICCV} robustly inside a RANSAC loop~\cite{Fischler81CACM,Chum08PAMI,Lebeda2012BMVC}.

\section{Experiments}
\label{sec:eval}

In this section, we present a thorough evaluation of the proposed pipeline and methods.
First, we present the evaluation protocol and the datasets used (Section~\ref{sec:eval:datasets}).
Second, we compare visual localization performance of four state-of-the-art local feature types (Section~\ref{sec:eval:local}) and four state-of-the-art global image representations as well as late fusion strategies to combine them (Section~\ref{sec:eval:global}).
Third, we present results obtained with different variants of our structure-based pipeline (Section~\ref{sec:eval:pipeline}). 
Fourth, we show how to leverage image sequences as well as camera rigs for visual localization (Section~\ref{sec:eval:rig}).
Fifth, we compare our SFM-based mapping method with our RGBD-based approach (Section~\ref{sec:eval:rgbd}).
Finally, we compare the best configurations of our pipeline with other top performing state-of-the-art methods on all datasets used in this paper.
In order to improve readability, in this section, we only present a summary of the results that best show the versatility and impact of \kapture.
More results can be found in Appendix~\ref{sec:experiments}.

\subsection{Protocol and datasets}
\label{sec:eval:datasets}

For all experiments, we followed our mapping and localization pipelines described in Section~\ref{sec:method}.
 Since, among others, we use COLMAP in this pipeline, we experimented with multiple parameter sets. 
Besides the default parameters, \textbf{config1}, we found that less strict bundle adjustment settings \textbf{config2} produce better results for some datasets (see  Table~\ref{tab:params} for the parameter configuration details). We indicate the used parameters for each experiments, but report the results for the best parameter configuration only.

\begin{table}[t]
\center
\resizebox{\linewidth}{!}{
\begin{tabular}{|l|c|c|}
\hline
\rowcolor[HTML]{F5E0C7} 
\textbf{\lstinline|COLMAP image\_registrator|} & \textbf{config1} & \textbf{config2} \\ \hline \hline
\lstinline|--Mapper.ba\_refine\_focal\_length| & 0 & 0 \\ \hline
\lstinline|--Mapper.ba\_refine\_principal\_point| & 0 & 0 \\ \hline
\lstinline|--Mapper.ba\_refine\_extra\_params| & 0 & 0 \\ \hline
\lstinline|--Mapper.min\_num\_matches| & 15 & 4 \\ \hline
\lstinline|--Mapper.init\_min\_num\_inliers| & 100 & 4 \\ \hline
\lstinline|--Mapper.abs\_pose\_min\_num\_inliers| & 30 & 4 \\ \hline
\lstinline|--Mapper.abs\_pose\_min\_inlier\_ratio| & 0.25 & 0.05 \\ \hline
\lstinline|--Mapper.ba\_local\_max\_num\_iterations| & 25 & 50 \\ \hline
\lstinline|--Mapper.abs\_pose\_max\_error| & 12 & 20 \\ \hline
\lstinline|--Mapper.filter\_max\_reproj\_error| & 4 & 12 \\ \hline
\end{tabular}
}
\vspace{0.1cm}
\caption{COLMAP parameter configurations used in our experiments.}
\label{tab:params}
\end{table}

For evaluation of our methods, we chose the datasets provided by the online visual localization benchmark~\cite{LTVL}
introduced in~\cite{SattlerCVPR18Benchmarking6DoFOutdoorLoc}, which consist of four outdoor datasets, namely Aachen Day-Night~v1.1.~\cite{SattlerCVPR18Benchmarking6DoFOutdoorLoc,Sattler2012BMVC,ZhangIJCV20ReferencePoseGenerationVisLoc}, RobotCar Seasons~v2~\cite{SattlerCVPR18Benchmarking6DoFOutdoorLoc}, Extended CMU-Seasons~\cite{SattlerCVPR18Benchmarking6DoFOutdoorLoc,Badino2011}, SILDa Weather and Time of Day~\cite{balntas2019silda}, and one indoor dataset, Inloc~\cite{Taira2019TPAMI,wijmans17rgbd}.
In addition, we use one more popular outdoor dataset, Cambridge Landmarks~\cite{KendallICCV15PoseNetCameraRelocalization}, and three more indoor datasets, 7-scenes~\cite{ShottonCVPR13SceneCoordinateRegression}, Baidu-mall~\cite{SunCVPR17DatasetBenchmarkingLocalization}, and Gangnam Station B2~\cite{LeeCVPR21LargeScaleLocalizationDatasetsCrowdedIndoorSpaces} for our experiments.
As evaluation metric, the five benchmark datasets, Baidu-Mall, and Gangnam Station use the percentage of query images which were localized within three pairs of translation and rotation error thresholds (the thresholds are provided in the respective results tables).
Cambridge Landmarks and 7-scenes use the median position and orientation error.

\begin{table*}[ttt]
\center
\resizebox{\linewidth}{!}{
\input{tables/lfeats_exps_arxiv}
}
\vspace{0.1cm}
\caption{Comparison of different local feature types on three datasets. The learned features R2D2, D2-Net, and ASLFeat often outperform SIFT, notably on difficult night images.}
\label{tab:local}
\vspace{-1mm}
\end{table*}

\begin{table*}[ttt]
\center
\resizebox{\linewidth}{!}{
\input{tables/fastr2d2}
}
\vspace{0.1cm}
\caption{Run-time (in seconds) and memory consumption comparison of R2D2 and Fast-R2D2 on four subsets of selected datasets. To minimize data I/O overhead, we did all data processing directly in the computers' RAM. Note, that for the large images of InLoc test, the 16GB GPU even ran out of memory while running original R2D2.}
\label{tab:fastr2d2_time}
\vspace{-4mm}
\end{table*}

\subsection{Local features}
\label{sec:eval:local}

Table~\ref{tab:local} compares four local features types on our SFM pipeline using image retrieval (top 20) with APGeM~\cite{RevaudICCV19LearningwithAPTrainingImgRetrievalListwiseLoss} features.
Next to the features R2D2~\cite{RevaudNIPS19R2D2ReliableRepeatableDetectorsDescriptors}, D2-Net~\cite{DusmanuCVPR19D2NetDeepLocalFeatures}, ASLFeat~\cite{aslfeatCVPR20}, and SIFT~\cite{LoweIJCV04DistinctiveImageFeaturesScaleInvariantKeypoints}, we present results obtained with a modified version of R2D2 that we call Fast-R2D2. 

Since the architecture of R2D2 is based on the fully convolutional encoder of L2-net~\cite{l2net}, the network will always process the image in its original resolution.
While this is a strong asset in the sense that it preserves fine details in the computed features, the computational cost is consequently heavier than that of a more standard convolution+pooling step.
In Fast-R2D2, we propose to merge both strategies by taking the R2D2 architecture and adding a single downsampling step inside the network to decrease the computation complexity while staying close to the original image resolution.
We restore the input resolution with an upsampling step right before the last convolutions.
This modification results in a threefold increase in computation speed and a significantly smaller memory footprint (see Table~\ref{tab:fastr2d2_time}) while, interestingly, achieving slightly better localization performance than the original architecture (Table~\ref{tab:local}).
More details about the architecture of Fast-R2D2 can be found in Appendix~\ref{sec:fastr2d2}.

The evaluation on three datasets in Table~\ref{tab:local} confirms the finding of \cite{RevaudNIPS19R2D2ReliableRepeatableDetectorsDescriptors,DusmanuCVPR19D2NetDeepLocalFeatures} that learned features often outperform classic SIFT, especially on the night images of Aachen and RobotCar.
On Cambridge Landmarks, SIFT performs comparable to R2D2, D2-Net, and ASLFeat.

\begin{table*}[ttt]
\center
\resizebox{\linewidth}{!}{
\input{tables/gfeat_exps_arxiv}
}
\vspace{0.1cm}
\caption{Comparison of different global image representations as well as a late fusion strategy on three datasets. The learned descriptors outperform the DenseVLAD on the night images of Aachen and GHARM consistently outperforms the individual representations.}
\label{tab:global}
\end{table*}

\subsection{Global image representations}
\label{sec:eval:global}

In this section, we use our SFM pipeline to compare different global descriptors used for image retrieval.
Similar to the experiments in the previous section, we fixed all parameters of the localization pipeline and only changed the image retrieval part.
We selected \textbf{DenseVLAD~\cite{ToriiCVPR15PlaceRecognitionByViewSynthesis}} and \textbf{NetVLAD~\cite{ArandjelovicCVPR16NetVLADPlaceRecognition}}, since they were already successfully used for visual localization~\cite{ToriiCVPR15PlaceRecognitionByViewSynthesis,SattlerCVPR17AreLargeScale3DModelsNecessaryForLocalization,SattlerCVPR19UnderstandingLimitationsPoseRegression,SarlinCVPR19FromCoarsetoFineHierarchicalLocalization,Germain20193DV,DusmanuCVPR19D2NetDeepLocalFeatures} before, and we added two new methods, \textbf{APGeM}~\cite{RevaudICCV19LearningwithAPTrainingImgRetrievalListwiseLoss} and \textbf{DELG}~\cite{CaoX20UnifyingDeepLocalGlobalFeatures} because both methods perform very well on the popular landmark retrieval benchmarks $\mathcal{R}$Oxford and $\mathcal{R}$Paris \cite{RadenovicCVPR18RevisitingOxfordParisImRetBenchmarking}. 
For our experiments, we used the code and models available at~\cite{densevlad,netvlad,apgem,delg}.

In addition to these image descriptors, we report results obtained by late fusion of all four of them.
Late fusion means that we first compute the similarities for each descriptor individually and then apply a fusion operator.
We used the operator \emph{generalized weighted harmonic mean} from \cite{csurka2012empirical} (GHARM) for this experiment since it tends to be more robust against outliers than geometric or arithmetic mean. 
More details about late fusion for image retrieval can be found in Appendix~\ref{sec:fusion}.
Table~\ref{tab:global} summarizes the results of our experiments.
We see that the learned image descriptors outperform the handcrafted descriptor DenseVLAD on the night images of Aachen and that GHARM consistently outperforms the individual representations. 
Late fusion, as it is robust to outliers, indeed is interesting to improve robustness and reliability.

\subsection{Pipeline variants}
\label{sec:eval:pipeline}

In this section, we present localization results obtained with different image pairing methods as well as different  image registration variants using 2D-3D correspondences.

\PAR{Pairing.} We compare the strategies \emph{image retrieval}, \emph{distance}, and \emph{frustum overlap} for image pair selection during the mapping process.
For localization, we use APGeM for image retrieval (top 20) and R2D2 for local feature matching.
Note that, contrary to the place recognition literature~\cite{ToriiPAMI15VisualPlaceRecognRepetitiveStructures,ArandjelovicACCV14DislocationDistinctivenessForLocation,SattlerCVPR16LargeScaleLocationRecognitionGeometricBurstiness}, where \emph{distance} is usually used as binary score (an image is relevant if it is below a position and orientation threshold $\tau_c$ resp.~$\tau_R$), we rank the images using the normalized distance score $s(q,t)= \frac{c_\text{diff}}{\tau_c} + \frac{R_\text{diff}}{\tau_R}$, where $c_\text{diff}$ and $R_\text{diff}$ are computed between the pose of image $q$ $(\mathbf{c}_\text{q},\mathtt{R}_\text{q})$ and the pose of image $t$ $(\mathbf{c}_\text{t},\mathtt{R}_\text{t})$.
The role of $\tau_c$ and $\tau_R$ is to normalize position distance and angle, making them more comparable.
We use $\tau_c=25m$ and $\tau_R=45^\circ$ in our experiment.
The results on two datasets in Table~\ref{tab:pairing} show that \emph{distance} performs best, followed by image retrieval.
This suggests that \emph{distance}, even if it does not take visual content into account and as a consequence might suffer from occlusions, is well suited to generate 3D maps without the need of processing intensive brute force matching or introducing bias by using specific global features.

\begin{table}[ttt]
\center
\resizebox{\linewidth}{!}{
\input{tables/mapping_pairsfile_exps}
}
\vspace{0.1cm}
\caption{Comparison of different image pairing strategies. Method \emph{distance} performs best, followed by image retrieval. Note that for \emph{image retrieval} we recommend to use the same descriptor as used for localization (APGeM in our experiments). All methods use, if available, the top 20 image pairs.}
\label{tab:pairing}
\vspace{-2mm}
\end{table}

\begin{table*}[ttt]
\center
\resizebox{\linewidth}{!}{
\input{tables/pipelines_exps_arxiv}
}
\vspace{0.1cm}
\caption{Comparison of different variants of our pipeline. We use APGeM-20 and R2D2 features for all variants. \emph{config2} is a set of relaxed bundle adjustment parameters for COLMAP that allow more images to be localized, even if the individual error is higher.}
\label{tab:pipeline}
\vspace{-4mm}
\end{table*}

\PAR{Image registration.} While many algorithms are directly implemented in \kapture~(\eg feature extraction, keypoint matching, image pairing, etc.), it also enables the easy use of methods provided in external libraries.
To demonstrate this, \kapture~provides native support for several open source tools (see Section~\ref{sec:kapture}), including interfaces to COLMAP~\cite{SchonbergerCVPR16StructureFromMotionRevisited}, Python bindings for pose estimators of COLMAP~\cite{Dusmanu2020pycolmap}, as well as RansacLib~\cite{Sattler2019Github,Lebeda2012BMVC,kneip2014opengv}, a template-based library that implements different RANSAC variants.

In this section we use these implementations to compare three variants of image registration that can be used with \kapture.
The variant COLMAP stands for COLMAP's image registrator which needs the dataset to be fully converted to COLMAP's format (\eg using \kapture). Pycolmap allows direct access to COLMAP's absolute camera pose estimators without the need of data import, and RansacLib provides easy access to multiple RANSAC implementations, also directly accessible from \kapture~format.
Table~\ref{tab:pipeline} shows that all variants perform similarly with a small advantage of COLMAP.
Note that we did not evaluate the full range of RANSAC implementations of RansacLib but we think that this is an interesting task for future work.

\subsection{Camera rigs and image sequences}
\label{sec:eval:rig}

In order to evaluate how camera rigs and image sequences can be used to improve localization performance, we implemented two post-processing steps applicable for any localization method and one algorithm that directly uses a multi-camera rig during localization. 
We evaluated them on RobotCar, Extended-CMU, and SILDa (these datasets provide synchronized camera rigs).
RobotCar provides the camera rig calibration (transforms between the cameras) as part of the dataset.
For Extended-CMU and SILDa, we estimated it using the training poses.

The first post-processing step (\emph{rig}) uses the camera rig calibration to compute the pose for all images of the rig given one successfully localized camera.
The second post-processing step assigns poses to non-localized images within query sequences (defined using timestamps) using linear interpolation between the two closest successfully localized images.
If this is not possible, we use the nearest neighbor.
Table~\ref{tab:rig} shows that the highest improvement over the single camera setup can be achieved with the camera rig.
Furthermore, utilizing the image sequences gives a little boost for the datasets we used for our experiments.

While these two post-processing methods can be used to spatially and temporally complete not successfully localized images, they do not directly leverage multi-camera rigs.
To demonstrate this, we implemented an interface to the generalized P3P (GP3P)~\cite{Kukelova2016CVPR,kneip2014opengv} implementation of \cite{Sattler2020multicampose,wald2020}, a method that utilizes known relative poses between multiple images for multi-camera pose estimation. 
We evaluated this method on the RobotCar dataset and observe that it performs better than using only one camera and similarly to using rig and sequence post-processing.
This method is useful if, \eg, the individual cameras on a rig do not find enough local feature matches to localize.
More investigation of such scenarios is an interesting direction for future work using \kapture.

\begin{table*}[ttt]
\center
\resizebox{\linewidth}{!}{
\input{tables/rig_arxiv}
}
\vspace{0.1cm}
\caption{Evaluation of rig and image sequence post-processing on three datasets. \emph{config1} is the default set of COLMAP parameters that result in accurate poses but might not be able to localize all images. We complete those with \emph{seq} and \emph{rig}.}
\label{tab:rig}
\vspace{-3mm}
\end{table*}

\subsection{RGBD-based localization}
\label{sec:eval:rgbd}

\begin{figure}[ttt]
    \centering
    \includegraphics[width=0.48\linewidth]{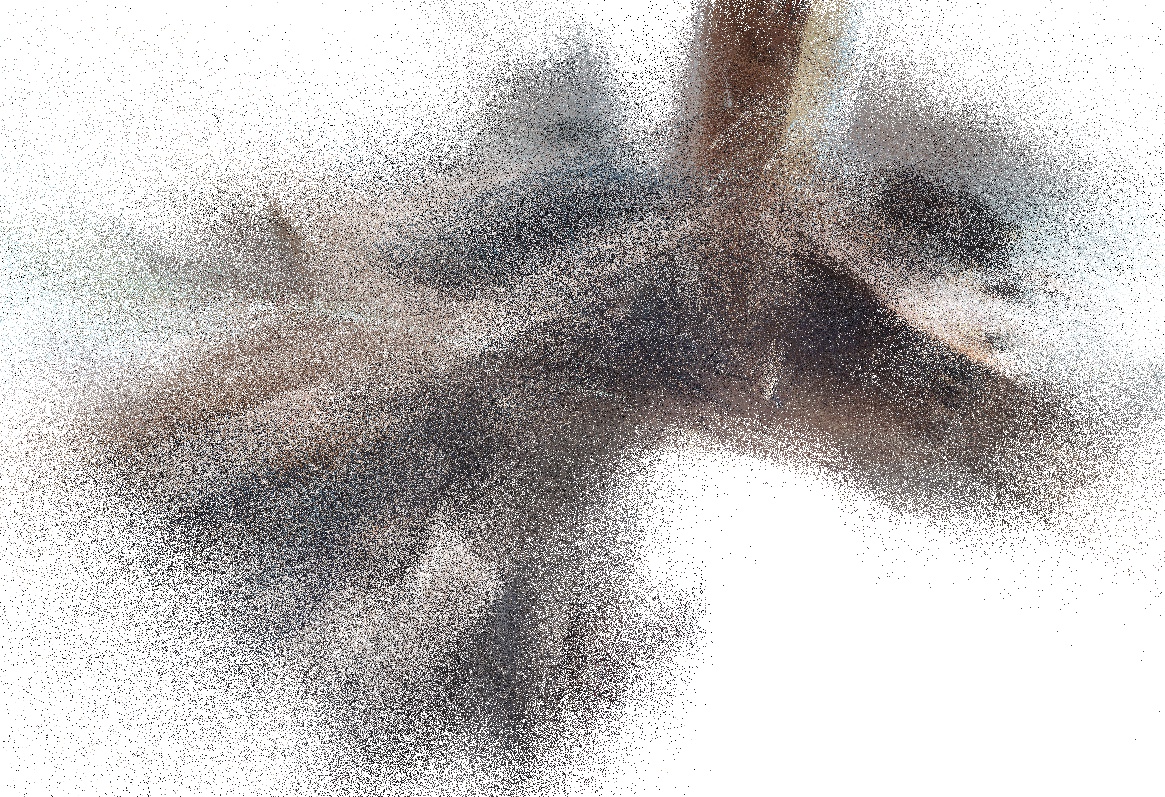}
    \includegraphics[width=0.48\linewidth]{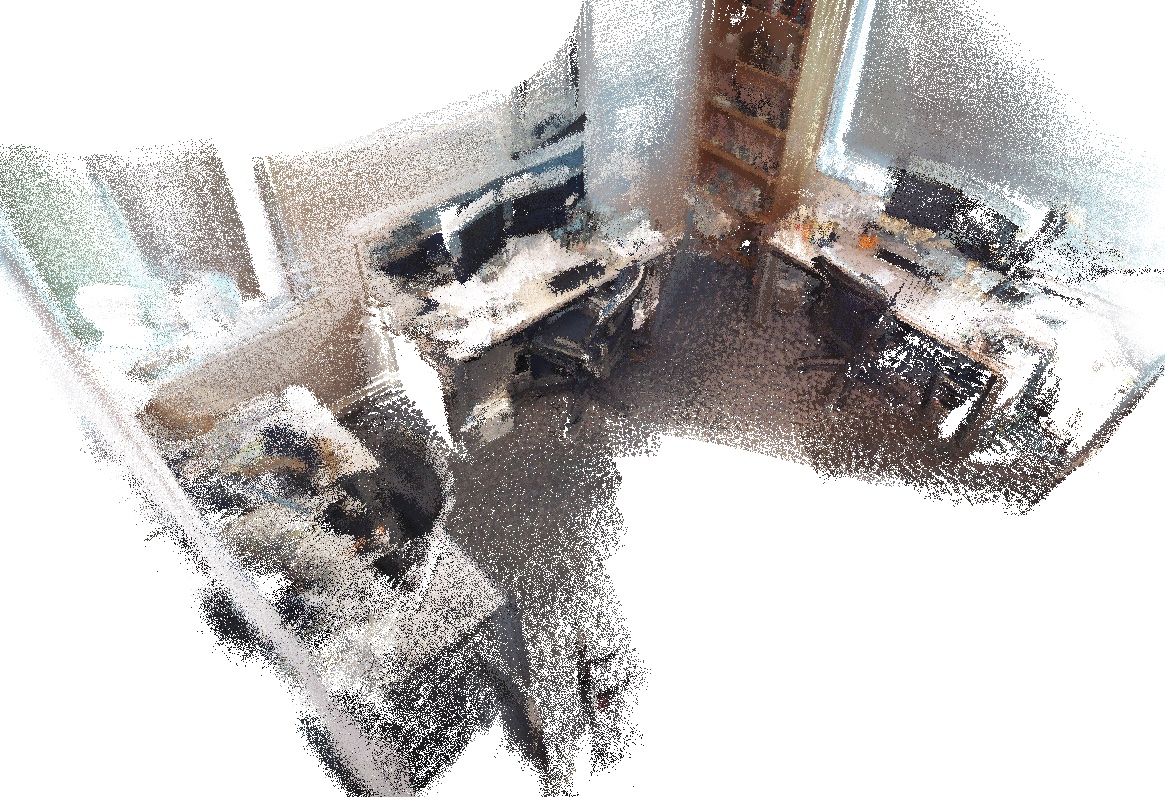}
    \caption{Comparison of the 7-scenes~\cite{ShottonCVPR13SceneCoordinateRegression} (office) 3D map created by triangulating R2D2 features (left) and by projecting the features to 3D using the provided depth maps.}
    \label{fig:map_office}
    \vspace{-3mm}
\end{figure}

An accurate 3D representation of the environment is crucial for precise camera localization. 
So far, in this paper, we constructed the maps by triangulating feature matches using the provided training images and poses.
In this section, we show the impact of using depth maps for this step on the RGBD dataset 7-scenes~\cite{ShottonCVPR13SceneCoordinateRegression}.
Figure~\ref{fig:map_office} shows two maps of the same scene, SFM (left) and RGBD (right), where a clear difference in reconstruction quality can be seen: the SFM map is noisier than the RGBD map.
We explain this by the fact that the RGBD camera is not calibrated and that the camera poses are not accurate enough for precise keypoint triangulation.
Contrary, when directly using depth from a dedicated sensor, the 3D reconstruction depends less on the camera poses (because the 3D points do not come from triangulation), which explains the cleaner map produced by RGBD.
Table~\ref{tab:rgbd} compares localization performance on these two maps (for both experiments we use APGeM top 20 and R2D2 for localization) confirming this observation.
We see a significant improvement in accuracy when using RGBD.

The second RGBD dataset we use in this paper is InLoc~\cite{Taira2019TPAMI,wijmans17rgbd} which represents an indoor scenario captured in multiple university buildings (only DUC1 and DUC2 are used here). 
There is only little overlap between the training images of this dataset which results in an SFM model which is, according to our experience, too sparse for visual localization pipelines. 
Instead, similarly as described above, we constructed the map using the provided 3D points which results in a very dense 3D reconstruction which can be used for the localization experiments presented in Table~\ref{tab:summary}.
We are convinced that \kapture~will enable more research on RGBD-based localization in the future.

\begin{table}[ttt]
\center
\resizebox{\linewidth}{!}{
\input{tables/rgbd_exps}
}
\vspace{0.1cm}
\caption{Comparison of SFM and RGBD mapping.}
\label{tab:rgbd}
\vspace{-2mm}
\end{table}

\subsection{Comparison to state-of-the-art}
\label{sec:eval:summary}

Finally, we compare our pipeline with top performing state-of-the-art methods on eight popular datasets.
In Table~\ref{tab:summary} (datasets from \cite{LTVL}), we see that our methods are top ranked on all datasets (with leading positions on two of them) but InLoc, which is very challenging for image retrieval and local feature matching. The top performing methods on InLoc either leverage semantic and depth cues~\cite{fan2020visual} or benefit from more advanced matching techniques~\cite{SarlinCVPR20SuperGlueLearningFeatureMatchingGNN}. 
Table~\ref{tab:summary_baidu} presents results on the Baidu-mall dataset.
The baseline results~\cite{SunCVPR17DatasetBenchmarkingLocalization} were computed using a map consisting of 3D points created by a Lidar scanner. 
Still, only using SFM with frustum-based pairing and top 50 DenseVLAD image features, our method outperforms the current state-of-the-art.
Table~\ref{tab:summary_median} shows a more comprehensive evaluation on 7-scenes~\cite{ShottonCVPR13SceneCoordinateRegression} and Cambridge Landmarks~\cite{KendallICCV15PoseNetCameraRelocalization}. 
We report results of the best performing algorithms in the literature and show that our methods outperform all of them (using RGBD on 7-scenes and SFM on Cambridge Landmarks).

\begin{table*}[ttt]
\center
\resizebox{\linewidth}{!}{
\input{tables/summary3}
}
\vspace{0.1cm}
\caption{Comparison of our methods with top performing state-of-the-art methods on the datasets from \cite{LTVL}. Our methods are top ranked on most of them.}
\label{tab:summary}
\end{table*}

\begin{table}[]
\center
\resizebox{\linewidth}{!}{
\input{tables/summary_baidu}
}
\vspace{0.1cm}
\caption{Comparison of our method with top performing state-of-the-art methods on the Baidu-Mall dataset. All baseline results were taken from \cite{SunCVPR17DatasetBenchmarkingLocalization} and, contrary to ours, computed using a Lidar map.}
\label{tab:summary_baidu}
\end{table}

\begin{table*}[ttt]
\center
\resizebox{\linewidth}{!}{
\input{tables/summary_median}
}
\vspace{0.1cm}
\caption{Comparison of our methods with top performing state-of-the-art methods on 7-scenes~\cite{ShottonCVPR13SceneCoordinateRegression} and Cambridge Landmarks~\cite{KendallICCV15PoseNetCameraRelocalization}. Our methods defines a new state-of-the-art on both.}
\label{tab:summary_median}
\vspace{-3mm}
\end{table*}

\section{Conclusion}

We present \kapture, a new, open, extendable data format and toolbox to facilitate processing and evaluation of visual localization and SFM.
To demonstrate \kapture, we propose a versatile pipeline that enables research on various aspects of visual localization.
We evaluate localization performance using multiple global and local feature types, using different variants for image registration, using multi-camera rigs as well as sequence processing for spacial and temporal fusion, and we show how depth data can be used to improve localization.
We present a modified version of R2D2 that is three times faster than the original version and increases localization performance on most datasets.
Furthermore, we show that late fusion of image representation consistently outperforms image retrieval using individual image descriptors.
Finally, our method ranks among the best methods on all datasets used in this paper, defining a new state-of-the-art on five of them.
We can conclude that a combination of robust global (\eg APGeM) and local (\eg R2D2) features is a good choice for visual localization on a large variety of application scenarios.
The \kapture~toolbox, all datasets used converted to \kapture~format, as well as the presented algorithms are publicly released at \url{github.com/naver/kapture} and \url{github.com/naver/kapture-localization}.

{\small
\bibliographystyle{ieee_fullname}
\bibliography{egbib}
}

\vspace{0.5cm}

\appendix

\noindent \textbf{\large APPENDIX}  \\

In the appendix, we first provide a detailed description of the proposed Fast-R2D2 architecture (Section~\ref{sec:fastr2d2}) as well as the proposed late fusion techniques (Section~\ref{sec:fusion}).
Second, we present additional experiments that complement the findings we discuss in the main paper (Section~\ref{sec:experiments}).

\section{Fast-R2D2}
\label{sec:fastr2d2}

R2D2~\cite{RevaudNIPS19R2D2ReliableRepeatableDetectorsDescriptors} is a sparse keypoint extractor that jointly performs detection and description but separately estimates keypoint reliability and keypoint repeatability.
Keypoints with high likelihoods on both aspects are chosen, which improves the overall feature matching pipeline. 
In this section, we revisit the architecture of the network
and present a significantly faster model (see Table~\ref{tab:fastr2d2_time}) that achieves the same order of precision as the original network, thus better suited for practical localization tasks in section~\ref{sec:fastr2d2arch}.
The details of the model training are explained in section~\ref{sec:fastr2d2:training} and we present the  hyper-parameter search that motivated the chosen architecture
in section~\ref{sec:fastr2d2:gridsearch}. The  code and models for both R2D2 and Fast-R2D2 can be found at \url{github.com/naver/r2d2}.

\subsection{Architecture}
\label{sec:fastr2d2arch}

The core of the architecture is based on the fully convolutional encoder of \cite{l2net}, to which three branches are appended that respectively predict the descriptors, their repeatability, and their reliability. 
These branches are left as in the original R2D2 paper and we focus here on the core of the network.
As described in Figure~\ref{fig:archs} (\textit{top}), it consists of a stack of dilated convolution blocks with increasing dilation factors. 
For the same number of parameters compared to a standard convolution and pooling block, the dilated layer increases the receptive field while preserving the input resolution for maximal detail preservation. 
However, through the different layers of the network, the depth of the channels increases, and so does the number of floating point operations. 
Computation time is not problematic for small scale evaluations but hinders applicability in practice for larger scale mapping and localization tasks. 

We claim that we can achieve the same level of performances in a much faster fashion.
To that aim and similarly to SuperPoint~\cite{AndersonCVPR18SuperpointInterestPoint}, we propose to incorporate an early down-sampling step in the network in the form of a pooling layer, process the lower resolution image, and up-sample the output to the original resolution right before the three branches prediction.
At the end, we up-sample the feature map to the input resolution using bi-linear interpolation.

\begin{figure}
\centering
\includegraphics[width=1.\linewidth]{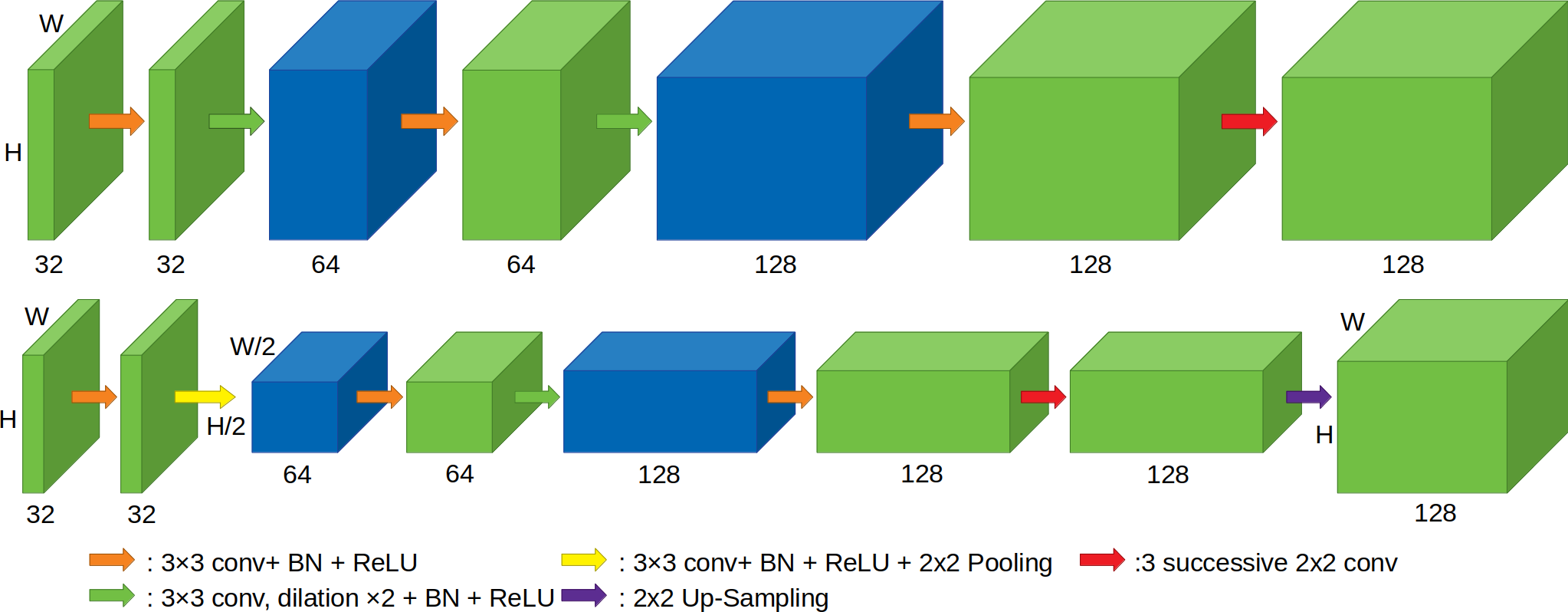} 
\caption{Comparison of the original core architecture~\cite{l2net} (\textit{top}) and the proposed one (\textit{bottom}). The three prediction branches (same as in \cite{RevaudNIPS19R2D2ReliableRepeatableDetectorsDescriptors}) are omitted for readability.}
\label{fig:archs}
\end{figure}

\subsection{Training and inference}
\label{sec:fastr2d2:training}

For the sake of fairness in the comparison, we train our faster architectures using the code of R2D2, with the exact same training data and losses. 
Both, the R2D2 and the Fast-R2D2 models were trained for $20$ epochs using synthetic image pairs generated by applying known transformations (homographies) on random web images~\cite{RadenovicCVPR18RevisitingOxfordParisImRetBenchmarking} and on images of the Aachen Day-Night dataset~\cite{SattlerCVPR18Benchmarking6DoFOutdoorLoc,Sattler2012BMVC,Zhang2020ARXIV}. 
We also use additional optical flow matches computed on Aachen Day-Night.
For inference, we use the original feature extraction code. Please refer to \cite{RevaudNIPS19R2D2ReliableRepeatableDetectorsDescriptors} for more details about the data generation, training, and inference procedures.

\subsection{Hyper-parameter selection}
\label{sec:fastr2d2:gridsearch}

In this section, we explain in more details the methodology we followed to find the optimal faster architecture.

In order to find the best set of hyper-parameters for the faster architecture, we grid search the possible variables, which are i)
the position of the pooling step: index between 0 and 6 (\eg a value of 2 means the pooling operation follows the convolution of index 2),
ii) the pooling type: AVG or MAX, and iii)  the down-sampling factor: 2 or 3.

We observe that any greater down-sampling factor, regardless of the architecture we tried, led to divergence of the training losses. Similarly, more than one pooling step induced large performance drops, so we experimented using only one.\\

\PAR{Models nomenclature.}
We denote the various models with $s$ for the pooling step position, $p$ for the pooling type, and $d$ for the down-sampling factor.

\begin{figure}
\centering
\includegraphics[width=1.\linewidth]{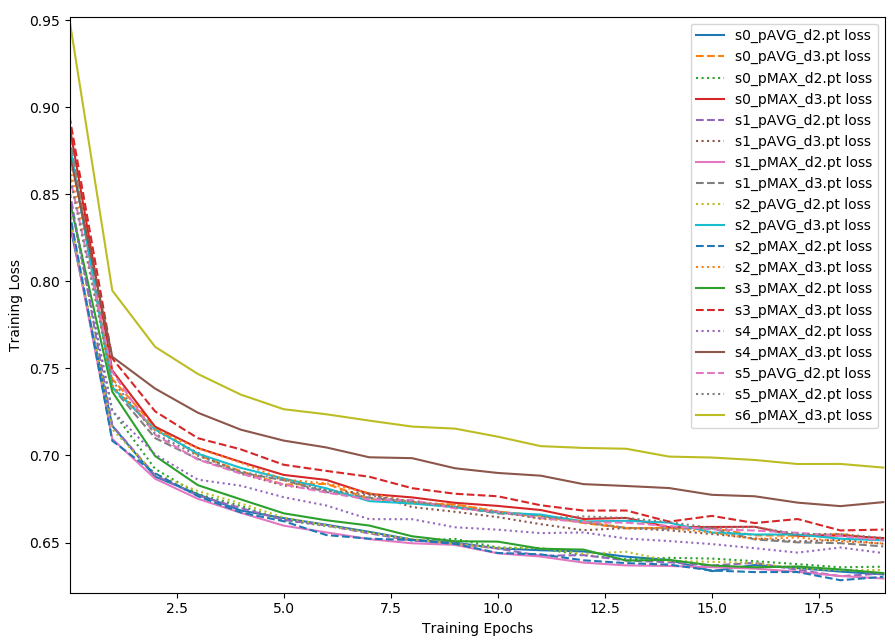}
\caption{Training losses of the explored down-sampling placements and types.}
\label{fig:archs}
\end{figure}

We plot the training losses of the various architectures in Figure~\ref{fig:archs}. Using this, we isolate good candidates for best performing faster architectures, which are 's1\_pMAX\_d2', 's2\_pMAX\_d2', 's0\_pAVG\_d2' and 's1\_pAVG\_d2'.

We then run localization experiments with these candidate models. 
In particular, using the same extraction procedure as~\cite{r2d2}, and the localization pipeline described in the main paper. 
We evaluate on the Aachen Day Night v1.1 public benchmark. 
We ran each experiment $4$ times and only the best result is kept, to account for small variations in the localization pipeline (due to RANSAC). 
Table~\ref{table:archsearch} summarizes the results of the chosen candidate architectures.

As can be seen in Table~\ref{tab:fastr2d2_time} of the main paper, the speed improvement of the chosen architecture is threefold. 
Interestingly, one would expect the down-sampling step to decrease accuracy in localization, however we observed that it was not the case and this faster architecture is among the best performing methods, as seen in the localization evaluations  Tables~\ref{tab:fastr2d2_time} and \ref{tab:summary} of the main paper.

\begin{table}[ttt]
\center
\resizebox{\linewidth}{!}{
\begin{tabular}{|l|c|c|c|c|c|c|c|}
\hline
\rowcolor[HTML]{F5E0C7} 
\textbf{Model} & \textbf{Avg.}  & \multicolumn{3}{c|}{\textbf{Day}} & \multicolumn{3}{c|}{\textbf{Night}} \\ \hline 
 & all bins & high & mid & low & high & mid & low \\  \hline 
s0\_pAVG\_d2 & 90.70 & 89.4 & 96.2 & 99.5 & 73.3 & 87.4 & 98.4 \\ \hline
s1\_pAVG\_d2 & 90.52 & 89.4 & 96.2 & 99.5 &	74.3 & 85.3 & 98.4 \\ \hline
\textbf{s1\_pMAX\_d2} & \textbf{90.98} & 89.9 & 96.4 & 99.5 &	74.3 & 87.4 & 98.4 \\ \hline
s2\_pMAX\_d2 & 90.90 & 90.2 & 96.2 & 99.4 &	74.3 & 87.4 & 97.9 \\ 
\hline
\end{tabular}
}
\vspace{0.1cm}
\caption{Localization performance of candidate architectures on Aachen Day Night v1.1. Final selected architecture in \textbf{bold}.}
\label{table:archsearch}
\end{table}

\section{Late fusion of global image representations}
\label{sec:fusion}
  
In the early literature, fusion techniques were often used to combine complementary visual cues measuring the similarity between images~\cite{JegouPAMI10AccurateImageSearchContextualDissimilarityMeasure,ZhangECCV12QuerySpecificFusionImageRetrieval} or different modalities~\cite{DepeursingeBC10FusionTechniquesCombiningTextualVisualRetr,csurka2012empirical,ZhengCVPR15QueryAdaptiveLateFusionReID}.
Mostly used fusion mechanisms are weighted score averaging
(\textbf{mean}) or powered product (\textbf{power}).
Power penalizes severely if one of the scores is low, thus, it can be interpreted as a logical "{\em and}".
Additionally, the \textbf{min} operator explicitly encodes the logical "{\em and}" (all scores need to be high) and \textbf{max} encodes the logical "{\em or}" (a single score needs to be high).

Csurka \etal in \cite{csurka2012empirical} show that the combination of these operators  in general further increases the robustness of the retrieval. 
Following this idea, together with the mentioned simple operators, we evaluate the following combined fusion operators.
\textbf{wmp} is a weighted combination of \textbf{mean} and \textbf{power}:
\begin{eqnarray*}
M_{wmp}(q,d)=\beta\cdot\sum_{i} \rho_i s_i(q,d) + (1-\beta)\cdot\prod_{i} s_i(q,d)^{\rho_i}
\end{eqnarray*}
and \textbf{wmm} is a weighted combination of \textbf{min} and \textbf{max}:
\begin{eqnarray*}
M_{wmm}(q,d)=(1-\beta)\max_i(s_i(q,d))+\beta\min_i(s_i(q,d))
\end{eqnarray*}
where $s_i(q,d)$ is the similarity score between the query $q$ and database image $d$ computed with the representation $i$. 
Generalized weighted harmonic mean
\textbf{gharm} can be obtained using the generalized f-mean operator:
\begin{eqnarray*}
M_f(x_1,...,x_n)=f^{-1}\left( \sum_{i=1}^n \frac{f(x_i)}{n}\right)
\end{eqnarray*}
with $f(x) = \frac{1}{\gamma+x}$, $x_i=\alpha_i s_i(q,d)$,  $\sum_{i=1}^n \alpha_i=1$, $n$ being the number of different image representations considered.

In addition to score aggregation, rank list aggregation methods are proposed in the literature~\cite{AhPineBC10LeveragingCrossMediaSimilarities,JegouPAMI10AccurateImageSearchContextualDissimilarityMeasure,ZhangECCV12QuerySpecificFusionImageRetrieval}. 
To complete our evaluation, we select \textbf{round robin} scheduling which creates a combined rank list by selecting new elements from each individual rank list in circular order.

\section{Further experimental comparisons}
\label{sec:experiments}

Contrary to the section~\ref{sec:eval}, here we present the results from above with complementary ones, this time organized per dataset.

In detail, if available, in Tables~\ref{tab:aachen}~-~\ref{tab:cambridge}, we present comparisons of: \begin{itemize}
    \item the six late fusion operators presented in Section~\ref{sec:fusion},
    \item the four global image representations used in Section~\ref{sec:eval:global} of the main paper,
    \item the five local feature types using in Section~\ref{sec:eval:local},
    \item state-of-the-art methods (no exhaustive list but the best methods),
    \item and two sets of COLMAP~\cite{SchonbergerCVPR16StructureFromMotionRevisited} parameters from Table~\ref{tab:params} (config1 and config2).
\end{itemize}
In addition to the main paper, we also present experimental results on the 12-scenes~\cite{Valentin20163DV} dataset.
We did not include these results in the main paper, because the dataset is already saturated at 100\% using the official error thresholds of 5cm and 5°.
Instead, in Table~\ref{tab:12scenes} we present results using tighter thresholds of 1cm and 1°.
Finally, Table~\ref{tab:pipeline_detailed} presents results of the three variants for image registration (see  Section~\ref{sec:eval:pipeline}) using top 1, 5, 10, and 20 retrieved images.

\begin{table*}[]
\center
\resizebox{0.9\linewidth}{!}{
\input{tables/aachen_v2}
}
\vspace{0.1cm}
\caption{Experiments summary for Aachen Day-Night~v1.1~\cite{SattlerCVPR18Benchmarking6DoFOutdoorLoc,Sattler2012BMVC,Zhang2020ARXIV} (sorted in descending order by average over all bins).}
\label{tab:aachen}
\end{table*}

\begin{table*}[]
\center
\resizebox{0.9\linewidth}{!}{
\input{tables/inloc_v2}
}
\vspace{0.1cm}
\caption{Experiments summary for InLoc~\cite{Taira2019TPAMI,wijmans17rgbd} (sorted in descending order by average over all bins). R2D2 with 40k keypoints.}
\label{tab:inloc}
\end{table*}

\begin{table*}[]
\center
\resizebox{0.9\linewidth}{!}{
\input{tables/robotcar_v2}
}
\vspace{0.1cm}
\caption{Experiments summary for RobotCar Seasons v2~\cite{SattlerCVPR18Benchmarking6DoFOutdoorLoc} (sorted in descending order by average over all bins).}
\label{tab:robotcar}
\end{table*}

\begin{table*}[]
\center
\resizebox{\linewidth}{!}{
\input{tables/ecmu_v2}
}
\vspace{0.1cm}
\caption{Experiments summary for Extended-CMU~\cite{SattlerCVPR18Benchmarking6DoFOutdoorLoc,Badino2011} (sorted in descending order by average over all bins).}
\label{tab:ecmu}
\end{table*}

\begin{table*}[]
\center
\resizebox{\linewidth}{!}{
\input{tables/silda_v2}
}
\vspace{0.1cm}
\caption{Experiments summary for SILDa~\cite{balntas2019silda} (sorted in descending order by average over all bins).}
\label{tab:silda}
\end{table*}

\begin{table*}[]
\center
\resizebox{0.6\linewidth}{!}{
\input{tables/baidu_suppl}
}
\vspace{0.1cm}
\caption{Comparison of variations of our method with top performing state-of-the-art methods on the Baidu-Mall dataset. All baseline results were taken from \cite{SunCVPR17DatasetBenchmarkingLocalization} and, contrary to ours, computed using a Lidar map. \emph{frustum-20} means that only the top 20 images are used, contrary to \emph{frustum}, which uses all overlapping pairs (same for \emph{distance-20} and \emph{APGeM-20}).}
\label{tab:baidu}
\end{table*}

\begin{table*}[]
\center
\resizebox{0.5\linewidth}{!}{
\input{tables/gangnam_b2_arxiv}
}
\vspace{0.1cm}
\caption{Experiments summary for Gangnam Station B2~\cite{LeeCVPR21LargeScaleLocalizationDatasetsCrowdedIndoorSpaces} (sorted in descending order by average over all bins).}
\label{tab:gangnam}
\end{table*}

\begin{table*}[]
\center
\resizebox{\linewidth}{!}{
\input{tables/7_scenes}
}
\vspace{0.1cm}
\caption{Experiments summary for 7-scenes~\cite{ShottonCVPR13SceneCoordinateRegression}.}
\label{tab:7scenes}
\end{table*}

\begin{table*}[]
\center
\resizebox{\linewidth}{!}{
\input{tables/12_scenes}
}
\vspace{0.1cm}
\caption{Experiments summary for 12-scenes~\cite{Valentin20163DV}.}
\label{tab:12scenes}
\end{table*}

\begin{table*}[]
\center
\resizebox{\linewidth}{!}{
\input{tables/cambridge_landmarks}
}
\vspace{0.1cm}
\caption{Experiments summary for Cambridge Landmarks~\cite{KendallICCV15PoseNetCameraRelocalization}. Note that the average median error is computed excluding the Street scene since only 2 methods (including ours) report results for this.}
\label{tab:cambridge}
\end{table*}

\begin{table*}[]
\center
\resizebox{\linewidth}{!}{
\input{tables/pipeline_exps_detailed_arxiv}
}
\vspace{0.1cm}
\caption{Comparison of different variants of our pipeline. We use R2D2 and APGeM with top 1, 5, 10, and 20 for all variants. The maps are created using pairs selected by frustum overlap.}
\label{tab:pipeline_detailed}
\end{table*}

\end{document}

%% file: tables/lfeats_exps_arxiv.tex
\begin{tabular}{ccccccccccll}
\hline
\rowcolor[HTML]{F5E0C7} 
\multicolumn{1}{|c|}{\cellcolor[HTML]{F5E0C7}\textbf{Datasets}} &
  \multicolumn{1}{c|}{\cellcolor[HTML]{F5E0C7}\textbf{Algorithm}} &
  \multicolumn{10}{c|}{\cellcolor[HTML]{F5E0C7}\textbf{percentage of successfully localized images}} \\ \hline
\multicolumn{1}{|c|}{} &
  \multicolumn{1}{c|}{} &
  \multicolumn{1}{c|}{} &
  \multicolumn{3}{c|}{day} &
  \multicolumn{3}{c|}{night} &
  \multicolumn{3}{c|}{} \\ \cline{4-9}
\multicolumn{1}{|c|}{} &
  \multicolumn{1}{c|}{\multirow{-2}{*}{\begin{tabular}[c]{@{}c@{}}APGeM-20\\      config2\end{tabular}}} &
  \multicolumn{1}{c|}{\multirow{-2}{*}{avg. all bins}} &
  \multicolumn{1}{c|}{high} &
  \multicolumn{1}{c|}{mid} &
  \multicolumn{1}{c|}{low} &
  \multicolumn{1}{c|}{high} &
  \multicolumn{1}{c|}{mid} &
  \multicolumn{1}{c|}{low} &
  \multicolumn{3}{c|}{} \\ \cline{2-9}
\multicolumn{1}{|c|}{} &
  \multicolumn{1}{c|}{R2D2} &
  \multicolumn{1}{c|}{90.32} &
  \multicolumn{1}{c|}{89.9} &
  \multicolumn{1}{c|}{96.5} &
  \multicolumn{1}{c|}{99.5} &
  \multicolumn{1}{c|}{71.2} &
  \multicolumn{1}{c|}{86.9} &
  \multicolumn{1}{c|}{97.9} &
  \multicolumn{3}{c|}{} \\ \cline{2-9}
\multicolumn{1}{|c|}{} &
  \multicolumn{1}{c|}{\textbf{Fast-R2D2}} &
  \multicolumn{1}{c|}{\textbf{90.98}} &
  \multicolumn{1}{c|}{89.8} &
  \multicolumn{1}{c|}{96.5} &
  \multicolumn{1}{c|}{99.5} &
  \multicolumn{1}{c|}{74.3} &
  \multicolumn{1}{c|}{87.4} &
  \multicolumn{1}{c|}{98.4} &
  \multicolumn{3}{c|}{} \\ \cline{2-9}
\multicolumn{1}{|c|}{} &
  \multicolumn{1}{c|}{D2-Net} &
  \multicolumn{1}{c|}{89.15} &
  \multicolumn{1}{c|}{85.8} &
  \multicolumn{1}{c|}{94.3} &
  \multicolumn{1}{c|}{98.8} &
  \multicolumn{1}{c|}{70.7} &
  \multicolumn{1}{c|}{86.9} &
  \multicolumn{1}{c|}{98.4} &
  \multicolumn{3}{c|}{} \\ \cline{2-9}
\multicolumn{1}{|c|}{} &
  \multicolumn{1}{c|}{COLMAP SIFT} &
  \multicolumn{1}{c|}{79.17} &
  \multicolumn{1}{c|}{87.7} &
  \multicolumn{1}{c|}{94.2} &
  \multicolumn{1}{c|}{98.4} &
  \multicolumn{1}{c|}{51.8} &
  \multicolumn{1}{c|}{65.4} &
  \multicolumn{1}{c|}{77.5} &
  \multicolumn{3}{c|}{} \\ \cline{2-9}
\multicolumn{1}{|c|}{\multirow{-7}{*}{\begin{tabular}[c]{@{}c@{}}Aachen \\      Day-Night v1.1\end{tabular}}} &
  \multicolumn{1}{c|}{ASLFeat} &
  \multicolumn{1}{c|}{89.97} &
  \multicolumn{1}{c|}{87.5} &
  \multicolumn{1}{c|}{95.4} &
  \multicolumn{1}{c|}{99.3} &
  \multicolumn{1}{c|}{73.8} &
  \multicolumn{1}{c|}{85.9} &
  \multicolumn{1}{c|}{97.9} &
  \multicolumn{3}{c|}{\multirow{-7}{*}{\begin{tabular}[c]{@{}c@{}}Accuracy   thresholds\\      Aachen\\      high: 0.25m, 2°\\      mid: 0.5m, 5°\\      low: 5m, 10°\end{tabular}}} \\ \hline
\multicolumn{12}{l}{} \\ \hline
\multicolumn{1}{|c|}{} &
  \multicolumn{1}{c|}{} &
  \multicolumn{1}{c|}{} &
  \multicolumn{3}{c|}{day} &
  \multicolumn{3}{c|}{night} &
  \multicolumn{3}{c|}{} \\ \cline{4-9}
\multicolumn{1}{|c|}{} &
  \multicolumn{1}{c|}{\multirow{-2}{*}{\begin{tabular}[c]{@{}c@{}}APGeM-20\\      config1 rig seq\end{tabular}}} &
  \multicolumn{1}{c|}{\multirow{-2}{*}{avg. all bins}} &
  \multicolumn{1}{c|}{high} &
  \multicolumn{1}{c|}{mid} &
  \multicolumn{1}{c|}{low} &
  \multicolumn{1}{c|}{high} &
  \multicolumn{1}{c|}{mid} &
  \multicolumn{1}{c|}{low} &
  \multicolumn{3}{c|}{} \\ \cline{2-9}
\multicolumn{1}{|c|}{} &
  \multicolumn{1}{c|}{R2D2} &
  \multicolumn{1}{c|}{79.17} &
  \multicolumn{1}{c|}{65.7} &
  \multicolumn{1}{c|}{95.1} &
  \multicolumn{1}{c|}{100.0} &
  \multicolumn{1}{c|}{43.6} &
  \multicolumn{1}{c|}{76.7} &
  \multicolumn{1}{c|}{93.9} &
  \multicolumn{3}{c|}{} \\ \cline{2-9}
\multicolumn{1}{|c|}{} &
  \multicolumn{1}{c|}{\textbf{Fast-R2D2}} &
  \multicolumn{1}{c|}{\textbf{80.85}} &
  \multicolumn{1}{c|}{65.9} &
  \multicolumn{1}{c|}{95.1} &
  \multicolumn{1}{c|}{100.0} &
  \multicolumn{1}{c|}{46.2} &
  \multicolumn{1}{c|}{81.4} &
  \multicolumn{1}{c|}{96.5} &
  \multicolumn{3}{c|}{} \\ \cline{2-9}
\multicolumn{1}{|c|}{} &
  \multicolumn{1}{c|}{D2-Net} &
  \multicolumn{1}{c|}{75.65} &
  \multicolumn{1}{c|}{60.9} &
  \multicolumn{1}{c|}{93.6} &
  \multicolumn{1}{c|}{99.9} &
  \multicolumn{1}{c|}{35.9} &
  \multicolumn{1}{c|}{72.0} &
  \multicolumn{1}{c|}{91.6} &
  \multicolumn{3}{c|}{} \\ \cline{2-9}
\multicolumn{1}{|c|}{} &
  \multicolumn{1}{c|}{COLMAP SIFT} &
  \multicolumn{1}{c|}{55.97} &
  \multicolumn{1}{c|}{64.2} &
  \multicolumn{1}{c|}{92.5} &
  \multicolumn{1}{c|}{97.9} &
  \multicolumn{1}{c|}{15.9} &
  \multicolumn{1}{c|}{27.3} &
  \multicolumn{1}{c|}{38.0} &
  \multicolumn{3}{c|}{} \\ \cline{2-9}
\multicolumn{1}{|c|}{\multirow{-7}{*}{\begin{tabular}[c]{@{}c@{}}RobotCar\\      Seasons v2\end{tabular}}} &
  \multicolumn{1}{c|}{ASLFeat} &
  \multicolumn{1}{c|}{78.98} &
  \multicolumn{1}{c|}{64.7} &
  \multicolumn{1}{c|}{94.9} &
  \multicolumn{1}{c|}{99.9} &
  \multicolumn{1}{c|}{45.2} &
  \multicolumn{1}{c|}{77.4} &
  \multicolumn{1}{c|}{91.8} &
  \multicolumn{3}{c|}{\multirow{-7}{*}{\begin{tabular}[c]{@{}c@{}}Accuracy   thresholds\\      RobotCar\\      high: 0.25m, 2°\\      mid: 0.5m, 5°\\      low: 5m, 10°\end{tabular}}} \\ \hline
\multicolumn{12}{l}{} \\ \hline
\multicolumn{1}{|c|}{} &
  \multicolumn{1}{c|}{} &
  \multicolumn{1}{c|}{} &
  \multicolumn{1}{c|}{} &
  \multicolumn{1}{c|}{} &
  \multicolumn{1}{c|}{} &
  \multicolumn{1}{c|}{} &
  \multicolumn{1}{c|}{} &
  \multicolumn{1}{c|}{} &
  \multicolumn{3}{c|}{} \\
\multicolumn{1}{|c|}{} &
  \multicolumn{1}{c|}{\multirow{-2}{*}{\begin{tabular}[c]{@{}c@{}}APGeM-20\\      config2\end{tabular}}} &
  \multicolumn{1}{c|}{\multirow{-2}{*}{avg. all scenes}} &
  \multicolumn{1}{c|}{\multirow{-2}{*}{\begin{tabular}[c]{@{}c@{}}Great\\      Court\end{tabular}}} &
  \multicolumn{1}{c|}{\multirow{-2}{*}{\begin{tabular}[c]{@{}c@{}}Kings\\      College\end{tabular}}} &
  \multicolumn{1}{c|}{\multirow{-2}{*}{\begin{tabular}[c]{@{}c@{}}Old\\      Hospital\end{tabular}}} &
  \multicolumn{1}{c|}{\multirow{-2}{*}{\begin{tabular}[c]{@{}c@{}}Shop\\      Facade\end{tabular}}} &
  \multicolumn{1}{c|}{\multirow{-2}{*}{\begin{tabular}[c]{@{}c@{}}StMarys\\      Church\end{tabular}}} &
  \multicolumn{1}{c|}{\multirow{-2}{*}{Street}} &
  \multicolumn{3}{c|}{} \\ \cline{2-9}
\multicolumn{1}{|c|}{} &
  \multicolumn{1}{c|}{R2D2} &
  \multicolumn{1}{c|}{93.98} &
  \multicolumn{1}{c|}{86.2} &
  \multicolumn{1}{c|}{99.7} &
  \multicolumn{1}{c|}{95.1} &
  \multicolumn{1}{c|}{99.0} &
  \multicolumn{1}{c|}{99.4} &
  \multicolumn{1}{c|}{84.5} &
  \multicolumn{3}{c|}{} \\ \cline{2-9}
\multicolumn{1}{|c|}{} &
  \multicolumn{1}{c|}{\textbf{Fast-R2D2}} &
  \multicolumn{1}{c|}{\textbf{94.43}} &
  \multicolumn{1}{c|}{89.2} &
  \multicolumn{1}{c|}{99.4} &
  \multicolumn{1}{c|}{94.5} &
  \multicolumn{1}{c|}{99.0} &
  \multicolumn{1}{c|}{99.6} &
  \multicolumn{1}{c|}{84.9} &
  \multicolumn{3}{c|}{} \\ \cline{2-9}
\multicolumn{1}{|c|}{} &
  \multicolumn{1}{c|}{D2-Net} &
  \multicolumn{1}{c|}{92.30} &
  \multicolumn{1}{c|}{85.5} &
  \multicolumn{1}{c|}{99.1} &
  \multicolumn{1}{c|}{91.2} &
  \multicolumn{1}{c|}{99.0} &
  \multicolumn{1}{c|}{99.6} &
  \multicolumn{1}{c|}{79.4} &
  \multicolumn{3}{c|}{} \\ \cline{2-9}
\multicolumn{1}{|c|}{} &
  \multicolumn{1}{c|}{COLMAP SIFT} &
  \multicolumn{1}{c|}{93.20} &
  \multicolumn{1}{c|}{85.0} &
  \multicolumn{1}{c|}{99.4} &
  \multicolumn{1}{c|}{95.1} &
  \multicolumn{1}{c|}{99.0} &
  \multicolumn{1}{c|}{99.4} &
  \multicolumn{1}{c|}{81.3} &
  \multicolumn{3}{c|}{} \\ \cline{2-9}
\multicolumn{1}{|c|}{\multirow{-7}{*}{Cambridge Landmarks}} &
  \multicolumn{1}{c|}{ASLFeat} &
  \multicolumn{1}{c|}{93.33} &
  \multicolumn{1}{c|}{87.9} &
  \multicolumn{1}{c|}{98.8} &
  \multicolumn{1}{c|}{92.3} &
  \multicolumn{1}{c|}{99.0} &
  \multicolumn{1}{c|}{99.6} &
  \multicolumn{1}{c|}{82.4} &
  \multicolumn{3}{c|}{\multirow{-7}{*}{\begin{tabular}[c]{@{}c@{}}Accuracy   threshold\\      Cambridge Landmarks\\      0.5m, 5°\end{tabular}}} \\ \hline
\end{tabular}

%% file: tables/fastr2d2.tex
\begin{tabular}{|c|c|c|c|c|c|c|c|c|}
\hline
\rowcolor[HTML]{F5E0C7} 
\textbf{Dataset} & \multicolumn{2}{c|}{\cellcolor[HTML]{F5E0C7}\textbf{Extended-CMU}} & \multicolumn{2}{c|}{\cellcolor[HTML]{F5E0C7}\textbf{RobotCar}} & \multicolumn{2}{c|}{\cellcolor[HTML]{F5E0C7}\textbf{InLoc train}} & \multicolumn{2}{c|}{\cellcolor[HTML]{F5E0C7}\textbf{InLoc test}} \\ \hline
Image size       & \multicolumn{2}{c|}{1024x768}                                      & \multicolumn{2}{c|}{1024x1024}                                 & \multicolumn{2}{c|}{1600x1200}                                    & \multicolumn{2}{c|}{4032x3024}                                   \\ \hline \hline
\rowcolor[HTML]{F5E0C7} 
\textbf{GPU}     & \textbf{R2D2}                 & \textbf{Fast-R2D2}                 & \textbf{R2D2}               & \textbf{Fast-R2D2}               & \textbf{R2D2}                 & \textbf{Fast-R2D2}                & \textbf{R2D2}                & \textbf{Fast-R2D2}                \\ \hline
Tesla P40 24GB   & 1.34                          & \textbf{0.46}                      & 1.76                        & \textbf{0.59}                    & 3.28                          & \textbf{1.1}                      & 15.08                        & \textbf{5.43}                     \\ \hline
Tesla V100 32GB  & 0.57                          & \textbf{0.21}                      & 0.75                        & \textbf{0.27}                    & 1.42                          & \textbf{0.5}                      & 6.75                         & \textbf{2.66}                     \\ \hline
Tesla P100 16GB  & 1.2                           & \textbf{0.39}                      & 1.58                        & \textbf{0.5}                     & 2.86                          & \textbf{0.9}                      & out-of-mem                   & \textbf{4.24}                     \\ \hline
\end{tabular}

%% file: tables/gfeat_exps_arxiv.tex
\begin{tabular}{cccccccccccc}
\hline
\rowcolor[HTML]{F5E0C7} 
\multicolumn{1}{|c|}{\cellcolor[HTML]{F5E0C7}\textbf{Datasets}} &
  \multicolumn{1}{c|}{\cellcolor[HTML]{F5E0C7}\textbf{Algorithm}} &
  \multicolumn{10}{c|}{\cellcolor[HTML]{F5E0C7}\textbf{percentage of successfully localized images}} \\ \hline
\multicolumn{1}{|c|}{} &
  \multicolumn{1}{c|}{} &
  \multicolumn{1}{c|}{} &
  \multicolumn{3}{c|}{day} &
  \multicolumn{3}{c|}{night} &
  \multicolumn{3}{c|}{} \\ \cline{4-9}
\multicolumn{1}{|c|}{} &
  \multicolumn{1}{c|}{\multirow{-2}{*}{\begin{tabular}[c]{@{}c@{}}R2D2   top20\\      config2\end{tabular}}} &
  \multicolumn{1}{c|}{\multirow{-2}{*}{avg. all bins}} &
  \multicolumn{1}{c|}{high} &
  \multicolumn{1}{c|}{mid} &
  \multicolumn{1}{c|}{low} &
  \multicolumn{1}{c|}{high} &
  \multicolumn{1}{c|}{mid} &
  \multicolumn{1}{c|}{low} &
  \multicolumn{3}{c|}{} \\ \cline{2-9}
\multicolumn{1}{|c|}{} &
  \multicolumn{1}{c|}{APGeM} &
  \multicolumn{1}{c|}{90.32} &
  \multicolumn{1}{c|}{89.9} &
  \multicolumn{1}{c|}{96.5} &
  \multicolumn{1}{c|}{99.5} &
  \multicolumn{1}{c|}{71.2} &
  \multicolumn{1}{c|}{86.9} &
  \multicolumn{1}{c|}{97.9} &
  \multicolumn{3}{c|}{} \\ \cline{2-9}
\multicolumn{1}{|c|}{} &
  \multicolumn{1}{c|}{DELG} &
  \multicolumn{1}{c|}{90.55} &
  \multicolumn{1}{c|}{90.0} &
  \multicolumn{1}{c|}{96.0} &
  \multicolumn{1}{c|}{99.2} &
  \multicolumn{1}{c|}{73.3} &
  \multicolumn{1}{c|}{87.4} &
  \multicolumn{1}{c|}{97.4} &
  \multicolumn{3}{c|}{} \\ \cline{2-9}
\multicolumn{1}{|c|}{} &
  \multicolumn{1}{c|}{NetVLAD} &
  \multicolumn{1}{c|}{90.60} &
  \multicolumn{1}{c|}{88.7} &
  \multicolumn{1}{c|}{95.1} &
  \multicolumn{1}{c|}{98.1} &
  \multicolumn{1}{c|}{74.3} &
  \multicolumn{1}{c|}{89.5} &
  \multicolumn{1}{c|}{97.9} &
  \multicolumn{3}{c|}{} \\ \cline{2-9}
\multicolumn{1}{|c|}{} &
  \multicolumn{1}{c|}{DenseVLAD} &
  \multicolumn{1}{c|}{84.93} &
  \multicolumn{1}{c|}{88.2} &
  \multicolumn{1}{c|}{94.2} &
  \multicolumn{1}{c|}{97.3} &
  \multicolumn{1}{c|}{62.8} &
  \multicolumn{1}{c|}{79.1} &
  \multicolumn{1}{c|}{88.0} &
  \multicolumn{3}{c|}{} \\ \cline{2-9}
\multicolumn{1}{|c|}{\multirow{-7}{*}{\begin{tabular}[c]{@{}c@{}}Aachen \\      Day-Night v1.1\end{tabular}}} &
  \multicolumn{1}{c|}{\textbf{GHARM}} &
  \multicolumn{1}{c|}{\textbf{91.68}} &
  \multicolumn{1}{c|}{90.5} &
  \multicolumn{1}{c|}{96.8} &
  \multicolumn{1}{c|}{99.4} &
  \multicolumn{1}{c|}{74.9} &
  \multicolumn{1}{c|}{90.1} &
  \multicolumn{1}{c|}{98.4} &
  \multicolumn{3}{c|}{\multirow{-7}{*}{\begin{tabular}[c]{@{}c@{}}Accuracy   thresholds\\      Aachen\\      Extended-CMU\\      high: 0.25m, 2°\\      mid: 0.5m, 5°\\      low: 5m, 10°\end{tabular}}} \\ \hline
\multicolumn{12}{l}{} \\ \hline
\multicolumn{1}{|c|}{} &
  \multicolumn{1}{c|}{} &
  \multicolumn{1}{c|}{} &
  \multicolumn{3}{c|}{urban} &
  \multicolumn{3}{c|}{suburban} &
  \multicolumn{3}{c|}{park} \\ \cline{4-12} 
\multicolumn{1}{|c|}{} &
  \multicolumn{1}{c|}{\multirow{-2}{*}{\begin{tabular}[c]{@{}c@{}}R2D2   top20\\      config1 rig seq\end{tabular}}} &
  \multicolumn{1}{c|}{\multirow{-2}{*}{avg. all bins}} &
  \multicolumn{1}{c|}{high} &
  \multicolumn{1}{c|}{mid} &
  \multicolumn{1}{c|}{low} &
  \multicolumn{1}{c|}{high} &
  \multicolumn{1}{c|}{mid} &
  \multicolumn{1}{c|}{low} &
  \multicolumn{1}{c|}{high} &
  \multicolumn{1}{c|}{mid} &
  \multicolumn{1}{c|}{low} \\ \cline{2-12} 
\multicolumn{1}{|c|}{} &
  \multicolumn{1}{c|}{APGeM} &
  \multicolumn{1}{c|}{94.87} &
  \multicolumn{1}{c|}{96.7} &
  \multicolumn{1}{c|}{98.9} &
  \multicolumn{1}{c|}{99.7} &
  \multicolumn{1}{c|}{94.4} &
  \multicolumn{1}{c|}{96.8} &
  \multicolumn{1}{c|}{99.2} &
  \multicolumn{1}{c|}{83.6} &
  \multicolumn{1}{c|}{89.0} &
  \multicolumn{1}{c|}{95.5} \\ \cline{2-12} 
\multicolumn{1}{|c|}{} &
  \multicolumn{1}{c|}{DELG} &
  \multicolumn{1}{c|}{95.00} &
  \multicolumn{1}{c|}{96.6} &
  \multicolumn{1}{c|}{98.8} &
  \multicolumn{1}{c|}{99.7} &
  \multicolumn{1}{c|}{94.1} &
  \multicolumn{1}{c|}{96.7} &
  \multicolumn{1}{c|}{99.1} &
  \multicolumn{1}{c|}{84.7} &
  \multicolumn{1}{c|}{89.6} &
  \multicolumn{1}{c|}{95.7} \\ \cline{2-12} 
\multicolumn{1}{|c|}{} &
  \multicolumn{1}{c|}{NetVLAD} &
  \multicolumn{1}{c|}{95.94} &
  \multicolumn{1}{c|}{97.1} &
  \multicolumn{1}{c|}{99.1} &
  \multicolumn{1}{c|}{99.8} &
  \multicolumn{1}{c|}{93.8} &
  \multicolumn{1}{c|}{96.3} &
  \multicolumn{1}{c|}{99.1} &
  \multicolumn{1}{c|}{88.1} &
  \multicolumn{1}{c|}{92.7} &
  \multicolumn{1}{c|}{97.5} \\ \cline{2-12} 
\multicolumn{1}{|c|}{} &
  \multicolumn{1}{c|}{DenseVLAD} &
  \multicolumn{1}{c|}{95.66} &
  \multicolumn{1}{c|}{96.1} &
  \multicolumn{1}{c|}{98.4} &
  \multicolumn{1}{c|}{99.4} &
  \multicolumn{1}{c|}{94.2} &
  \multicolumn{1}{c|}{96.7} &
  \multicolumn{1}{c|}{99.1} &
  \multicolumn{1}{c|}{87.9} &
  \multicolumn{1}{c|}{92.3} &
  \multicolumn{1}{c|}{96.8} \\ \cline{2-12} 
\multicolumn{1}{|c|}{\multirow{-7}{*}{\begin{tabular}[c]{@{}c@{}}Extended-\\      CMU\end{tabular}}} &
  \multicolumn{1}{c|}{\textbf{GHARM}} &
  \multicolumn{1}{c|}{\textbf{96.38}} &
  \multicolumn{1}{c|}{97.0} &
  \multicolumn{1}{c|}{99.1} &
  \multicolumn{1}{c|}{99.8} &
  \multicolumn{1}{c|}{95.0} &
  \multicolumn{1}{c|}{97.0} &
  \multicolumn{1}{c|}{99.4} &
  \multicolumn{1}{c|}{89.2} &
  \multicolumn{1}{c|}{93.4} &
  \multicolumn{1}{c|}{97.5} \\ \hline
\end{tabular}

%% file: tables/mapping_pairsfile_exps.tex
\begin{tabular}{ccclllcll}
\hline
\rowcolor[HTML]{F5E0C7} 
\multicolumn{1}{|c|}{\cellcolor[HTML]{F5E0C7}\textbf{Datasets}} & \multicolumn{1}{c|}{\cellcolor[HTML]{F5E0C7}\textbf{Algorithm}} & \multicolumn{7}{c|}{\cellcolor[HTML]{F5E0C7}\textbf{percentage of successfully localized images}} \\ \hline
\multicolumn{1}{|c|}{} & \multicolumn{1}{c|}{} & \multicolumn{1}{c|}{} & \multicolumn{3}{c|}{day} & \multicolumn{3}{c|}{night} \\ \cline{4-9} 
\multicolumn{1}{|c|}{} & \multicolumn{1}{c|}{\multirow{-2}{*}{\begin{tabular}[c]{@{}c@{}}R2D2\\      APGeM-20\end{tabular}}} & \multicolumn{1}{c|}{\multirow{-2}{*}{avg. all   bins}} & \multicolumn{1}{c|}{high} & \multicolumn{1}{c|}{mid} & \multicolumn{1}{c|}{low} & \multicolumn{1}{c|}{high} & \multicolumn{1}{c|}{mid} & \multicolumn{1}{c|}{low} \\ \cline{2-9} 
\multicolumn{1}{|c|}{} & \multicolumn{1}{c|}{AP-GeM} & \multicolumn{1}{c|}{90.33} & \multicolumn{1}{l|}{89.4} & \multicolumn{1}{l|}{96.2} & \multicolumn{1}{l|}{99.4} & \multicolumn{1}{l|}{71.7} & \multicolumn{1}{l|}{86.9} & \multicolumn{1}{l|}{98.4} \\ \cline{2-9} 
\multicolumn{1}{|c|}{} & \multicolumn{1}{c|}{\textbf{distance}} & \multicolumn{1}{c|}{\textbf{90.50}} & \multicolumn{1}{l|}{89.4} & \multicolumn{1}{l|}{96.6} & \multicolumn{1}{l|}{99.4} & \multicolumn{1}{l|}{72.3} & \multicolumn{1}{l|}{87.4} & \multicolumn{1}{l|}{97.9} \\ \cline{2-9} 
\multicolumn{1}{|c|}{\multirow{-5}{*}{\begin{tabular}[c]{@{}c@{}}Aachen \\      Day-Night v1.1\end{tabular}}} & \multicolumn{1}{c|}{frustum} & \multicolumn{1}{c|}{86.28} & \multicolumn{1}{l|}{86.7} & \multicolumn{1}{l|}{84.9} & \multicolumn{1}{l|}{99.0} & \multicolumn{1}{l|}{68.1} & \multicolumn{1}{l|}{83.2} & \multicolumn{1}{l|}{95.8} \\ \hline
\multicolumn{9}{l}{} \\ \hline
\multicolumn{1}{|c|}{} & \multicolumn{1}{c|}{} & \multicolumn{1}{c|}{} & \multicolumn{3}{c|}{all} & \multicolumn{3}{c|}{} \\ \cline{4-6}
\multicolumn{1}{|c|}{} & \multicolumn{1}{c|}{\multirow{-2}{*}{\begin{tabular}[c]{@{}c@{}}R2D2\\      APGeM-20\end{tabular}}} & \multicolumn{1}{c|}{\multirow{-2}{*}{avg. all bins}} & \multicolumn{1}{c|}{high} & \multicolumn{1}{c|}{mid} & \multicolumn{1}{c|}{low} & \multicolumn{3}{c|}{} \\ \cline{2-6}
\multicolumn{1}{|c|}{} & \multicolumn{1}{c|}{AP-GeM} & \multicolumn{1}{c|}{71.30} & \multicolumn{1}{l|}{57.5} & \multicolumn{1}{l|}{73.4} & \multicolumn{1}{l|}{83.0} & \multicolumn{3}{c|}{} \\ \cline{2-6}
\multicolumn{1}{|c|}{} & \multicolumn{1}{c|}{\textbf{distance}} & \multicolumn{1}{c|}{\textbf{71.97}} & \multicolumn{1}{l|}{58.3} & \multicolumn{1}{l|}{74.2} & \multicolumn{1}{l|}{83.4} & \multicolumn{3}{c|}{} \\ \cline{2-6}
\multicolumn{1}{|c|}{\multirow{-5}{*}{Baidu-mall}} & \multicolumn{1}{c|}{frustum} & \multicolumn{1}{c|}{71.63} & \multicolumn{1}{l|}{58.6} & \multicolumn{1}{l|}{73.5} & \multicolumn{1}{l|}{82.8} & \multicolumn{3}{c|}{\multirow{-5}{*}{\begin{tabular}[c]{@{}c@{}}Accuracy   thresholds\\      Aachen, Baidu-mall\\      high: 0.25m, 2°\\      mid: 0.5m, 5°\\      low: 5m, 10°\end{tabular}}} \\ \hline
\end{tabular}

%% file: tables/pipelines_exps_arxiv.tex
\begin{tabular}{ccccccccccll}
\hline
\rowcolor[HTML]{F5E0C7} 
\multicolumn{1}{|c|}{\cellcolor[HTML]{F5E0C7}\textbf{Datasets}} &
  \multicolumn{1}{c|}{\cellcolor[HTML]{F5E0C7}\textbf{Algorithm}} &
  \multicolumn{10}{c|}{\cellcolor[HTML]{F5E0C7}\textbf{percentage of successfully localized images}} \\ \hline
\multicolumn{1}{|c|}{} &
  \multicolumn{1}{c|}{} &
  \multicolumn{1}{c|}{} &
  \multicolumn{3}{c|}{day} &
  \multicolumn{3}{c|}{night} &
  \multicolumn{3}{c|}{} \\ \cline{4-9}
\multicolumn{1}{|c|}{} &
  \multicolumn{1}{c|}{\multirow{-2}{*}{\begin{tabular}[c]{@{}c@{}}R2D2\\      frustum,APGeM-20\end{tabular}}} &
  \multicolumn{1}{c|}{\multirow{-2}{*}{avg. all bins}} &
  \multicolumn{1}{c|}{high} &
  \multicolumn{1}{c|}{mid} &
  \multicolumn{1}{c|}{low} &
  \multicolumn{1}{c|}{high} &
  \multicolumn{1}{c|}{mid} &
  \multicolumn{1}{c|}{low} &
  \multicolumn{3}{c|}{} \\ \cline{2-9}
\multicolumn{1}{|c|}{} &
  \multicolumn{1}{c|}{\textbf{COLMAP config2}} &
  \multicolumn{1}{c|}{\textbf{90.63}} &
  \multicolumn{1}{c|}{89.8} &
  \multicolumn{1}{c|}{96.5} &
  \multicolumn{1}{c|}{99.4} &
  \multicolumn{1}{c|}{73.3} &
  \multicolumn{1}{c|}{86.9} &
  \multicolumn{1}{c|}{97.9} &
  \multicolumn{3}{c|}{} \\ \cline{2-9}
\multicolumn{1}{|c|}{} &
  \multicolumn{1}{c|}{\textbf{pycolmap}} &
  \multicolumn{1}{c|}{\textbf{90.63}} &
  \multicolumn{1}{c|}{89.9} &
  \multicolumn{1}{c|}{96.4} &
  \multicolumn{1}{c|}{99.4} &
  \multicolumn{1}{c|}{73.8} &
  \multicolumn{1}{c|}{86.4} &
  \multicolumn{1}{c|}{97.9} &
  \multicolumn{3}{c|}{} \\ \cline{2-9}
\multicolumn{1}{|c|}{\multirow{-5}{*}{\begin{tabular}[c]{@{}c@{}}Aachen \\      Day-Night v1.1\end{tabular}}} &
  \multicolumn{1}{c|}{ransaclib} &
  \multicolumn{1}{c|}{90.62} &
  \multicolumn{1}{c|}{90.1} &
  \multicolumn{1}{c|}{96.5} &
  \multicolumn{1}{c|}{99.5} &
  \multicolumn{1}{c|}{73.3} &
  \multicolumn{1}{c|}{85.9} &
  \multicolumn{1}{c|}{98.4} &
  \multicolumn{3}{c|}{\multirow{-5}{*}{\begin{tabular}[c]{@{}c@{}}Accuracy   thresholds\\      Aachen\\      high: 0.25m, 2°\\      mid: 0.5m, 5°\\      low: 5m, 10°\end{tabular}}} \\ \hline
\multicolumn{12}{l}{} \\ \hline
\multicolumn{1}{|c|}{} &
  \multicolumn{1}{c|}{} &
  \multicolumn{1}{c|}{} &
  \multicolumn{3}{c|}{test} &
  \multicolumn{3}{c|}{validation} &
  \multicolumn{3}{c|}{} \\ \cline{4-9}
\multicolumn{1}{|c|}{} &
  \multicolumn{1}{c|}{\multirow{-2}{*}{\begin{tabular}[c]{@{}c@{}}R2D2\\      distance-50,APGeM-20\end{tabular}}} &
  \multicolumn{1}{c|}{\multirow{-2}{*}{avg. all bins}} &
  \multicolumn{1}{c|}{high} &
  \multicolumn{1}{c|}{mid} &
  \multicolumn{1}{c|}{low} &
  \multicolumn{1}{c|}{high} &
  \multicolumn{1}{c|}{mid} &
  \multicolumn{1}{c|}{low} &
  \multicolumn{3}{c|}{} \\ \cline{2-9}
\multicolumn{1}{|c|}{} &
  \multicolumn{1}{c|}{COLMAP config2} &
  \multicolumn{1}{c|}{53.35} &
  \multicolumn{1}{c|}{42.6} &
  \multicolumn{1}{c|}{60.4} &
  \multicolumn{1}{c|}{64.8} &
  \multicolumn{1}{c|}{35.8} &
  \multicolumn{1}{c|}{56.4} &
  \multicolumn{1}{c|}{60.1} &
  \multicolumn{3}{c|}{} \\ \cline{2-9}
\multicolumn{1}{|c|}{} &
  \multicolumn{1}{c|}{pycolmap} &
  \multicolumn{1}{c|}{55.97} &
  \multicolumn{1}{c|}{44.3} &
  \multicolumn{1}{c|}{62.5} &
  \multicolumn{1}{c|}{66.9} &
  \multicolumn{1}{c|}{38.1} &
  \multicolumn{1}{c|}{59.6} &
  \multicolumn{1}{c|}{64.4} &
  \multicolumn{3}{c|}{} \\ \cline{2-9}
\multicolumn{1}{|c|}{\multirow{-5}{*}{Gangnam Station B2}} &
  \multicolumn{1}{c|}{\textbf{ransaclib}} &
  \multicolumn{1}{c|}{\textbf{56.12}} &
  \multicolumn{1}{c|}{44.5} &
  \multicolumn{1}{c|}{62.4} &
  \multicolumn{1}{c|}{67.0} &
  \multicolumn{1}{c|}{38.7} &
  \multicolumn{1}{c|}{59.9} &
  \multicolumn{1}{c|}{64.2} &
  \multicolumn{3}{c|}{\multirow{-5}{*}{\begin{tabular}[c]{@{}c@{}}Accuracy   thresholds\\      Gangnam B2\\      high: 0.1m, 1°\\      mid: 0.25m, 2°\\      low: 1m, 5°\end{tabular}}} \\ \hline
\multicolumn{12}{l}{} \\ \hline
\multicolumn{1}{|c|}{} &
  \multicolumn{1}{c|}{} &
  \multicolumn{1}{c|}{} &
  \multicolumn{3}{c|}{all} &
  \multicolumn{6}{c|}{} \\ \cline{4-6}
\multicolumn{1}{|c|}{} &
  \multicolumn{1}{c|}{\multirow{-2}{*}{\begin{tabular}[c]{@{}c@{}}R2D2\\      frustum,APGeM-20\end{tabular}}} &
  \multicolumn{1}{c|}{\multirow{-2}{*}{avg. all bins}} &
  \multicolumn{1}{c|}{high} &
  \multicolumn{1}{c|}{mid} &
  \multicolumn{1}{c|}{low} &
  \multicolumn{6}{c|}{} \\ \cline{2-6}
\multicolumn{1}{|c|}{} &
  \multicolumn{1}{c|}{COLMAP config2} &
  \multicolumn{1}{c|}{71.77} &
  \multicolumn{1}{c|}{58.6} &
  \multicolumn{1}{c|}{73.7} &
  \multicolumn{1}{c|}{83.0} &
  \multicolumn{6}{c|}{} \\ \cline{2-6}
\multicolumn{1}{|c|}{} &
  \multicolumn{1}{c|}{pycolmap} &
  \multicolumn{1}{c|}{72.97} &
  \multicolumn{1}{c|}{60.0} &
  \multicolumn{1}{c|}{75.0} &
  \multicolumn{1}{c|}{83.9} &
  \multicolumn{6}{c|}{} \\ \cline{2-6}
\multicolumn{1}{|c|}{\multirow{-5}{*}{Baidu-mall}} &
  \multicolumn{1}{c|}{\textbf{ransaclib}} &
  \multicolumn{1}{c|}{\textbf{73.13}} &
  \multicolumn{1}{c|}{59.9} &
  \multicolumn{1}{c|}{75.6} &
  \multicolumn{1}{c|}{83.9} &
  \multicolumn{6}{c|}{\multirow{-5}{*}{\begin{tabular}[c]{@{}c@{}}Accuracy thresholds\\      Baidu-mall\\      high: 0.25m, 2°\\      mid: 0.5m, 5°\\      low: 5m, 10°\end{tabular}}} \\ \hline
\end{tabular}

%% file: tables/rig_arxiv.tex
\begin{tabular}{cccccccccccc}
\hline
\rowcolor[HTML]{F5E0C7} 
\multicolumn{1}{|c|}{\cellcolor[HTML]{F5E0C7}\textbf{Datasets}} &
  \multicolumn{1}{c|}{\cellcolor[HTML]{F5E0C7}\textbf{Algorithm}} &
  \multicolumn{10}{c|}{\cellcolor[HTML]{F5E0C7}\textbf{Rig   and sequence post-processing, percentage of localized images}} \\ \hline
\multicolumn{1}{|c|}{} &
  \multicolumn{1}{c|}{} &
  \multicolumn{1}{c|}{} &
  \multicolumn{3}{c|}{day} &
  \multicolumn{3}{c|}{night} &
  \multicolumn{3}{c|}{} \\ \cline{4-9}
\multicolumn{1}{|c|}{} &
  \multicolumn{1}{c|}{\multirow{-2}{*}{\begin{tabular}[c]{@{}c@{}}R2D2   APGeM top20\\      config1\end{tabular}}} &
  \multicolumn{1}{c|}{\multirow{-2}{*}{avg. all   bins}} &
  \multicolumn{1}{c|}{high} &
  \multicolumn{1}{c|}{mid} &
  \multicolumn{1}{c|}{low} &
  \multicolumn{1}{c|}{high} &
  \multicolumn{1}{c|}{mid} &
  \multicolumn{1}{c|}{low} &
  \multicolumn{3}{c|}{} \\ \cline{2-9}
\multicolumn{1}{|c|}{} &
  \multicolumn{1}{c|}{rig + seq} &
  \multicolumn{1}{c|}{79.17} &
  \multicolumn{1}{c|}{65.7} &
  \multicolumn{1}{c|}{95.1} &
  \multicolumn{1}{c|}{100} &
  \multicolumn{1}{c|}{43.6} &
  \multicolumn{1}{c|}{76.7} &
  \multicolumn{1}{c|}{93.9} &
  \multicolumn{3}{c|}{} \\ \cline{2-9}
\multicolumn{1}{|c|}{} &
  \multicolumn{1}{c|}{rig} &
  \multicolumn{1}{c|}{77.85} &
  \multicolumn{1}{c|}{65.7} &
  \multicolumn{1}{c|}{95.1} &
  \multicolumn{1}{c|}{100} &
  \multicolumn{1}{c|}{43.6} &
  \multicolumn{1}{c|}{76} &
  \multicolumn{1}{c|}{86.7} &
  \multicolumn{3}{c|}{} \\ \cline{2-9}
\multicolumn{1}{|c|}{} &
  \multicolumn{1}{c|}{none} &
  \multicolumn{1}{c|}{76.25} &
  \multicolumn{1}{c|}{65.7} &
  \multicolumn{1}{c|}{95.1} &
  \multicolumn{1}{c|}{99.9} &
  \multicolumn{1}{c|}{41.5} &
  \multicolumn{1}{c|}{73} &
  \multicolumn{1}{c|}{82.3} &
  \multicolumn{3}{c|}{} \\ \cline{2-9}
\multicolumn{1}{|c|}{} &
  \multicolumn{1}{c|}{\textbf{pycolmap rig}} &
  \multicolumn{1}{c|}{\textbf{79.70}} &
  \multicolumn{1}{c|}{61.8} &
  \multicolumn{1}{c|}{94.6} &
  \multicolumn{1}{c|}{100} &
  \multicolumn{1}{c|}{48.0} &
  \multicolumn{1}{c|}{81.6} &
  \multicolumn{1}{c|}{92.2} &
  \multicolumn{3}{c|}{} \\ \cline{2-9}
\multicolumn{1}{|c|}{\multirow{-7}{*}{\begin{tabular}[c]{@{}c@{}}RobotCar\\      Seasons v2\end{tabular}}} &
  \multicolumn{1}{c|}{multicamerapose} &
  \multicolumn{1}{c|}{79.28} &
  \multicolumn{1}{c|}{61.1} &
  \multicolumn{1}{c|}{94.5} &
  \multicolumn{1}{c|}{100} &
  \multicolumn{1}{c|}{49.0} &
  \multicolumn{1}{c|}{80.7} &
  \multicolumn{1}{c|}{90.4} &
  \multicolumn{3}{c|}{\multirow{-7}{*}{\begin{tabular}[c]{@{}c@{}}Accuracy   thresholds\\      high: 0.25m, 2°\\      mid: 0.5m, 5°\\      low: 5m, 10°\end{tabular}}} \\ \hline
\multicolumn{12}{c}{} \\ \hline
\multicolumn{1}{|c|}{} &
  \multicolumn{1}{c|}{} &
  \multicolumn{1}{c|}{} &
  \multicolumn{3}{c|}{urban} &
  \multicolumn{3}{c|}{suburban} &
  \multicolumn{3}{c|}{park} \\ \cline{4-12} 
\multicolumn{1}{|c|}{} &
  \multicolumn{1}{c|}{\multirow{-2}{*}{\begin{tabular}[c]{@{}c@{}}R2D2   APGeM top20\\      config1\end{tabular}}} &
  \multicolumn{1}{c|}{\multirow{-2}{*}{avg. all   bins}} &
  \multicolumn{1}{c|}{high} &
  \multicolumn{1}{c|}{mid} &
  \multicolumn{1}{c|}{low} &
  \multicolumn{1}{c|}{high} &
  \multicolumn{1}{c|}{mid} &
  \multicolumn{1}{c|}{low} &
  \multicolumn{1}{c|}{high} &
  \multicolumn{1}{c|}{mid} &
  \multicolumn{1}{c|}{low} \\ \cline{2-12} 
\multicolumn{1}{|c|}{} &
  \multicolumn{1}{c|}{\textbf{rig + seq}} &
  \multicolumn{1}{c|}{\textbf{94.87}} &
  \multicolumn{1}{c|}{96.7} &
  \multicolumn{1}{c|}{98.9} &
  \multicolumn{1}{c|}{99.7} &
  \multicolumn{1}{c|}{94.4} &
  \multicolumn{1}{c|}{96.8} &
  \multicolumn{1}{c|}{99.2} &
  \multicolumn{1}{c|}{83.6} &
  \multicolumn{1}{c|}{89} &
  \multicolumn{1}{c|}{95.5} \\ \cline{2-12} 
\multicolumn{1}{|c|}{} &
  \multicolumn{1}{c|}{rig} &
  \multicolumn{1}{c|}{94.30} &
  \multicolumn{1}{c|}{96.5} &
  \multicolumn{1}{c|}{98.8} &
  \multicolumn{1}{c|}{99.5} &
  \multicolumn{1}{c|}{94.3} &
  \multicolumn{1}{c|}{96.7} &
  \multicolumn{1}{c|}{99.1} &
  \multicolumn{1}{c|}{83.1} &
  \multicolumn{1}{c|}{87.9} &
  \multicolumn{1}{c|}{92.8} \\ \cline{2-12} 
\multicolumn{1}{|c|}{\multirow{-5}{*}{\begin{tabular}[c]{@{}c@{}}Extended-\\      CMU\end{tabular}}} &
  \multicolumn{1}{c|}{none} &
  \multicolumn{1}{c|}{89.11} &
  \multicolumn{1}{c|}{95.8} &
  \multicolumn{1}{c|}{98.1} &
  \multicolumn{1}{c|}{98.8} &
  \multicolumn{1}{c|}{88.9} &
  \multicolumn{1}{c|}{91.1} &
  \multicolumn{1}{c|}{93.4} &
  \multicolumn{1}{c|}{75.5} &
  \multicolumn{1}{c|}{78.4} &
  \multicolumn{1}{c|}{82} \\ \hline
\multicolumn{12}{c}{} \\ \hline
\multicolumn{1}{|c|}{} &
  \multicolumn{1}{c|}{} &
  \multicolumn{1}{c|}{} &
  \multicolumn{3}{c|}{evening} &
  \multicolumn{3}{c|}{snow} &
  \multicolumn{3}{c|}{night} \\ \cline{4-12} 
\multicolumn{1}{|c|}{} &
  \multicolumn{1}{c|}{\multirow{-2}{*}{\begin{tabular}[c]{@{}c@{}}R2D2   APGeM top20\\      config1\end{tabular}}} &
  \multicolumn{1}{c|}{\multirow{-2}{*}{avg. all   bins}} &
  \multicolumn{1}{c|}{high} &
  \multicolumn{1}{c|}{mid} &
  \multicolumn{1}{c|}{low} &
  \multicolumn{1}{c|}{high} &
  \multicolumn{1}{c|}{mid} &
  \multicolumn{1}{c|}{low} &
  \multicolumn{1}{c|}{high} &
  \multicolumn{1}{c|}{mid} &
  \multicolumn{1}{c|}{low} \\ \cline{2-12} 
\multicolumn{1}{|c|}{} &
  \multicolumn{1}{c|}{rig + seq} &
  \multicolumn{1}{c|}{49.97} &
  \multicolumn{1}{c|}{31.9} &
  \multicolumn{1}{c|}{66.6} &
  \multicolumn{1}{c|}{92.5} &
  \multicolumn{1}{c|}{0.5} &
  \multicolumn{1}{c|}{5.8} &
  \multicolumn{1}{c|}{89.2} &
  \multicolumn{1}{c|}{30.5} &
  \multicolumn{1}{c|}{54.2} &
  \multicolumn{1}{c|}{78.5} \\ \cline{2-12} 
\multicolumn{1}{|c|}{} &
  \multicolumn{1}{c|}{rig} &
  \multicolumn{1}{c|}{49.97} &
  \multicolumn{1}{c|}{31.9} &
  \multicolumn{1}{c|}{66.6} &
  \multicolumn{1}{c|}{92.5} &
  \multicolumn{1}{c|}{0.5} &
  \multicolumn{1}{c|}{5.8} &
  \multicolumn{1}{c|}{89.2} &
  \multicolumn{1}{c|}{30.5} &
  \multicolumn{1}{c|}{54.2} &
  \multicolumn{1}{c|}{78.5} \\ \cline{2-12} 
\multicolumn{1}{|c|}{} &
  \multicolumn{1}{c|}{none} &
  \multicolumn{1}{c|}{46.39} &
  \multicolumn{1}{c|}{31.8} &
  \multicolumn{1}{c|}{66.3} &
  \multicolumn{1}{c|}{89.4} &
  \multicolumn{1}{c|}{0.3} &
  \multicolumn{1}{c|}{3.9} &
  \multicolumn{1}{c|}{64.9} &
  \multicolumn{1}{c|}{30} &
  \multicolumn{1}{c|}{53.4} &
  \multicolumn{1}{c|}{77.5} \\ \cline{2-12} 
\multicolumn{1}{|c|}{\multirow{-6}{*}{\begin{tabular}[c]{@{}c@{}}SILDa Weather\\       and\\      Time of Day\end{tabular}}} &
  \multicolumn{1}{c|}{\textbf{pycolmap rig}} &
  \multicolumn{1}{c|}{\textbf{53.17}} &
  \multicolumn{1}{c|}{37.0} &
  \multicolumn{1}{c|}{71.0} &
  \multicolumn{1}{c|}{93.4} &
  \multicolumn{1}{c|}{2.4} &
  \multicolumn{1}{c|}{16.4} &
  \multicolumn{1}{c|}{91.1} &
  \multicolumn{1}{c|}{32.9} &
  \multicolumn{1}{c|}{54.6} &
  \multicolumn{1}{c|}{79.7} \\ \hline
\end{tabular}

%% file: tables/rgbd_exps.tex
\begin{tabular}{|c|c|c|c|c|c|c|c|c|}
\hline
\cellcolor[HTML]{F5E0C7} & \multicolumn{8}{c|}{\cellcolor[HTML]{F5E0C7}\textbf{7-scenes,   percentage of successfully localized images (0.05m, 5.0°)}} \\ \cline{2-9} 
\multirow{-2}{*}{\cellcolor[HTML]{F5E0C7}\textbf{Method}} & avg. all scenes & chess & fire & heads & office & pumpkin & stairs & redkitchen \\ \hline
SFM & 66.11 & 89.6 & 83.6 & 92.4 & 67.9 & 44.5 & 36.8 & 48.0 \\ \hline
\textbf{RGBD} & \textbf{82.47} & 96.6 & 94.1 & 98.1 & 85.7 & 62.8 & 66.2 & 73.8 \\ \hline
\end{tabular}

%% file: tables/summary3.tex
\begin{tabular}{cccccccccccc}
\hline
\rowcolor[HTML]{F5E0C7} 
\multicolumn{1}{|c|}{\cellcolor[HTML]{F5E0C7}\textbf{Datasets}} & \multicolumn{1}{c|}{\cellcolor[HTML]{F5E0C7}\textbf{Algorithm}} & \multicolumn{10}{c|}{\cellcolor[HTML]{F5E0C7}\textbf{ours   compared to top state-of-the-art methods}} \\ \hline
\multicolumn{1}{|c|}{} & \multicolumn{1}{c|}{} & \multicolumn{1}{c|}{} & \multicolumn{3}{c|}{day} & \multicolumn{3}{c|}{night} & \multicolumn{3}{c|}{} \\ \cline{4-9}
\multicolumn{1}{|c|}{} & \multicolumn{1}{c|}{\multirow{-2}{*}{}} & \multicolumn{1}{c|}{\multirow{-2}{*}{avg. all   bins}} & \multicolumn{1}{c|}{high} & \multicolumn{1}{c|}{mid} & \multicolumn{1}{c|}{low} & \multicolumn{1}{c|}{high} & \multicolumn{1}{c|}{mid} & \multicolumn{1}{c|}{low} & \multicolumn{3}{c|}{} \\ \cline{2-9}
\multicolumn{1}{|c|}{} & \multicolumn{1}{c|}{\textbf{ours (R2D2 40k,GHARM-50)}} & \multicolumn{1}{c|}{\textbf{92.45}} & \multicolumn{1}{c|}{90.9} & \multicolumn{1}{c|}{96.7} & \multicolumn{1}{c|}{99.5} & \multicolumn{1}{c|}{78.5} & \multicolumn{1}{c|}{91.1} & \multicolumn{1}{c|}{98.0} & \multicolumn{3}{c|}{} \\ \cline{2-9}
\multicolumn{1}{|c|}{} & \multicolumn{1}{c|}{hloc+SuperGlue~\cite{SarlinCVPR20SuperGlueLearningFeatureMatchingGNN}} & \multicolumn{1}{c|}{92.15} & \multicolumn{1}{c|}{89.8} & \multicolumn{1}{c|}{96.1} & \multicolumn{1}{c|}{99.4} & \multicolumn{1}{c|}{77.0} & \multicolumn{1}{c|}{90.6} & \multicolumn{1}{c|}{100.0} & \multicolumn{3}{c|}{} \\ \cline{2-9}
\multicolumn{1}{|c|}{} & \multicolumn{1}{c|}{ours (Fast-R2D2,GHARM-20)} & \multicolumn{1}{c|}{91.75} & \multicolumn{1}{c|}{90.3} & \multicolumn{1}{c|}{96.6} & \multicolumn{1}{c|}{99.6} & \multicolumn{1}{c|}{74.9} & \multicolumn{1}{c|}{90.1} & \multicolumn{1}{c|}{99.0} & \multicolumn{3}{c|}{} \\ \cline{2-9}
\multicolumn{1}{|c|}{\multirow{-6}{*}{\begin{tabular}[c]{@{}c@{}}Aachen \\      Day-Night v1.1\end{tabular}}} & \multicolumn{1}{c|}{RLOCS\_v1.0~\cite{fan2020visual}} & \multicolumn{1}{c|}{89.90} & \multicolumn{1}{c|}{86.0} & \multicolumn{1}{c|}{94.8} & \multicolumn{1}{c|}{98.8} & \multicolumn{1}{c|}{72.3} & \multicolumn{1}{c|}{88.5} & \multicolumn{1}{c|}{99.0} & \multicolumn{3}{c|}{\multirow{-6}{*}{\begin{tabular}[c]{@{}c@{}}Accuracy   thresholds Aachen\\      high: 0.25m, 2°\\      mid: 0.5m, 5°\\      low: 5m, 10°\end{tabular}}} \\ \hline
\multicolumn{12}{c}{} \\ \hline
\multicolumn{1}{|c|}{} & \multicolumn{1}{c|}{} & \multicolumn{1}{c|}{} & \multicolumn{3}{c|}{DUC1} & \multicolumn{3}{c|}{DUC2} & \multicolumn{3}{c|}{} \\ \cline{4-9}
\multicolumn{1}{|c|}{} & \multicolumn{1}{c|}{\multirow{-2}{*}{}} & \multicolumn{1}{c|}{\multirow{-2}{*}{avg all   bins}} & \multicolumn{1}{c|}{high} & \multicolumn{1}{c|}{mid} & \multicolumn{1}{c|}{low} & \multicolumn{1}{c|}{high} & \multicolumn{1}{c|}{mid} & \multicolumn{1}{c|}{low} & \multicolumn{3}{c|}{} \\ \cline{2-9}
\multicolumn{1}{|c|}{} & \multicolumn{1}{c|}{\textbf{RLOCS\_v1.0~\cite{fan2020visual}}} & \multicolumn{1}{c|}{\textbf{70.10}} & \multicolumn{1}{c|}{47.0} & \multicolumn{1}{c|}{71.2} & \multicolumn{1}{c|}{84.8} & \multicolumn{1}{c|}{58.8} & \multicolumn{1}{c|}{77.9} & \multicolumn{1}{c|}{80.9} & \multicolumn{3}{c|}{} \\ \cline{2-9}
\multicolumn{1}{|c|}{} & \multicolumn{1}{c|}{hloc+SuperGlue~\cite{SarlinCVPR20SuperGlueLearningFeatureMatchingGNN}} & \multicolumn{1}{c|}{68.57} & \multicolumn{1}{c|}{49.0} & \multicolumn{1}{c|}{68.7} & \multicolumn{1}{c|}{80.8} & \multicolumn{1}{c|}{53.4} & \multicolumn{1}{c|}{77.1} & \multicolumn{1}{c|}{82.4} & \multicolumn{3}{c|}{} \\ \cline{2-9}
\multicolumn{1}{|c|}{} & \multicolumn{1}{c|}{ours (Fast-R2D2,RGBD, GHARM-50)} & \multicolumn{1}{c|}{59.55} & \multicolumn{1}{c|}{40.4} & \multicolumn{1}{c|}{63.1} & \multicolumn{1}{c|}{78.3} & \multicolumn{1}{c|}{42.7} & \multicolumn{1}{c|}{64.9} & \multicolumn{1}{c|}{67.9} & \multicolumn{3}{c|}{} \\ \cline{2-9}
\multicolumn{1}{|c|}{} & \multicolumn{1}{c|}{DensePV+S (w/ scan-graph)~\cite{taira2018rightplace}} & \multicolumn{1}{c|}{58.62} & \multicolumn{1}{c|}{40.9} & \multicolumn{1}{c|}{63.6} & \multicolumn{1}{c|}{71.7} & \multicolumn{1}{c|}{42.7} & \multicolumn{1}{c|}{61.8} & \multicolumn{1}{c|}{71.0} & \multicolumn{3}{c|}{} \\ \cline{2-9}
\multicolumn{1}{|c|}{\multirow{-7}{*}{InLoc}} & \multicolumn{1}{c|}{InLoc~\cite{Taira2019TPAMI}} & \multicolumn{1}{c|}{54.80} & \multicolumn{1}{c|}{40.9} & \multicolumn{1}{c|}{58.1} & \multicolumn{1}{c|}{70.2} & \multicolumn{1}{c|}{35.9} & \multicolumn{1}{c|}{54.2} & \multicolumn{1}{c|}{69.5} & \multicolumn{3}{c|}{\multirow{-7}{*}{\begin{tabular}[c]{@{}c@{}}Accuracy   thresholds InLoc\\      high: 0.25m, 10°\\      mid: 0.5m, 10°\\      low: 1m, 10°\end{tabular}}} \\ \hline
\multicolumn{12}{c}{} \\ \hline
\multicolumn{1}{|c|}{} & \multicolumn{1}{c|}{} & \multicolumn{1}{c|}{} & \multicolumn{3}{c|}{day} & \multicolumn{3}{c|}{night} & \multicolumn{3}{c|}{} \\ \cline{4-9}
\multicolumn{1}{|c|}{} & \multicolumn{1}{c|}{\multirow{-2}{*}{}} & \multicolumn{1}{c|}{\multirow{-2}{*}{avg all   bins}} & \multicolumn{1}{c|}{high} & \multicolumn{1}{c|}{mid} & \multicolumn{1}{c|}{low} & \multicolumn{1}{c|}{high} & \multicolumn{1}{c|}{mid} & \multicolumn{1}{c|}{low} & \multicolumn{3}{c|}{} \\ \cline{2-9}
\multicolumn{1}{|c|}{} & \multicolumn{1}{c|}{\textbf{ours (Fast-R2D2,APGeM-20,rig+seq)}} & \multicolumn{1}{c|}{\textbf{80.85}} & \multicolumn{1}{c|}{65.9} & \multicolumn{1}{c|}{95.1} & \multicolumn{1}{c|}{100.0} & \multicolumn{1}{c|}{46.2} & \multicolumn{1}{c|}{81.4} & \multicolumn{1}{c|}{96.5} & \multicolumn{3}{c|}{} \\ \cline{2-9}
\multicolumn{1}{|c|}{} & \multicolumn{1}{c|}{hloc+SuperGlue~\cite{SarlinCVPR20SuperGlueLearningFeatureMatchingGNN}} & \multicolumn{1}{c|}{80.77} & \multicolumn{1}{c|}{63.8} & \multicolumn{1}{c|}{95.0} & \multicolumn{1}{c|}{100.0} & \multicolumn{1}{c|}{45.0} & \multicolumn{1}{c|}{86.2} & \multicolumn{1}{c|}{94.6} & \multicolumn{3}{c|}{} \\ \cline{2-9}
\multicolumn{1}{|c|}{\multirow{-5}{*}{\begin{tabular}[c]{@{}c@{}}RobotCar\\      Seasons v2\end{tabular}}} & \multicolumn{1}{c|}{ours (Fast-R2D2,GHARM-20,rig+seq)} & \multicolumn{1}{c|}{80.17} & \multicolumn{1}{c|}{65.9} & \multicolumn{1}{c|}{95.1} & \multicolumn{1}{c|}{100.0} & \multicolumn{1}{c|}{45.2} & \multicolumn{1}{c|}{80.4} & \multicolumn{1}{c|}{94.4} & \multicolumn{3}{c|}{\multirow{-5}{*}{\begin{tabular}[c]{@{}c@{}}Accuracy   thresholds others\\      high: 0.25m, 2°\\      mid: 0.5m, 5°\\      low: 5m, 10°\end{tabular}}} \\ \hline
\multicolumn{12}{c}{} \\ \hline
\multicolumn{1}{|c|}{} & \multicolumn{1}{c|}{} & \multicolumn{1}{c|}{} & \multicolumn{3}{c|}{urban} & \multicolumn{3}{c|}{suburban} & \multicolumn{3}{c|}{park} \\ \cline{4-12} 
\multicolumn{1}{|c|}{} & \multicolumn{1}{c|}{\multirow{-2}{*}{}} & \multicolumn{1}{c|}{\multirow{-2}{*}{avg all   bins}} & \multicolumn{1}{c|}{high} & \multicolumn{1}{c|}{mid} & \multicolumn{1}{c|}{low} & \multicolumn{1}{c|}{high} & \multicolumn{1}{c|}{mid} & \multicolumn{1}{c|}{low} & \multicolumn{1}{c|}{high} & \multicolumn{1}{c|}{mid} & \multicolumn{1}{c|}{low} \\ \cline{2-12} 
\multicolumn{1}{|c|}{} & \multicolumn{1}{c|}{\textbf{FGSN~\cite{larsson2019fine}}} & \multicolumn{1}{c|}{\textbf{98.98}} & \multicolumn{1}{c|}{94.1} & \multicolumn{1}{c|}{99.3} & \multicolumn{1}{c|}{100.0} & \multicolumn{1}{c|}{100.0} & \multicolumn{1}{c|}{100.0} & \multicolumn{1}{c|}{100.0} & \multicolumn{1}{c|}{97.6} & \multicolumn{1}{c|}{99.9} & \multicolumn{1}{c|}{99.9} \\ \cline{2-12} 
\multicolumn{1}{|c|}{} & \multicolumn{1}{c|}{hloc+SuperGlue~\cite{SarlinCVPR20SuperGlueLearningFeatureMatchingGNN}} & \multicolumn{1}{c|}{98.38} & \multicolumn{1}{c|}{98.1} & \multicolumn{1}{c|}{99.8} & \multicolumn{1}{c|}{99.9} & \multicolumn{1}{c|}{98.3} & \multicolumn{1}{c|}{99.5} & \multicolumn{1}{c|}{100.0} & \multicolumn{1}{c|}{94.2} & \multicolumn{1}{c|}{97.1} & \multicolumn{1}{c|}{98.5} \\ \cline{2-12} 
\multicolumn{1}{|c|}{} & \multicolumn{1}{c|}{ours (R2D2,GHARM-20,rig+seq)} & \multicolumn{1}{c|}{96.87} & \multicolumn{1}{c|}{97.3} & \multicolumn{1}{c|}{99.4} & \multicolumn{1}{c|}{99.8} & \multicolumn{1}{c|}{95.6} & \multicolumn{1}{c|}{97.2} & \multicolumn{1}{c|}{99.4} & \multicolumn{1}{c|}{90.8} & \multicolumn{1}{c|}{94.6} & \multicolumn{1}{c|}{97.7} \\ \cline{2-12} 
\multicolumn{1}{|c|}{\multirow{-6}{*}{\begin{tabular}[c]{@{}c@{}}Extended-\\      CMU\end{tabular}}} & \multicolumn{1}{c|}{Active Search v1.1~\cite{SattlerPAMI17EfficientPrioritizedMatching}} & \multicolumn{1}{c|}{70.48} & \multicolumn{1}{c|}{81.0} & \multicolumn{1}{c|}{87.3} & \multicolumn{1}{c|}{92.4} & \multicolumn{1}{c|}{62.6} & \multicolumn{1}{c|}{70.9} & \multicolumn{1}{c|}{81.0} & \multicolumn{1}{c|}{45.5} & \multicolumn{1}{c|}{51.6} & \multicolumn{1}{c|}{62.0} \\ \hline
\multicolumn{12}{c}{} \\ \hline
\multicolumn{1}{|c|}{} & \multicolumn{1}{c|}{} & \multicolumn{1}{c|}{} & \multicolumn{3}{c|}{evening} & \multicolumn{3}{c|}{snow} & \multicolumn{3}{c|}{night} \\ \cline{4-12} 
\multicolumn{1}{|c|}{} & \multicolumn{1}{c|}{\multirow{-2}{*}{}} & \multicolumn{1}{c|}{\multirow{-2}{*}{avg all   bins}} & \multicolumn{1}{c|}{high} & \multicolumn{1}{c|}{mid} & \multicolumn{1}{c|}{low} & \multicolumn{1}{c|}{high} & \multicolumn{1}{c|}{mid} & \multicolumn{1}{c|}{low} & \multicolumn{1}{c|}{high} & \multicolumn{1}{c|}{mid} & \multicolumn{1}{c|}{low} \\ \cline{2-12} 
\multicolumn{1}{|c|}{} & \multicolumn{1}{c|}{\textbf{hloc+SuperGlue~\cite{SarlinCVPR20SuperGlueLearningFeatureMatchingGNN}}} & \multicolumn{1}{c|}{\textbf{51.59}} & \multicolumn{1}{c|}{35.5} & \multicolumn{1}{c|}{75.0} & \multicolumn{1}{c|}{97.1} & \multicolumn{1}{c|}{0.0} & \multicolumn{1}{c|}{2.4} & \multicolumn{1}{c|}{86.3} & \multicolumn{1}{c|}{31.7} & \multicolumn{1}{c|}{54.4} & \multicolumn{1}{c|}{81.9} \\ \cline{2-12} 
\multicolumn{1}{|c|}{} & \multicolumn{1}{c|}{ours (R2D2,GHARM-20,rig+seq)} & \multicolumn{1}{c|}{50.22} & \multicolumn{1}{c|}{32.4} & \multicolumn{1}{c|}{67.4} & \multicolumn{1}{c|}{93.3} & \multicolumn{1}{c|}{0.2} & \multicolumn{1}{c|}{4.1} & \multicolumn{1}{c|}{88.9} & \multicolumn{1}{c|}{30.4} & \multicolumn{1}{c|}{54.2} & \multicolumn{1}{c|}{81.1} \\ \cline{2-12} 
\multicolumn{1}{|c|}{} & \multicolumn{1}{c|}{NetVLAD (top50) D2-Net~\cite{DusmanuCVPR19D2NetDeepLocalFeatures}} & \multicolumn{1}{c|}{49.33} & \multicolumn{1}{c|}{29.6} & \multicolumn{1}{c|}{67.8} & \multicolumn{1}{c|}{94.8} & \multicolumn{1}{c|}{6.0} & \multicolumn{1}{c|}{16.4} & \multicolumn{1}{c|}{72.3} & \multicolumn{1}{c|}{25.6} & \multicolumn{1}{c|}{51.6} & \multicolumn{1}{c|}{79.9} \\ \cline{2-12} 
\multicolumn{1}{|c|}{\multirow{-6}{*}{\begin{tabular}[c]{@{}c@{}}SILDa Weather\\       and \\      Time of Day\end{tabular}}} & \multicolumn{1}{c|}{ours (Fast-R2D2,GHARM-20,rig+seq)} & \multicolumn{1}{c|}{49.18} & \multicolumn{1}{c|}{28.9} & \multicolumn{1}{c|}{65.9} & \multicolumn{1}{c|}{92.3} & \multicolumn{1}{c|}{0.2} & \multicolumn{1}{c|}{3.4} & \multicolumn{1}{c|}{89.7} & \multicolumn{1}{c|}{28.6} & \multicolumn{1}{c|}{53.4} & \multicolumn{1}{c|}{80.2} \\ \hline
\end{tabular}

%% file: tables/summary_baidu.tex
\begin{tabular}{|c|c|}
\hline
\rowcolor[HTML]{F5E0C7} 
\textbf{Method} & \textbf{\%localized (1m,   5.0°)} \\ \hline
\textbf{ours   (R2D2,frustum,DenseVLAD-50)} & \textbf{85.0} \\ \hline
RootSIFT + VVS & 84.8 \\ \hline
Direct Matching & 83.3 \\ \hline
COV + RootSIFT & 75.4 \\ \hline
\end{tabular}

%% file: tables/summary_median.tex
\begin{tabular}{ccccccccl}
\hline
\rowcolor[HTML]{F5E0C7} 
\multicolumn{1}{|c|}{\cellcolor[HTML]{F5E0C7}\textbf{Datasets}} & \multicolumn{1}{c|}{\cellcolor[HTML]{F5E0C7}\textbf{Method}} & \multicolumn{7}{c|}{\cellcolor[HTML]{F5E0C7}\textbf{Median   6D Localization Errors}} \\ \hline
\multicolumn{1}{|c|}{} & \multicolumn{1}{c|}{} & \multicolumn{1}{c|}{Great Court} & \multicolumn{1}{c|}{K. College} & \multicolumn{1}{c|}{Old Hospital} & \multicolumn{1}{c|}{Shop Facade} & \multicolumn{1}{c|}{St M. Church} & \multicolumn{1}{c|}{Street} & \multicolumn{1}{l|}{} \\ \cline{2-8}
\multicolumn{1}{|c|}{} & \multicolumn{1}{c|}{PoseNet~\cite{kendall2017geometric}} & \multicolumn{1}{c|}{7.00m, 3.7°} & \multicolumn{1}{c|}{0.99m, 1.1°} & \multicolumn{1}{c|}{2.17m, 2.9°} & \multicolumn{1}{c|}{1.05m, 4.0°} & \multicolumn{1}{c|}{1.49m, 3.4°} & \multicolumn{1}{c|}{20.7m, 25.7°} & \multicolumn{1}{l|}{} \\ \cline{2-8}
\multicolumn{1}{|c|}{} & \multicolumn{1}{c|}{ActiveSearch~\cite{SattlerPAMI17EfficientPrioritizedMatching}} & \multicolumn{1}{c|}{0.24m, 0.1°} & \multicolumn{1}{c|}{0.13m, 0.2°} & \multicolumn{1}{c|}{0.20m, 0.4°} & \multicolumn{1}{c|}{0.04m, 0.2°} & \multicolumn{1}{c|}{0.08m, 0.3°} & \multicolumn{1}{c|}{N/A} & \multicolumn{1}{l|}{} \\ \cline{2-8}
\multicolumn{1}{|c|}{} & \multicolumn{1}{c|}{InLoc~\cite{Taira2019TPAMI}} & \multicolumn{1}{c|}{1.20m, 0.6°} & \multicolumn{1}{c|}{0.46m, 0.8°} & \multicolumn{1}{c|}{0.48m, 1.0°} & \multicolumn{1}{c|}{0.11m, 0.5°} & \multicolumn{1}{c|}{0.18m 0.6°} & \multicolumn{1}{c|}{N/A} & \multicolumn{1}{l|}{} \\ \cline{2-8}
\multicolumn{1}{|c|}{} & \multicolumn{1}{c|}{DSAC++~\cite{BrachmannCVPR18LearningLessIsMore6DLocalization}} & \multicolumn{1}{c|}{0.40m, 0.2°} & \multicolumn{1}{c|}{0.18m, 0.3°} & \multicolumn{1}{c|}{0.20m, 0.3°} & \multicolumn{1}{c|}{0.06m, 0.3°} & \multicolumn{1}{c|}{0.13m, 0.4°} & \multicolumn{1}{c|}{-} & \multicolumn{1}{l|}{} \\ \cline{2-8}
\multicolumn{1}{|c|}{} & \multicolumn{1}{c|}{DSAC*~\cite{brachmann2020visual}} & \multicolumn{1}{c|}{0.49m, 0.3°} & \multicolumn{1}{c|}{0.15m, 0.3°} & \multicolumn{1}{c|}{0.21m, 0.4°} & \multicolumn{1}{c|}{0.05m, 0.3°} & \multicolumn{1}{c|}{0.13m, 0.4°} & \multicolumn{1}{c|}{-} & \multicolumn{1}{l|}{} \\ \cline{2-8}
\multicolumn{1}{|c|}{} & \multicolumn{1}{c|}{HACNet~\cite{hieraSceneCoordCVPR20}} & \multicolumn{1}{c|}{0.28m, 0.2°} & \multicolumn{1}{c|}{0.18m, 0.3°} & \multicolumn{1}{c|}{0.19m, 0.3°} & \multicolumn{1}{c|}{0.06m, 0.3°} & \multicolumn{1}{c|}{0.09m, 0.3°} & \multicolumn{1}{c|}{N/A} & \multicolumn{1}{l|}{} \\ \cline{2-8}
\multicolumn{1}{|c|}{} & \multicolumn{1}{c|}{PixLoc~\cite{sarlin2021feature}} & \multicolumn{1}{c|}{0.30m, 0.1°} & \multicolumn{1}{c|}{0.14m, 0.2°} & \multicolumn{1}{c|}{0.16m, 0.3°} & \multicolumn{1}{c|}{0.05m, 0.2°} & \multicolumn{1}{c|}{0.10m, 0.3°} & \multicolumn{1}{c|}{N/A} & \multicolumn{1}{l|}{} \\ \cline{2-8}
\multicolumn{1}{|c|}{} & \multicolumn{1}{c|}{hloc + SuperGlue~\cite{SarlinCVPR20SuperGlueLearningFeatureMatchingGNN}} & \multicolumn{1}{c|}{0.16m, 0.1°} & \multicolumn{1}{c|}{0.12m, 0.2°} & \multicolumn{1}{c|}{0.15m, 0.3°} & \multicolumn{1}{c|}{0.04m, 0.2°} & \multicolumn{1}{c|}{0.07m, 0.2°} & \multicolumn{1}{c|}{N/A} & \multicolumn{1}{l|}{} \\ \cline{2-8}
\multicolumn{1}{|c|}{\multirow{-10}{*}{\begin{tabular}[c]{@{}c@{}}Cambridge\\      Landmarks\end{tabular}}} & \multicolumn{1}{c|}{\textbf{ours (R2D2,APGeM-20)}} & \multicolumn{1}{c|}{0.10m, 0.0°} & \multicolumn{1}{c|}{0.05m, 0.1°} & \multicolumn{1}{c|}{0.09m, 0.2°} & \multicolumn{1}{c|}{0.02m, 0.1°} & \multicolumn{1}{c|}{0.03m, 0.1°} & \multicolumn{1}{c|}{0.10m, 0.3°} & \multicolumn{1}{l|}{\multirow{-10}{*}{}} \\ \hline
\multicolumn{9}{l}{} \\ \hline
\multicolumn{1}{|c|}{} & \multicolumn{1}{c|}{} & \multicolumn{1}{c|}{Chess} & \multicolumn{1}{c|}{Fire} & \multicolumn{1}{c|}{Heads} & \multicolumn{1}{c|}{Office} & \multicolumn{1}{c|}{Pumpkin} & \multicolumn{1}{c|}{Stairs} & \multicolumn{1}{c|}{Kitchen} \\ \cline{2-9} 
\multicolumn{1}{|c|}{} & \multicolumn{1}{c|}{PoseNet~\cite{kendall2017geometric}} & \multicolumn{1}{c|}{0.13m, 4.5°} & \multicolumn{1}{c|}{0.27m, 11.3°} & \multicolumn{1}{c|}{0.17m, 13.0°} & \multicolumn{1}{c|}{0.19m, 5.6°} & \multicolumn{1}{c|}{0.26m, 4.8°} & \multicolumn{1}{c|}{0.35m, 12.4°} & \multicolumn{1}{c|}{0.23m, 5.4°} \\ \cline{2-9} 
\multicolumn{1}{|c|}{} & \multicolumn{1}{c|}{ActiveSearch~\cite{SattlerPAMI17EfficientPrioritizedMatching}} & \multicolumn{1}{c|}{0.03m, 0.9°} & \multicolumn{1}{c|}{0.02m, 1.0°} & \multicolumn{1}{c|}{0.01m, 0.8°} & \multicolumn{1}{c|}{0.04m, 1.2°} & \multicolumn{1}{c|}{0.07m, 1.7°} & \multicolumn{1}{c|}{0.04m, 1.0°} & \multicolumn{1}{c|}{0.05m, 1.7°} \\ \cline{2-9} 
\multicolumn{1}{|c|}{} & \multicolumn{1}{c|}{InLoc~\cite{Taira2019TPAMI}} & \multicolumn{1}{c|}{0.03m, 1.1°} & \multicolumn{1}{c|}{0.03m, 1.1°} & \multicolumn{1}{c|}{0.02m, 1.2°} & \multicolumn{1}{c|}{0.03m, 1.1°} & \multicolumn{1}{c|}{0.05m, 1.6°} & \multicolumn{1}{c|}{0.09m, 2.5°} & \multicolumn{1}{c|}{0.04m, 1.3°} \\ \cline{2-9} 
\multicolumn{1}{|c|}{} & \multicolumn{1}{c|}{DSAC++~\cite{BrachmannCVPR18LearningLessIsMore6DLocalization}} & \multicolumn{1}{c|}{0.02m, 0.5°} & \multicolumn{1}{c|}{0.02m, 0.9°} & \multicolumn{1}{c|}{0.01m, 0.8°} & \multicolumn{1}{c|}{0.03m, 0.7°} & \multicolumn{1}{c|}{0.04m, 1.1°} & \multicolumn{1}{c|}{0.09m, 2.6°} & \multicolumn{1}{c|}{0.04m, 1.1°} \\ \cline{2-9} 
\multicolumn{1}{|c|}{} & \multicolumn{1}{c|}{HACNet~\cite{hieraSceneCoordCVPR20}} & \multicolumn{1}{c|}{0.02m, 0.7°} & \multicolumn{1}{c|}{0.02m, 0.9°} & \multicolumn{1}{c|}{0.01m, 0.9°} & \multicolumn{1}{c|}{0.03m, 0.8°} & \multicolumn{1}{c|}{0.04m, 1.0°} & \multicolumn{1}{c|}{0.03m, 0.8°} & \multicolumn{1}{c|}{0.04m, 1.2°} \\ \cline{2-9} 
\multicolumn{1}{|c|}{} & \multicolumn{1}{c|}{PixLoc~\cite{sarlin2021feature}} & \multicolumn{1}{c|}{0.02m, 0.8°} & \multicolumn{1}{c|}{0.02m, 0.7°} & \multicolumn{1}{c|}{0.01m, 0.8°} & \multicolumn{1}{c|}{0.03m, 0.8°} & \multicolumn{1}{c|}{0.04m, 1.2°} & \multicolumn{1}{c|}{0.05m, 1.3°} & \multicolumn{1}{c|}{0.03m, 1.2°} \\ \cline{2-9} 
\multicolumn{1}{|c|}{} & \multicolumn{1}{c|}{hloc + SuperGlue~\cite{SarlinCVPR20SuperGlueLearningFeatureMatchingGNN}} & \multicolumn{1}{c|}{0.02m, 0.9°} & \multicolumn{1}{c|}{0.02m, 0.9°} & \multicolumn{1}{c|}{0.01m, 0.8°} & \multicolumn{1}{c|}{0.03m, 0.9°} & \multicolumn{1}{c|}{0.05m, 1.3°} & \multicolumn{1}{c|}{0.05m, 1.5°} & \multicolumn{1}{c|}{0.04m, 1.4°} \\ \cline{2-9} 
\multicolumn{1}{|c|}{\multirow{-9}{*}{7-Scenes}} & \multicolumn{1}{c|}{\textbf{\begin{tabular}[c]{@{}c@{}}ours\\      (R2D2,RGBD,APGeM-20)\end{tabular}}} & \multicolumn{1}{c|}{0.02m, 0.8°} & \multicolumn{1}{c|}{0.02m, 0.8°} & \multicolumn{1}{c|}{0.01m, 0.7°} & \multicolumn{1}{c|}{0.03m, 0.8°} & \multicolumn{1}{c|}{0.04m, 1.1°} & \multicolumn{1}{c|}{0.04m, 1.1°} & \multicolumn{1}{c|}{0.03m, 1.1°} \\ \hline
\end{tabular}

%% file: tables/aachen_v2.tex
\begin{tabular}{|l|c|c|c|c|c|c|c|}
\hline
\rowcolor[HTML]{F5E0C7} 
\multicolumn{1}{|c|}{\cellcolor[HTML]{F5E0C7}} & \cellcolor[HTML]{F5E0C7} & \multicolumn{3}{c|}{\cellcolor[HTML]{F5E0C7}\textbf{day}} & \multicolumn{3}{c|}{\cellcolor[HTML]{F5E0C7}\textbf{night}} \\ \cline{3-8} 
\rowcolor[HTML]{F5E0C7} 
\multicolumn{1}{|c|}{\multirow{-2}{*}{\cellcolor[HTML]{F5E0C7}\textbf{Algorithm}}} & \multirow{-2}{*}{\cellcolor[HTML]{F5E0C7}\textbf{\begin{tabular}[c]{@{}c@{}}avg. all\\      bins\end{tabular}}} & \textbf{high} & \textbf{mid} & \textbf{low} & \textbf{high} & \textbf{mid} & \textbf{low} \\ \hline
\textbf{R2D2   40k,GHARM-50,config2} & \textbf{92.45} & 90.9 & 96.7 & 99.5 & 78.5 & 91.1 & 98.0 \\ \hline
hloc+SuperGlue~\cite{SarlinCVPR20SuperGlueLearningFeatureMatchingGNN} & 92.15 & 89.8 & 96.1 & 99.4 & 77.0 & 90.6 & 100.0 \\ \hline
Fast-R2D2,GHARM-20,config2 & 91.75 & 90.3 & 96.6 & 99.6 & 74.9 & 90.1 & 99.0 \\ \hline
R2D2,GHARM-20,config2 & 91.68 & 90.5 & 96.8 & 99.4 & 74.9 & 90.1 & 98.4 \\ \hline
R2D2,mean and   power-20,config2 & 91.33 & 90.7 & 96.8 & 99.3 & 73.3 & 89.5 & 98.4 \\ \hline
Fast-R2D2,APGeM-20,config2 & 90.98 & 89.8 & 96.5 & 99.5 & 74.3 & 87.4 & 98.4 \\ \hline
R2D2,round   robin-20,config2 & 90.70 & 89.6 & 96.6 & 99.4 & 72.8 & 87.4 & 98.4 \\ \hline
R2D2,min and   max-20,config2 & 90.70 & 89.9 & 96.4 & 99.2 & 73.3 & 88.0 & 97.4 \\ \hline
R2D2,max-20,config2 & 90.65 & 89.2 & 96.0 & 99.0 & 73.8 & 88.0 & 97.9 \\ \hline
R2D2,NetVLAD-20,config2 & 90.60 & 88.7 & 95.1 & 98.1 & 74.3 & 89.5 & 97.9 \\ \hline
R2D2,DELG-20,config2 & 90.55 & 90.0 & 96.0 & 99.2 & 73.3 & 87.4 & 97.4 \\ \hline
R2D2,APGeM-20,config2 & 90.32 & 89.9 & 96.5 & 99.5 & 71.2 & 86.9 & 97.9 \\ \hline
ASLFeat~\cite{aslfeatCVPR20},APGeM-20,config2 & 89.97 & 87.5 & 95.4 & 99.3 & 73.8 & 85.9 & 97.9 \\ \hline
RLOCS\_v1.0~\cite{fan2020visual} & 89.90 & 86.0 & 94.8 & 98.8 & 72.3 & 88.5 & 99.0 \\ \hline
D2-Net,APGeM-20,config2 & 89.15 & 85.8 & 94.3 & 98.8 & 70.7 & 86.9 & 98.4 \\ \hline
R2D2,min-20,config2 & 88.40 & 89.1 & 95.0 & 98.2 & 71.2 & 84.8 & 92.1 \\ \hline
R2D2,DenseVLAD-20,config2 & 84.93 & 88.2 & 94.2 & 97.3 & 62.8 & 79.1 & 88.0 \\ \hline
R2D2,APGeM-20,config1 & 83.72 & 60.8 & 80.9 & 99.3 & 75.4 & 88.0 & 97.9 \\ \hline
COLMAP   SIFT,APGeM-20,config2 & 79.17 & 87.7 & 94.2 & 98.4 & 51.8 & 65.4 & 77.5 \\ \hline
COLMAP~\cite{SchonbergerCVPR16StructureFromMotionRevisited} & 66.18 & 80.2 & 88.3 & 93.0 & 33.0 & 46.1 & 56.5 \\ \hline
\end{tabular}

%% file: tables/inloc_v2.tex
\begin{tabular}{|l|c|c|c|c|c|c|c|}
\hline
\rowcolor[HTML]{F5E0C7} 
\multicolumn{1}{|c|}{\cellcolor[HTML]{F5E0C7}} & \cellcolor[HTML]{F5E0C7} & \multicolumn{3}{c|}{\cellcolor[HTML]{F5E0C7}\textbf{DUC1}} & \multicolumn{3}{c|}{\cellcolor[HTML]{F5E0C7}\textbf{DUC2}} \\ \cline{3-8} 
\rowcolor[HTML]{F5E0C7} 
\multicolumn{1}{|c|}{\multirow{-2}{*}{\cellcolor[HTML]{F5E0C7}\textbf{Algorithm}}} & \multirow{-2}{*}{\cellcolor[HTML]{F5E0C7}\textbf{\begin{tabular}[c]{@{}c@{}}avg. all\\      bins\end{tabular}}} & \textbf{high} & \textbf{mid} & \textbf{low} & \textbf{high} & \textbf{mid} & \textbf{low} \\ \hline
\textbf{RLOCS\_v1.0~\cite{fan2020visual}} & \textbf{70.10} & 47.00 & 71.2 & 84.8 & 58.8 & 77.9 & 80.9 \\ \hline
hloc+SuperGlue~\cite{SarlinCVPR20SuperGlueLearningFeatureMatchingGNN} & 68.57 & 49.00 & 68.7 & 80.8 & 53.4 & 77.1 & 82.4 \\ \hline
R2D2,mean and   power-50,config2 & 59.77 & 39.90 & 57.6 & 71.7 & 48.9 & 67.2 & 73.3 \\ \hline
Fast-R2D2,GHARM-50,config2 & 59.55 & 40.40 & 63.1 & 78.3 & 42.7 & 64.9 & 67.9 \\ \hline
R2D2,GHARM-50,config2 & 58.68 & 39.90 & 58.1 & 71.7 & 45.8 & 65.6 & 71.0 \\ \hline
DensePV+S (w/   scan-graph)~\cite{taira2018rightplace} & 58.62 & 40.90 & 63.6 & 71.7 & 42.7 & 61.8 & 71.0 \\ \hline
R2D2,min and   max-50,config2 & 56.93 & 37.90 & 56.6 & 70.7 & 44.3 & 62.6 & 69.5 \\ \hline
R2D2,NetVLAD-50,config2 & 56.03 & 36.90 & 58.1 & 70.2 & 38.2 & 62.6 & 70.2 \\ \hline
R2D2,DenseVLAD-50,config2 & 56.02 & 33.80 & 51.5 & 67.7 & 45.0 & 65.6 & 72.5 \\ \hline
R2D2,min-50,config2 & 55.85 & 32.80 & 54.5 & 67.7 & 42.7 & 65.6 & 71.8 \\ \hline
InLoc~\cite{Taira2019TPAMI} & 54.80 & 40.90 & 58.1 & 70.2 & 35.9 & 54.2 & 69.5 \\ \hline
R2D2,DELG-50,config2 & 55.17 & 38.40 & 56.1 & 71.7 & 37.4 & 59.5 & 67.9 \\ \hline
Fast-R2D2,APGeM-50,config2 & 54.48 & 34.80 & 56.6 & 73.7 & 35.9 & 59.5 & 66.4 \\ \hline
R2D2,max-50,config2 & 54.20 & 39.90 & 55.1 & 69.2 & 38.9 & 58.0 & 64.1 \\ \hline
R2D2,APGeM-50,config2 & 53.68 & 38.40 & 58.6 & 70.2 & 37.4 & 55.7 & 61.8 \\ \hline
R2D2,round   robin-50,config2 & 47.13 & 33.80 & 51.5 & 66.2 & 29.8 & 48.1 & 53.4 \\ \hline
R2D2,APGeM-50,config1 & 41.02 & 28.30 & 43.4 & 60.6 & 25.2 & 39.7 & 48.9 \\ \hline
COLMAP~\cite{SchonbergerCVPR16StructureFromMotionRevisited} & 32.13 & 28.80 & 43.4 & 59.6 & 13.7 & 20.6 & 26.7 \\ \hline
\end{tabular}

%% file: tables/robotcar_v2.tex
\begin{tabular}{|l|c|c|c|c|c|c|c|}
\hline
\rowcolor[HTML]{F5E0C7} 
\multicolumn{1}{|c|}{\cellcolor[HTML]{F5E0C7}} & \cellcolor[HTML]{F5E0C7} & \multicolumn{3}{c|}{\cellcolor[HTML]{F5E0C7}\textbf{day}} & \multicolumn{3}{c|}{\cellcolor[HTML]{F5E0C7}\textbf{night}} \\ \cline{3-8} 
\rowcolor[HTML]{F5E0C7} 
\multicolumn{1}{|c|}{\multirow{-2}{*}{\cellcolor[HTML]{F5E0C7}\textbf{Algorithm}}} & \multirow{-2}{*}{\cellcolor[HTML]{F5E0C7}\textbf{\begin{tabular}[c]{@{}c@{}}avg. all\\      bins\end{tabular}}} & \textbf{high} & \textbf{mid} & \textbf{low} & \textbf{high} & \textbf{mid} & \textbf{low} \\ \hline
\textbf{Fast-R2D2,APGeM-20,config1   rig seq} & \textbf{80.85} & 65.9 & 95.1 & 100.0 & 46.2 & 81.4 & 96.5 \\ \hline
hloc+SuperGlue~\cite{SarlinCVPR20SuperGlueLearningFeatureMatchingGNN} & 80.77 & 63.8 & 95.0 & 100.0 & 45.0 & 86.2 & 94.6 \\ \hline
Fast-R2D2,GHARM-20,config1   rig seq & 80.17 & 65.9 & 95.1 & 100.0 & 45.2 & 80.4 & 94.4 \\ \hline
R2D2,mean and   power-20,config1 rig seq & 79.93 & 65.9 & 95.1 & 100.0 & 46.4 & 79.7 & 92.5 \\ \hline
R2D2,GHARM-20,config1   rig seq & 79.20 & 66.0 & 95.1 & 100.0 & 46.2 & 76.5 & 91.4 \\ \hline
R2D2,APGeM-20,config1   rig seq & 79.17 & 65.7 & 95.1 & 100.0 & 43.6 & 76.7 & 93.9 \\ \hline
ASLFeat~\cite{aslfeatCVPR20},APGeM-20,config1   rig seq & 78.98 & 64.7 & 94.9 & 99.9 & 45.2 & 77.4 & 91.8 \\ \hline
R2D2,round   robin-20,config1 rig seq & 78.78 & 65.9 & 95.1 & 100.0 & 42.4 & 75.1 & 94.2 \\ \hline
R2D2,DenseVLAD-20,config1   rig seq & 78.17 & 65.7 & 95.1 & 100.0 & 41.3 & 74.1 & 92.8 \\ \hline
R2D2,APGeM-20,config1   rig & 77.85 & 65.7 & 95.1 & 100.0 & 43.6 & 76.0 & 86.7 \\ \hline
R2D2,APGeM-20,config2 & 77.75 & 65.9 & 95.1 & 99.9 & 43.4 & 74.8 & 87.4 \\ \hline
R2D2,min-20,config1   rig seq & 77.55 & 65.8 & 95.1 & 100.0 & 40.8 & 72.0 & 91.6 \\ \hline
R2D2,min and   max-20,config1 rig seq & 76.80 & 65.7 & 95.1 & 100.0 & 41.3 & 72.5 & 86.2 \\ \hline
R2D2,NetVLAD-20,config1   rig seq & 76.28 & 65.6 & 95.1 & 100.0 & 35.7 & 70.4 & 90.9 \\ \hline
R2D2,APGeM-20,config1 & 76.25 & 65.7 & 95.1 & 99.9 & 41.5 & 73.0 & 82.3 \\ \hline
R2D2,DELG-20,config1   rig seq & 73.50 & 65.6 & 94.6 & 99.6 & 37.8 & 64.6 & 78.8 \\ \hline
R2D2,max-20,config1   rig seq & 73.30 & 65.6 & 94.7 & 99.7 & 36.6 & 63.2 & 80.0 \\ \hline
COLMAP~\cite{SchonbergerCVPR16StructureFromMotionRevisited}   config1 rig seq & 50.38 & 64.2 & 94.2 & 99.8 & 8.9 & 15.6 & 19.6 \\ \hline
COLMAP~\cite{SchonbergerCVPR16StructureFromMotionRevisited}   config1 rig & 49.33 & 64.2 & 93.8 & 98.8 & 8.4 & 14.9 & 15.9 \\ \hline
COLMAP~\cite{SchonbergerCVPR16StructureFromMotionRevisited}   config1 & 47.90 & 62.9 & 91.3 & 94.9 & 8.4 & 14.5 & 15.4 \\ \hline
\end{tabular}

%% file: tables/ecmu_v2.tex
\begin{tabular}{|l|c|c|c|c|c|c|c|c|c|c|}
\hline
\rowcolor[HTML]{F5E0C7} 
\multicolumn{1}{|c|}{\cellcolor[HTML]{F5E0C7}} & \cellcolor[HTML]{F5E0C7} & \multicolumn{3}{c|}{\cellcolor[HTML]{F5E0C7}\textbf{urban}} & \multicolumn{3}{c|}{\cellcolor[HTML]{F5E0C7}\textbf{suburban}} & \multicolumn{3}{c|}{\cellcolor[HTML]{F5E0C7}\textbf{park}} \\ \cline{3-11} 
\rowcolor[HTML]{F5E0C7} 
\multicolumn{1}{|c|}{\multirow{-2}{*}{\cellcolor[HTML]{F5E0C7}\textbf{Algorithm}}} & \multirow{-2}{*}{\cellcolor[HTML]{F5E0C7}\textbf{\begin{tabular}[c]{@{}c@{}}avg all\\      bins\end{tabular}}} & \textbf{high} & \textbf{mid} & \textbf{low} & \textbf{high} & \textbf{mid} & \textbf{low} & \textbf{high} & \textbf{mid} & \textbf{low} \\ \hline
\textbf{FGSN~\cite{larsson2019fine}} & \textbf{98.98} & 94.1 & 99.3 & 100.0 & 100.0 & 100.0 & 100.0 & 97.6 & 99.9 & 99.9 \\ \hline
hloc+SuperGlue~\cite{SarlinCVPR20SuperGlueLearningFeatureMatchingGNN} & 98.38 & 98.1 & 99.8 & 99.9 & 98.3 & 99.5 & 100.0 & 94.2 & 97.1 & 98.5 \\ \hline
Fast-R2D2,GHARM-20,config1   rig seq & 96.87 & 97.3 & 99.4 & 99.8 & 95.6 & 97.2 & 99.4 & 90.8 & 94.6 & 97.7 \\ \hline
R2D2,GHARM-20,config1   rig seq & 96.38 & 97.0 & 99.1 & 99.8 & 95.0 & 97.0 & 99.4 & 89.2 & 93.4 & 97.5 \\ \hline
R2D2,mean and   power-20,config1 rig seq & 96.36 & 97.0 & 99.2 & 99.7 & 94.7 & 97.0 & 99.5 & 89.2 & 93.4 & 97.5 \\ \hline
R2D2,round   robin-20,config1 rig seq & 96.27 & 96.9 & 99.1 & 99.7 & 94.8 & 97.0 & 99.4 & 88.8 & 93.1 & 97.6 \\ \hline
R2D2,min and   max-20,config1 rig seq & 96.24 & 96.7 & 98.9 & 99.6 & 95.0 & 97.1 & 99.5 & 88.6 & 93.2 & 97.6 \\ \hline
R2D2,NetVLAD-20,config1   rig seq & 95.94 & 97.1 & 99.1 & 99.8 & 93.8 & 96.3 & 99.1 & 88.1 & 92.7 & 97.5 \\ \hline
R2D2,min-20,config1   rig seq & 95.77 & 96.3 & 98.5 & 99.4 & 94.2 & 96.6 & 99.2 & 88.0 & 92.4 & 97.3 \\ \hline
R2D2,DenseVLAD-20,config1   rig seq & 95.66 & 96.1 & 98.4 & 99.4 & 94.2 & 96.7 & 99.1 & 87.9 & 92.3 & 96.8 \\ \hline
Fast-R2D2,APGeM-20,config1   rig seq & 95.53 & 97.2 & 99.3 & 99.8 & 95.0 & 97.0 & 99.2 & 85.9 & 90.8 & 95.6 \\ \hline
R2D2,max-20,config1   rig seq & 95.04 & 96.8 & 98.9 & 99.7 & 94.4 & 97.0 & 99.3 & 84.1 & 89.2 & 96.0 \\ \hline
DenseVLAD   D2-Net~\cite{DusmanuCVPR19D2NetDeepLocalFeatures} & 95.02 & 94.0 & 97.7 & 99.1 & 93.0 & 95.7 & 98.3 & 89.2 & 93.2 & 95.0 \\ \hline
R2D2,DELG-20,config1   rig seq & 95.00 & 96.6 & 98.8 & 99.7 & 94.1 & 96.7 & 99.1 & 84.7 & 89.6 & 95.7 \\ \hline
R2D2,APGeM-20,config1   rig seq & 94.87 & 96.7 & 98.9 & 99.7 & 94.4 & 96.8 & 99.2 & 83.6 & 89.0 & 95.5 \\ \hline
R2D2,APGeM-20,config1   rig & 94.30 & 96.5 & 98.8 & 99.5 & 94.3 & 96.7 & 99.1 & 83.1 & 87.9 & 92.8 \\ \hline
R2D2,APGeM-20,config2 & 90.71 & 95.9 & 98.1 & 98.9 & 89.5 & 92.1 & 95.2 & 78.3 & 82.0 & 86.4 \\ \hline
R2D2,APGeM-20,config1 & 89.11 & 95.8 & 98.1 & 98.8 & 88.9 & 91.1 & 93.4 & 75.5 & 78.4 & 82.0 \\ \hline
Active Search   v1.1~\cite{SattlerPAMI17EfficientPrioritizedMatching} & 70.48 & 81.0 & 87.3 & 92.4 & 62.6 & 70.9 & 81.0 & 45.5 & 51.6 & 62.0 \\ \hline
COLMAP~\cite{SchonbergerCVPR16StructureFromMotionRevisited}   config1 rig seq & 62.73 & 82.7 & 87.1 & 92.1 & 60.1 & 68.0 & 78.8 & 28.3 & 30.5 & 37.0 \\ \hline
COLMAP~\cite{SchonbergerCVPR16StructureFromMotionRevisited}   config1 rig & 53.98 & 77.7 & 80.3 & 80.8 & 52.1 & 56.0 & 57.5 & 26.2 & 27.5 & 27.7 \\ \hline
COLMAP~\cite{SchonbergerCVPR16StructureFromMotionRevisited}   config1 & 39.50 & 63.0 & 65.1 & 65.3 & 35.2 & 37.4 & 38.3 & 16.7 & 17.2 & 17.3 \\ \hline
\end{tabular}

%% file: tables/silda_v2.tex
\begin{tabular}{|l|c|c|c|c|c|c|c|c|c|c|}
\hline
\rowcolor[HTML]{F5E0C7} 
\multicolumn{1}{|c|}{\cellcolor[HTML]{F5E0C7}} & \cellcolor[HTML]{F5E0C7} & \multicolumn{3}{c|}{\cellcolor[HTML]{F5E0C7}\textbf{evening}} & \multicolumn{3}{c|}{\cellcolor[HTML]{F5E0C7}\textbf{snow}} & \multicolumn{3}{c|}{\cellcolor[HTML]{F5E0C7}\textbf{night}} \\ \cline{3-11} 
\rowcolor[HTML]{F5E0C7} 
\multicolumn{1}{|c|}{\multirow{-2}{*}{\cellcolor[HTML]{F5E0C7}\textbf{Algorithm}}} & \multirow{-2}{*}{\cellcolor[HTML]{F5E0C7}\textbf{\begin{tabular}[c]{@{}c@{}}avg. all\\      bins\end{tabular}}} & \textbf{high} & \textbf{mid} & \textbf{low} & \textbf{high} & \textbf{mid} & \textbf{low} & \textbf{high} & \textbf{mid} & \textbf{low} \\ \hline
\textbf{hloc+SuperGlue~\cite{SarlinCVPR20SuperGlueLearningFeatureMatchingGNN}} & \textbf{51.59} & 35.5 & 75.0 & 97.1 & 0.0 & 2.4 & 86.3 & 31.7 & 54.4 & 81.9 \\ \hline
R2D2,GHARM-20,config1   rig seq & 50.22 & 32.4 & 67.4 & 93.3 & 0.2 & 4.1 & 88.9 & 30.4 & 54.2 & 81.1 \\ \hline
R2D2,min and   max-20,config1 rig seq & 50.00 & 32.1 & 67.6 & 93.4 & 0.2 & 3.3 & 88.7 & 30.5 & 54.0 & 80.2 \\ \hline
R2D2,APGeM-20,config1   rig & 49.97 & 31.9 & 66.6 & 92.5 & 0.5 & 5.8 & 89.2 & 30.5 & 54.2 & 78.5 \\ \hline
R2D2,APGeM-20,config1   rig seq & 49.97 & 31.9 & 66.6 & 92.5 & 0.5 & 5.8 & 89.2 & 30.5 & 54.2 & 78.5 \\ \hline
R2D2,mean and   power-20,config1 rig seq & 49.88 & 31.7 & 67.0 & 93.5 & 0.0 & 2.7 & 88.5 & 30.4 & 54.4 & 80.7 \\ \hline
Fast-R2D2,APGeM-20,config1   rig seq & 49.76 & 32.3 & 65.8 & 91.8 & 0.3 & 3.9 & 89.4 & 30.2 & 54.3 & 79.8 \\ \hline
Fast-R2D2,APGeM-20,config1   rig & 49.74 & 32.3 & 65.8 & 91.7 & 0.3 & 3.9 & 89.4 & 30.2 & 54.3 & 79.8 \\ \hline
R2D2,round   robin-20,config1 rig seq & 49.61 & 31.8 & 66.9 & 92.0 & 0.0 & 3.4 & 89.6 & 29.3 & 54.0 & 79.5 \\ \hline
R2D2,DELG-20,config1   rig seq & 49.42 & 31.3 & 66.4 & 92.1 & 0.2 & 7.5 & 85.6 & 30.2 & 54.1 & 77.4 \\ \hline
NetVLAD   (top50) D2-Net~\cite{DusmanuCVPR19D2NetDeepLocalFeatures} & 49.33 & 29.6 & 67.8 & 94.8 & 6.0 & 16.4 & 72.3 & 25.6 & 51.6 & 79.9 \\ \hline
Fast-R2D2,GHARM-20,config1   rig seq & 49.18 & 28.9 & 65.9 & 92.3 & 0.2 & 3.4 & 89.7 & 28.6 & 53.4 & 80.2 \\ \hline
Fast-R2D2,GHARM-20,config1   rig & 49.17 & 28.9 & 65.9 & 92.2 & 0.2 & 3.4 & 89.7 & 28.6 & 53.4 & 80.2 \\ \hline
R2D2,NetVLAD-20,config1   rig seq & 48.99 & 31.6 & 66.5 & 91.0 & 0.0 & 2.7 & 89.6 & 28.9 & 52.1 & 78.5 \\ \hline
R2D2,max-20,config1   rig seq & 48.89 & 31.8 & 65.9 & 92.4 & 0.3 & 2.7 & 84.6 & 30.4 & 54.1 & 77.8 \\ \hline
R2D2,APGeM-20,config2 & 48.00 & 33.1 & 67.8 & 92.3 & 0.3 & 5.3 & 70.7 & 30.3 & 53.5 & 78.7 \\ \hline
Fast-R2D2,APGeM-20,config2 & 47.49 & 32.3 & 66.9 & 90.9 & 0.3 & 5.5 & 68.8 & 30.3 & 53.3 & 79.1 \\ \hline
R2D2,min-20,config1   rig seq & 46.47 & 30.1 & 63.9 & 82.5 & 0.0 & 1.9 & 76.2 & 28.1 & 54.1 & 81.4 \\ \hline
R2D2,APGeM-20,config1 & 46.39 & 31.8 & 66.3 & 89.4 & 0.3 & 3.9 & 64.9 & 30.0 & 53.4 & 77.5 \\ \hline
Fast-R2D2,GHARM-20,config1 & 46.08 & 28.7 & 65.4 & 88.8 & 0.2 & 3.4 & 66.6 & 28.6 & 53.2 & 79.8 \\ \hline
Fast-R2D2,APGeM-20,config1 & 45.79 & 32.2 & 65.4 & 88.3 & 0.2 & 2.4 & 61.8 & 29.8 & 53.4 & 78.6 \\ \hline
COLMAP~\cite{SchonbergerCVPR16StructureFromMotionRevisited}   rig seq & 43.01 & 32.3 & 67.9 & 93.4 & 0.3 & 11.3 & 68.2 & 18.3 & 33.8 & 61.6 \\ \hline
COLMAP~\cite{SchonbergerCVPR16StructureFromMotionRevisited}   rig & 42.99 & 32.3 & 67.9 & 93.4 & 0.3 & 11.3 & 68.0 & 18.3 & 33.8 & 61.6 \\ \hline
R2D2,DenseVLAD-20,config1   rig seq & 42.90 & 27.9 & 59.7 & 75.6 & 0.0 & 1.9 & 59.1 & 29.3 & 52.9 & 79.7 \\ \hline
COLMAP~\cite{SchonbergerCVPR16StructureFromMotionRevisited} & 40.57 & 32.1 & 67.2 & 92.4 & 0.3 & 8.6 & 56.5 & 17.3 & 31.8 & 58.9 \\ \hline
\end{tabular}

%% file: tables/baidu_suppl.tex
\begin{tabular}{|c|c|}
\hline
\rowcolor[HTML]{F5E0C7} 
\textbf{Method} & \textbf{\%localized (1m,   5.0°)} \\ \hline
\textbf{R2D2,frustum,DenseVLAD-50,config2} & \textbf{85.0} \\ \hline
RootSIFT + VVS & 84.8 \\ \hline
R2D2,frustum,NetVLAD-50,config2 & 84.1 \\ \hline
Direct Matching & 83.3 \\ \hline
R2D2,frustum,APGeM-50,config2 & 82.6 \\ \hline
R2D2,frustum,DELG-50,config2 & 82.6 \\ \hline
R2D2,distance-20,APGeM-20,config2 & 79.9 \\ \hline
R2D2,APGeM-20,config2 & 79.1 \\ \hline
R2D2,frustum-20,APGeM-20,config2 & 79.0 \\ \hline
COV + RootSIFT & 75.4 \\ \hline
\end{tabular}

%% file: tables/gangnam_b2_arxiv.tex
\begin{tabular}{|c|cccc|}
\hline
\rowcolor[HTML]{F5E0C7} 
\textbf{Algorithm}      & \multicolumn{4}{c|}{\cellcolor[HTML]{F5E0C7}\textbf{Localization   Errors}}                        \\ \hline
                        & \multicolumn{1}{c|}{}               & \multicolumn{3}{c|}{test}                                    \\ \cline{3-5} 
\multirow{-2}{*}{}         & \multicolumn{1}{c|}{\multirow{-2}{*}{avg. all bins}} & \multicolumn{1}{c|}{high} & \multicolumn{1}{c|}{mid}  & low  \\ \hline
\textbf{Ospace}         & \multicolumn{1}{c|}{\textbf{75.13}} & \multicolumn{1}{c|}{57.6} & \multicolumn{1}{c|}{80.6} & 87.2 \\ \hline
MegLoc                  & \multicolumn{1}{c|}{72.07}          & \multicolumn{1}{c|}{55.9} & \multicolumn{1}{c|}{77.5} & 82.8 \\ \hline
R2D2,GHARM-50,ransaclib & \multicolumn{1}{c|}{66.23}          & \multicolumn{1}{c|}{51.2} & \multicolumn{1}{c|}{71.5} & 76.0 \\ \hline
ASLFeat,OpenIBL-50,config2 & \multicolumn{1}{c|}{61.03}                           & \multicolumn{1}{c|}{48.0} & \multicolumn{1}{c|}{65.2} & 69.9 \\ \hline
R2D2-APGeM-50,config2   & \multicolumn{1}{c|}{59.77}          & \multicolumn{1}{c|}{46.0} & \multicolumn{1}{c|}{64.6} & 68.7 \\ \hline
R2D2-OpenIBL-20,config2 & \multicolumn{1}{c|}{57.70}          & \multicolumn{1}{c|}{44.1} & \multicolumn{1}{c|}{62.0} & 67.0 \\ \hline
Onavi                   & \multicolumn{1}{c|}{56.50}          & \multicolumn{1}{c|}{43.4} & \multicolumn{1}{c|}{61.0} & 65.1 \\ \hline
R2D2-APGeM-20,config2   & \multicolumn{1}{c|}{55.30}          & \multicolumn{1}{c|}{42.4} & \multicolumn{1}{c|}{59.7} & 63.8 \\ \hline
\end{tabular}

%% file: tables/7_scenes.tex
\begin{tabular}{|c|c|c|c|c|c|c|c|c|}
\hline
\rowcolor[HTML]{F5E0C7} 
\textbf{Method} & \multicolumn{8}{c|}{\cellcolor[HTML]{F5E0C7}\textbf{Median 6D Localization Errors}} \\ \hline
 & Avg. Median Error & Chess & Fire & Heads & Office & Pumpkin & Stairs & Kitchen \\ \hline
PoseNet~\cite{kendall2017geometric} & 0.23m, 8.1° & 0.13m, 4.5° & 0.27m, 11.3° & 0.17m, 13.0° & 0.19m, 5.6° & 0.26m, 4.8° & 0.35m, 12.4° & 0.23m, 5.4° \\ \hline
ActiveSearch~\cite{SattlerPAMI17EfficientPrioritizedMatching} & 0.04m, 1.2° & 0.03m, 0.9° & 0.02m, 1.0° & 0.01m, 0.8° & 0.04m, 1.2° & 0.07m, 1.7° & 0.04m, 1.0° & 0.05m, 1.7° \\ \hline
InLoc~\cite{Taira2019TPAMI} & 0.04m, 1.4° & 0.03m, 1.1° & 0.03m, 1.1° & 0.02m, 1.2° & 0.03m, 1.1° & 0.05m, 1.6° & 0.09m, 2.5° & 0.04m, 1.3° \\ \hline
DSAC++~\cite{BrachmannCVPR18LearningLessIsMore6DLocalization} & 0.04m, 1.1° & 0.02m, 0.5° & 0.02m, 0.9° & 0.01m, 0.8° & 0.03m, 0.7° & 0.04m, 1.1° & 0.09m, 2.6° & 0.04m, 1.1° \\ \hline
DSAC*~\cite{brachmann2020visual} & 0.02m, 1.3° & 0.01m, 1.0° & 0.01m, 1.1° & 0.01m, 1.9° & 0.01m, 1.0° & 0.02m, 1.2° & 0.03m, 1.2° & 0.02m, 1.4° \\ \hline
\textbf{HACNet~\cite{hieraSceneCoordCVPR20}} & 0.03m, \textbf{0.9°} & 0.02m, 0.7° & 0.02m, 0.9° & 0.01m, 0.9° & 0.03m, 0.8° & 0.04m, 1.0° & 0.03m, 0.8° & 0.04m, 1.2° \\ \hline
PixLoc~\cite{sarlin2021feature} & 0.03m, 1.0° & 0.02m, 0.8° & 0.02m, 0.7° & 0.01m, 0.8° & 0.03m, 0.8° & 0.04m, 1.2° & 0.05m, 1.3° & 0.03m, 1.2° \\ \hline
hloc +   SuperGlue\cite{SarlinCVPR20SuperGlueLearningFeatureMatchingGNN} & 0.03m, 1.1° & 0.02m, 0.9° & 0.02m, 0.9° & 0.01m, 0.8° & 0.03m, 0.9° & 0.05m, 1.3° & 0.05m, 1.5° & 0.04m, 1.4° \\ \hline
\textbf{Relocalisation   Cascade~\cite{Cavallari_2020}} & \textbf{0.01m}, 1.2° & N/A & N/A & N/A & N/A & N/A & N/A & N/A \\ \hline
\textbf{\begin{tabular}[c]{@{}c@{}}R2D2,RGBD,APGeM-20,config2\end{tabular}} & 0.03m, \textbf{0.9°} & 0.02m, 0.8° & 0.02m, 0.8° & 0.01m, 0.7° & 0.03m, 0.8° & 0.04m, 1.1° & 0.04m, 1.1° & 0.03m, 1.1° \\ \hline
\end{tabular}

%% file: tables/12_scenes.tex
\begin{tabular}{cccccccccccccc}
\hline
\multicolumn{1}{|c|}{\cellcolor[HTML]{F5E0C7}} & \multicolumn{13}{c|}{\cellcolor[HTML]{F5E0C7}\textbf{12-scenes,   Median 6D Localization Errors (mm, °)}} \\ \cline{2-14} 
\multicolumn{1}{|c|}{\multirow{-2}{*}{\cellcolor[HTML]{F5E0C7}\textbf{Method}}} & \multicolumn{1}{c|}{} & \multicolumn{2}{c|}{apt1} & \multicolumn{4}{c|}{apt2} & \multicolumn{4}{c|}{office1} & \multicolumn{2}{c|}{office2} \\ \hline
\multicolumn{1}{|c|}{} & \multicolumn{1}{c|}{avg.} & \multicolumn{1}{c|}{kitchen} & \multicolumn{1}{c|}{living} & \multicolumn{1}{c|}{bed} & \multicolumn{1}{c|}{kitchen} & \multicolumn{1}{c|}{living} & \multicolumn{1}{c|}{luke} & \multicolumn{1}{c|}{gates362} & \multicolumn{1}{c|}{gates381} & \multicolumn{1}{c|}{lounge} & \multicolumn{1}{c|}{manolis} & \multicolumn{1}{c|}{5a} & \multicolumn{1}{c|}{5b} \\ \hline
\multicolumn{1}{|c|}{\textbf{R2D2,APGeM-20,config2}} & \multicolumn{1}{c|}{\textbf{9.3, 0.389}} & \multicolumn{1}{c|}{6.8, 0.389} & \multicolumn{1}{c|}{8.2, 0.332} & \multicolumn{1}{c|}{9.5, 0.363} & \multicolumn{1}{c|}{6.9, 0.351} & \multicolumn{1}{c|}{9.5, 0.347} & \multicolumn{1}{c|}{12.0, 0.533} & \multicolumn{1}{c|}{8.7, 0.377} & \multicolumn{1}{c|}{9.6, 0.453} & \multicolumn{1}{c|}{11.5, 0.389} & \multicolumn{1}{c|}{7.4, 0.334} & \multicolumn{1}{c|}{11.1, 0.448} & \multicolumn{1}{c|}{10.3, 0.349} \\ \hline
\multicolumn{1}{|c|}{R2D2,RGBD,APGeM-20,config2} & \multicolumn{1}{c|}{9.5, 0.397} & \multicolumn{1}{c|}{6.4, 0.377} & \multicolumn{1}{c|}{8.8, 0.332} & \multicolumn{1}{c|}{8.5, 0.369} & \multicolumn{1}{c|}{7.9, 0.416} & \multicolumn{1}{c|}{8.5, 0.318} & \multicolumn{1}{c|}{11.8, 0.514} & \multicolumn{1}{c|}{8.8, 0.363} & \multicolumn{1}{c|}{10.1, 0.455} & \multicolumn{1}{c|}{13.3, 0.444} & \multicolumn{1}{c|}{7.6, 0.346} & \multicolumn{1}{c|}{9.7, 0.421} & \multicolumn{1}{c|}{12.6, 0.414} \\ \hline
\multicolumn{14}{l}{} \\ \hline
\multicolumn{1}{|c|}{\cellcolor[HTML]{F5E0C7}} & \multicolumn{13}{c|}{\cellcolor[HTML]{F5E0C7}\textbf{12-scenes, percentage   of   successfully localized images   (0.01m, 1.0°)}} \\ \cline{2-14} 
\multicolumn{1}{|c|}{\multirow{-2}{*}{\cellcolor[HTML]{F5E0C7}\textbf{Method}}} & \multicolumn{1}{c|}{} & \multicolumn{2}{c|}{apt1} & \multicolumn{4}{c|}{apt2} & \multicolumn{4}{c|}{office1} & \multicolumn{2}{c|}{office2} \\ \hline
\multicolumn{1}{|c|}{} & \multicolumn{1}{c|}{avg.} & \multicolumn{1}{c|}{kitchen} & \multicolumn{1}{c|}{living} & \multicolumn{1}{c|}{bed} & \multicolumn{1}{c|}{kitchen} & \multicolumn{1}{c|}{living} & \multicolumn{1}{c|}{luke} & \multicolumn{1}{c|}{gates362} & \multicolumn{1}{c|}{gates381} & \multicolumn{1}{c|}{lounge} & \multicolumn{1}{c|}{manolis} & \multicolumn{1}{c|}{5a} & \multicolumn{1}{c|}{5b} \\ \hline
\multicolumn{1}{|c|}{\textbf{R2D2,AP-GeM-20,config2}} & \multicolumn{1}{c|}{\textbf{56.28}} & \multicolumn{1}{c|}{73.95} & \multicolumn{1}{c|}{61.26} & \multicolumn{1}{c|}{54.90} & \multicolumn{1}{c|}{76.67} & \multicolumn{1}{c|}{55.01} & \multicolumn{1}{c|}{37.82} & \multicolumn{1}{c|}{58.81} & \multicolumn{1}{c|}{53.09} & \multicolumn{1}{c|}{40.37} & \multicolumn{1}{c|}{71.02} & \multicolumn{1}{c|}{44.27} & \multicolumn{1}{c|}{48.15} \\ \hline
\multicolumn{1}{|c|}{R2D2,RGBD,AP-GeM-20,config2} & \multicolumn{1}{c|}{54.33} & \multicolumn{1}{c|}{77.03} & \multicolumn{1}{c|}{57.61} & \multicolumn{1}{c|}{57.35} & \multicolumn{1}{c|}{63.33} & \multicolumn{1}{c|}{61.89} & \multicolumn{1}{c|}{38.94} & \multicolumn{1}{c|}{59.84} & \multicolumn{1}{c|}{48.91} & \multicolumn{1}{c|}{28.13} & \multicolumn{1}{c|}{66.62} & \multicolumn{1}{c|}{53.32} & \multicolumn{1}{c|}{39.01} \\ \hline
\end{tabular}

%% file: tables/cambridge_landmarks.tex
\begin{tabular}{|c|c|c|c|c|c|c|c|}
\hline
\rowcolor[HTML]{F5E0C7} 
\textbf{Method} & \textbf{} & \multicolumn{6}{c|}{\cellcolor[HTML]{F5E0C7}\textbf{Median 6D Localization Errors}} \\ \hline
 & Avg. Median Error* & Great Court & K. College & Old Hospital & Shop Facade & St M. Church & Street \\ \hline
PoseNet~\cite{kendall2017geometric} & 2.54m, 3.0° & 7.00m, 3.7° & 0.99m, 1.1° & 2.17m, 2.9° & 1.05m, 4.0° & 1.49m, 3.4° & 20.7m, 25.7° \\ \hline
ActiveSearch~\cite{SattlerPAMI17EfficientPrioritizedMatching} & 0.14m, 0.2° & 0.24m, 0.1° & 0.13m, 0.2° & 0.20m, 0.4° & 0.04m, 0.2° & 0.08m, 0.3° & N/A \\ \hline
InLoc~\cite{Taira2019TPAMI} & 0.49m, 0.7° & 1.20m, 0.6° & 0.46m, 0.8° & 0.48m, 1.0° & 0.11m, 0.5° & 0.18m 0.6° & N/A \\ \hline
DSAC++~\cite{BrachmannCVPR18LearningLessIsMore6DLocalization} & 0.19m, 0.3° & 0.40m, 0.2° & 0.18m, 0.3° & 0.20m, 0.3° & 0.06m, 0.3° & 0.13m, 0.4° & - \\ \hline
DSAC*~\cite{brachmann2020visual} & 0.21m, 0.3° & 0.49m, 0.3° & 0.15m, 0.3° & 0.21m, 0.4° & 0.05m, 0.3° & 0.13m, 0.4° & - \\ \hline
HACNet~\cite{hieraSceneCoordCVPR20} & 0.16m, 0.3° & 0.28m, 0.2° & 0.18m, 0.3° & 0.19m, 0.3° & 0.06m, 0.3° & 0.09m, 0.3° & N/A \\ \hline
PixLoc~\cite{sarlin2021feature} & 0.15m, 0.2° & 0.30m, 0.1° & 0.14m, 0.2° & 0.16m, 0.3° & 0.05m, 0.2° & 0.10m, 0.3° & N/A \\ \hline
hloc +   SuperGlue~\cite{SarlinCVPR20SuperGlueLearningFeatureMatchingGNN} & 0.11m, 0.2° & 0.16m, 0.1° & 0.12m, 0.2° & 0.15m, 0.3° & 0.04m, 0.2° & 0.07m, 0.2° & N/A \\ \hline
\textbf{R2D2,APGeM-20,config2} & \textbf{0.06m, 0.1°} & 0.10m, 0.0° & 0.05m, 0.1° & 0.09m, 0.2° & 0.02m, 0.1° & 0.03m, 0.1° & 0.10m, 0.3° \\ \hline
\end{tabular}

%% file: tables/pipeline_exps_detailed_arxiv.tex
\begin{tabular}{clccccccccll}
\hline
\rowcolor[HTML]{F5E0C7} 
\multicolumn{1}{|c|}{\cellcolor[HTML]{F5E0C7}\textbf{Datasets}} &
  \multicolumn{1}{c|}{\cellcolor[HTML]{F5E0C7}\textbf{Algorithm}} &
  \multicolumn{10}{c|}{\cellcolor[HTML]{F5E0C7}\textbf{percentage of successfully localized images}} \\ \hline
\multicolumn{1}{|c|}{} &
  \multicolumn{1}{c|}{} &
  \multicolumn{1}{c|}{} &
  \multicolumn{3}{c|}{day} &
  \multicolumn{3}{c|}{night} &
  \multicolumn{3}{c|}{} \\ \cline{4-9}
\multicolumn{1}{|c|}{} &
  \multicolumn{1}{c|}{\multirow{-2}{*}{local   features,map pairs,query pairs,method}} &
  \multicolumn{1}{c|}{\multirow{-2}{*}{avg. all   bins}} &
  \multicolumn{1}{c|}{high} &
  \multicolumn{1}{c|}{mid} &
  \multicolumn{1}{c|}{low} &
  \multicolumn{1}{c|}{high} &
  \multicolumn{1}{c|}{mid} &
  \multicolumn{1}{c|}{low} &
  \multicolumn{3}{c|}{} \\ \cline{2-9}
\multicolumn{1}{|c|}{} &
  \multicolumn{1}{l|}{\textbf{R2D2,frustum,APGeM-20,COLMAP config2}} &
  \multicolumn{1}{c|}{\textbf{90.63}} &
  \multicolumn{1}{c|}{89.8} &
  \multicolumn{1}{c|}{96.5} &
  \multicolumn{1}{c|}{99.4} &
  \multicolumn{1}{c|}{73.3} &
  \multicolumn{1}{c|}{86.9} &
  \multicolumn{1}{c|}{97.9} &
  \multicolumn{3}{c|}{} \\ \cline{2-9}
\multicolumn{1}{|c|}{} &
  \multicolumn{1}{l|}{\textbf{R2D2,frustum,APGeM-20,pycolmap}} &
  \multicolumn{1}{c|}{\textbf{90.63}} &
  \multicolumn{1}{c|}{89.9} &
  \multicolumn{1}{c|}{96.4} &
  \multicolumn{1}{c|}{99.4} &
  \multicolumn{1}{c|}{73.8} &
  \multicolumn{1}{c|}{86.4} &
  \multicolumn{1}{c|}{97.9} &
  \multicolumn{3}{c|}{} \\ \cline{2-9}
\multicolumn{1}{|c|}{} &
  \multicolumn{1}{l|}{R2D2,frustum,APGeM-20,ransaclib} &
  \multicolumn{1}{c|}{90.62} &
  \multicolumn{1}{c|}{90.1} &
  \multicolumn{1}{c|}{96.5} &
  \multicolumn{1}{c|}{99.5} &
  \multicolumn{1}{c|}{73.3} &
  \multicolumn{1}{c|}{85.9} &
  \multicolumn{1}{c|}{98.4} &
  \multicolumn{3}{c|}{} \\ \cline{2-9}
\multicolumn{1}{|c|}{} &
  \multicolumn{1}{l|}{R2D2,frustum,APGeM-10,COLMAP config2} &
  \multicolumn{1}{c|}{89.27} &
  \multicolumn{1}{c|}{88.8} &
  \multicolumn{1}{c|}{96.1} &
  \multicolumn{1}{c|}{99.4} &
  \multicolumn{1}{c|}{69.1} &
  \multicolumn{1}{c|}{85.9} &
  \multicolumn{1}{c|}{96.3} &
  \multicolumn{3}{c|}{} \\ \cline{2-9}
\multicolumn{1}{|c|}{} &
  \multicolumn{1}{l|}{R2D2,frustum,APGeM-10,pycolmap} &
  \multicolumn{1}{c|}{89.52} &
  \multicolumn{1}{c|}{87.3} &
  \multicolumn{1}{c|}{95.3} &
  \multicolumn{1}{c|}{99.5} &
  \multicolumn{1}{c|}{71.2} &
  \multicolumn{1}{c|}{86.9} &
  \multicolumn{1}{c|}{96.9} &
  \multicolumn{3}{c|}{} \\ \cline{2-9}
\multicolumn{1}{|c|}{} &
  \multicolumn{1}{l|}{R2D2,frustum,APGeM-10,ransaclib} &
  \multicolumn{1}{c|}{89.67} &
  \multicolumn{1}{c|}{88.4} &
  \multicolumn{1}{c|}{96.2} &
  \multicolumn{1}{c|}{99.5} &
  \multicolumn{1}{c|}{71.2} &
  \multicolumn{1}{c|}{85.3} &
  \multicolumn{1}{c|}{97.4} &
  \multicolumn{3}{c|}{} \\ \cline{2-9}
\multicolumn{1}{|c|}{} &
  \multicolumn{1}{l|}{R2D2,frustum,APGeM-5,COLMAP config2} &
  \multicolumn{1}{c|}{88.17} &
  \multicolumn{1}{c|}{87.4} &
  \multicolumn{1}{c|}{95.4} &
  \multicolumn{1}{c|}{99.0} &
  \multicolumn{1}{c|}{65.5} &
  \multicolumn{1}{c|}{84.8} &
  \multicolumn{1}{c|}{96.9} &
  \multicolumn{3}{c|}{} \\ \cline{2-9}
\multicolumn{1}{|c|}{} &
  \multicolumn{1}{l|}{R2D2,frustum,APGeM-5,pycolmap} &
  \multicolumn{1}{c|}{88.12} &
  \multicolumn{1}{c|}{87.3} &
  \multicolumn{1}{c|}{95.4} &
  \multicolumn{1}{c|}{98.9} &
  \multicolumn{1}{c|}{65.5} &
  \multicolumn{1}{c|}{85.3} &
  \multicolumn{1}{c|}{96.3} &
  \multicolumn{3}{c|}{} \\ \cline{2-9}
\multicolumn{1}{|c|}{} &
  \multicolumn{1}{l|}{R2D2,frustum,APGeM-5,ransaclib} &
  \multicolumn{1}{c|}{88.08} &
  \multicolumn{1}{c|}{86.2} &
  \multicolumn{1}{c|}{94.3} &
  \multicolumn{1}{c|}{99.2} &
  \multicolumn{1}{c|}{65.5} &
  \multicolumn{1}{c|}{85.9} &
  \multicolumn{1}{c|}{97.4} &
  \multicolumn{3}{c|}{} \\ \cline{2-9}
\multicolumn{1}{|c|}{} &
  \multicolumn{1}{l|}{R2D2,frustum,APGeM-1,COLMAP config2} &
  \multicolumn{1}{c|}{84.67} &
  \multicolumn{1}{c|}{81.8} &
  \multicolumn{1}{c|}{92.1} &
  \multicolumn{1}{c|}{97.2} &
  \multicolumn{1}{c|}{62.3} &
  \multicolumn{1}{c|}{75.4} &
  \multicolumn{1}{c|}{99.2} &
  \multicolumn{3}{c|}{} \\ \cline{2-9}
\multicolumn{1}{|c|}{} &
  \multicolumn{1}{l|}{R2D2,frustum,APGeM-1,pycolmap} &
  \multicolumn{1}{c|}{82.92} &
  \multicolumn{1}{c|}{81.9} &
  \multicolumn{1}{c|}{91.4} &
  \multicolumn{1}{c|}{97.0} &
  \multicolumn{1}{c|}{60.2} &
  \multicolumn{1}{c|}{75.9} &
  \multicolumn{1}{c|}{91.1} &
  \multicolumn{3}{c|}{} \\ \cline{2-9}
\multicolumn{1}{|c|}{\multirow{-14}{*}{\begin{tabular}[c]{@{}c@{}}Aachen \\      Day-Night v1.1\end{tabular}}} &
  \multicolumn{1}{l|}{R2D2,frustum,APGeM-1,ransaclib} &
  \multicolumn{1}{c|}{83.30} &
  \multicolumn{1}{c|}{81.9} &
  \multicolumn{1}{c|}{90.8} &
  \multicolumn{1}{c|}{97.2} &
  \multicolumn{1}{c|}{60.7} &
  \multicolumn{1}{c|}{77.0} &
  \multicolumn{1}{c|}{92.2} &
  \multicolumn{3}{c|}{\multirow{-14}{*}{\begin{tabular}[c]{@{}c@{}}Accuracy   thresholds\\      Aachen\\      high: 0.25m, 2°\\      mid: 0.5m, 5°\\      low: 5m, 10°\end{tabular}}} \\ \hline
\multicolumn{12}{l}{} \\ \hline
\multicolumn{1}{|c|}{} &
  \multicolumn{1}{c|}{} &
  \multicolumn{1}{c|}{} &
  \multicolumn{3}{c|}{test} &
  \multicolumn{3}{c|}{validation} &
  \multicolumn{3}{c|}{} \\ \cline{4-9}
\multicolumn{1}{|c|}{} &
  \multicolumn{1}{c|}{\multirow{-2}{*}{local   features,map pairs,query pairs,method}} &
  \multicolumn{1}{c|}{\multirow{-2}{*}{avg. all bins}} &
  \multicolumn{1}{c|}{high} &
  \multicolumn{1}{c|}{mid} &
  \multicolumn{1}{c|}{low} &
  \multicolumn{1}{c|}{high} &
  \multicolumn{1}{c|}{mid} &
  \multicolumn{1}{c|}{low} &
  \multicolumn{3}{c|}{} \\ \cline{2-9}
\multicolumn{1}{|c|}{} &
  \multicolumn{1}{l|}{R2D2,distance-50,APGeM-20,COLMAP config2} &
  \multicolumn{1}{c|}{53.35} &
  \multicolumn{1}{c|}{42.6} &
  \multicolumn{1}{c|}{60.4} &
  \multicolumn{1}{c|}{64.8} &
  \multicolumn{1}{c|}{35.8} &
  \multicolumn{1}{c|}{56.4} &
  \multicolumn{1}{c|}{60.1} &
  \multicolumn{3}{c|}{} \\ \cline{2-9}
\multicolumn{1}{|c|}{} &
  \multicolumn{1}{l|}{R2D2,distance-50,APGeM-20,pycolmap} &
  \multicolumn{1}{c|}{55.97} &
  \multicolumn{1}{c|}{44.3} &
  \multicolumn{1}{c|}{62.5} &
  \multicolumn{1}{c|}{66.9} &
  \multicolumn{1}{c|}{38.1} &
  \multicolumn{1}{c|}{59.6} &
  \multicolumn{1}{c|}{64.4} &
  \multicolumn{3}{c|}{} \\ \cline{2-9}
\multicolumn{1}{|c|}{} &
  \multicolumn{1}{l|}{\textbf{R2D2,distance-50,APGeM-20,ransaclib}} &
  \multicolumn{1}{c|}{\textbf{56.12}} &
  \multicolumn{1}{c|}{44.5} &
  \multicolumn{1}{c|}{62.4} &
  \multicolumn{1}{c|}{67.0} &
  \multicolumn{1}{c|}{38.7} &
  \multicolumn{1}{c|}{59.9} &
  \multicolumn{1}{c|}{64.2} &
  \multicolumn{3}{c|}{} \\ \cline{2-9}
\multicolumn{1}{|c|}{} &
  \multicolumn{1}{l|}{R2D2,distance-50,APGeM-10,COLMAP   config2} &
  \multicolumn{1}{c|}{51.45} &
  \multicolumn{1}{c|}{39.9} &
  \multicolumn{1}{c|}{57.3} &
  \multicolumn{1}{c|}{61.8} &
  \multicolumn{1}{c|}{34.6} &
  \multicolumn{1}{c|}{55.5} &
  \multicolumn{1}{c|}{59.6} &
  \multicolumn{3}{c|}{} \\ \cline{2-9}
\multicolumn{1}{|c|}{} &
  \multicolumn{1}{l|}{R2D2,distance-50,APGeM-10,pycolmap} &
  \multicolumn{1}{c|}{52.67} &
  \multicolumn{1}{c|}{41.0} &
  \multicolumn{1}{c|}{58.1} &
  \multicolumn{1}{c|}{62.5} &
  \multicolumn{1}{c|}{35.7} &
  \multicolumn{1}{c|}{56.7} &
  \multicolumn{1}{c|}{62.0} &
  \multicolumn{3}{c|}{} \\ \cline{2-9}
\multicolumn{1}{|c|}{} &
  \multicolumn{1}{l|}{R2D2,distance-50,APGeM-10,ransaclib} &
  \multicolumn{1}{c|}{52.92} &
  \multicolumn{1}{c|}{41.0} &
  \multicolumn{1}{c|}{58.3} &
  \multicolumn{1}{c|}{62.5} &
  \multicolumn{1}{c|}{36.2} &
  \multicolumn{1}{c|}{57.6} &
  \multicolumn{1}{c|}{61.9} &
  \multicolumn{3}{c|}{} \\ \cline{2-9}
\multicolumn{1}{|c|}{} &
  \multicolumn{1}{l|}{R2D2,distance-50,APGeM-5,COLMAP config2} &
  \multicolumn{1}{c|}{46.98} &
  \multicolumn{1}{c|}{37.0} &
  \multicolumn{1}{c|}{53.0} &
  \multicolumn{1}{c|}{57.0} &
  \multicolumn{1}{c|}{31.1} &
  \multicolumn{1}{c|}{49.5} &
  \multicolumn{1}{c|}{54.3} &
  \multicolumn{3}{c|}{} \\ \cline{2-9}
\multicolumn{1}{|c|}{} &
  \multicolumn{1}{l|}{R2D2,distance-50,APGeM-5,pycolmap} &
  \multicolumn{1}{c|}{47.63} &
  \multicolumn{1}{c|}{37.4} &
  \multicolumn{1}{c|}{53.1} &
  \multicolumn{1}{c|}{57.4} &
  \multicolumn{1}{c|}{31.9} &
  \multicolumn{1}{c|}{50.2} &
  \multicolumn{1}{c|}{55.8} &
  \multicolumn{3}{c|}{} \\ \cline{2-9}
\multicolumn{1}{|c|}{} &
  \multicolumn{1}{l|}{R2D2,distance-50,APGeM-5,ransaclib} &
  \multicolumn{1}{c|}{47.75} &
  \multicolumn{1}{c|}{37.4} &
  \multicolumn{1}{c|}{53.6} &
  \multicolumn{1}{c|}{57.7} &
  \multicolumn{1}{c|}{32.0} &
  \multicolumn{1}{c|}{50.3} &
  \multicolumn{1}{c|}{55.5} &
  \multicolumn{3}{c|}{} \\ \cline{2-9}
\multicolumn{1}{|c|}{} &
  \multicolumn{1}{l|}{R2D2,distance-50,APGeM-1,COLMAP config2} &
  \multicolumn{1}{c|}{34.35} &
  \multicolumn{1}{c|}{27.0} &
  \multicolumn{1}{c|}{39.0} &
  \multicolumn{1}{c|}{43.2} &
  \multicolumn{1}{c|}{21.9} &
  \multicolumn{1}{c|}{35.5} &
  \multicolumn{1}{c|}{39.5} &
  \multicolumn{3}{c|}{} \\ \cline{2-9}
\multicolumn{1}{|c|}{} &
  \multicolumn{1}{l|}{R2D2,distance-50,APGeM-1,pycolmap} &
  \multicolumn{1}{c|}{34.42} &
  \multicolumn{1}{c|}{27.4} &
  \multicolumn{1}{c|}{38.9} &
  \multicolumn{1}{c|}{42.8} &
  \multicolumn{1}{c|}{22.4} &
  \multicolumn{1}{c|}{35.5} &
  \multicolumn{1}{c|}{39.5} &
  \multicolumn{3}{c|}{} \\ \cline{2-9}
\multicolumn{1}{|c|}{\multirow{-14}{*}{Gangnam Station B2}} &
  \multicolumn{1}{l|}{R2D2,distance-50,APGeM-1,ransaclib} &
  \multicolumn{1}{c|}{34.88} &
  \multicolumn{1}{c|}{27.4} &
  \multicolumn{1}{c|}{39.6} &
  \multicolumn{1}{c|}{43.2} &
  \multicolumn{1}{c|}{22.5} &
  \multicolumn{1}{c|}{36.5} &
  \multicolumn{1}{c|}{40.1} &
  \multicolumn{3}{c|}{\multirow{-14}{*}{\begin{tabular}[c]{@{}c@{}}Accuracy   thresholds\\      Gangnam B2\\      high: 0.1m, 1°\\      mid: 0.25m, 2°\\      low: 1m, 5°\end{tabular}}} \\ \hline
\multicolumn{12}{l}{} \\ \hline
\multicolumn{1}{|c|}{} &
  \multicolumn{1}{c|}{} &
  \multicolumn{1}{c|}{} &
  \multicolumn{3}{c|}{all} &
  \multicolumn{6}{c|}{} \\ \cline{4-6}
\multicolumn{1}{|c|}{} &
  \multicolumn{1}{c|}{\multirow{-2}{*}{local   features,map pairs,query pairs,method}} &
  \multicolumn{1}{c|}{\multirow{-2}{*}{avg. all bins}} &
  \multicolumn{1}{c|}{high} &
  \multicolumn{1}{c|}{mid} &
  \multicolumn{1}{c|}{low} &
  \multicolumn{6}{c|}{} \\ \cline{2-6}
\multicolumn{1}{|c|}{} &
  \multicolumn{1}{l|}{R2D2,frustum,APGeM-20,COLMAP config2} &
  \multicolumn{1}{c|}{71.77} &
  \multicolumn{1}{c|}{58.6} &
  \multicolumn{1}{c|}{73.7} &
  \multicolumn{1}{c|}{83.0} &
  \multicolumn{6}{c|}{} \\ \cline{2-6}
\multicolumn{1}{|c|}{} &
  \multicolumn{1}{l|}{R2D2,frustum,APGeM-20,pycolmap} &
  \multicolumn{1}{c|}{72.97} &
  \multicolumn{1}{c|}{60.0} &
  \multicolumn{1}{c|}{75.0} &
  \multicolumn{1}{c|}{83.9} &
  \multicolumn{6}{c|}{} \\ \cline{2-6}
\multicolumn{1}{|c|}{} &
  \multicolumn{1}{l|}{\textbf{R2D2,frustum,APGeM-20,ransaclib}} &
  \multicolumn{1}{c|}{\textbf{73.13}} &
  \multicolumn{1}{c|}{59.9} &
  \multicolumn{1}{c|}{75.6} &
  \multicolumn{1}{c|}{83.9} &
  \multicolumn{6}{c|}{} \\ \cline{2-6}
\multicolumn{1}{|c|}{} &
  \multicolumn{1}{l|}{R2D2,frustum,APGeM-10,COLMAP   config2} &
  \multicolumn{1}{c|}{69.30} &
  \multicolumn{1}{c|}{56.3} &
  \multicolumn{1}{c|}{71.4} &
  \multicolumn{1}{c|}{80.2} &
  \multicolumn{6}{c|}{} \\ \cline{2-6}
\multicolumn{1}{|c|}{} &
  \multicolumn{1}{l|}{R2D2,frustum,APGeM-10,pycolmap} &
  \multicolumn{1}{c|}{70.03} &
  \multicolumn{1}{c|}{57.6} &
  \multicolumn{1}{c|}{71.9} &
  \multicolumn{1}{c|}{80.6} &
  \multicolumn{6}{c|}{} \\ \cline{2-6}
\multicolumn{1}{|c|}{} &
  \multicolumn{1}{l|}{R2D2,frustum,APGeM-10,ransaclib} &
  \multicolumn{1}{c|}{70.17} &
  \multicolumn{1}{c|}{57.7} &
  \multicolumn{1}{c|}{72.2} &
  \multicolumn{1}{c|}{80.6} &
  \multicolumn{6}{c|}{} \\ \cline{2-6}
\multicolumn{1}{|c|}{} &
  \multicolumn{1}{l|}{R2D2,frustum,APGeM-5,COLMAP config2} &
  \multicolumn{1}{c|}{66.83} &
  \multicolumn{1}{c|}{55.1} &
  \multicolumn{1}{c|}{68.4} &
  \multicolumn{1}{c|}{77.0} &
  \multicolumn{6}{c|}{} \\ \cline{2-6}
\multicolumn{1}{|c|}{} &
  \multicolumn{1}{l|}{R2D2,frustum,APGeM-5,pycolmap} &
  \multicolumn{1}{c|}{67.07} &
  \multicolumn{1}{c|}{55.3} &
  \multicolumn{1}{c|}{68.5} &
  \multicolumn{1}{c|}{77.4} &
  \multicolumn{6}{c|}{} \\ \cline{2-6}
\multicolumn{1}{|c|}{} &
  \multicolumn{1}{l|}{R2D2,frustum,APGeM-5,ransaclib} &
  \multicolumn{1}{c|}{67.57} &
  \multicolumn{1}{c|}{56.0} &
  \multicolumn{1}{c|}{69.4} &
  \multicolumn{1}{c|}{77.3} &
  \multicolumn{6}{c|}{} \\ \cline{2-6}
\multicolumn{1}{|c|}{} &
  \multicolumn{1}{l|}{R2D2,frustum,APGeM-1,COLMAP config2} &
  \multicolumn{1}{c|}{55.83} &
  \multicolumn{1}{c|}{45.2} &
  \multicolumn{1}{c|}{57.2} &
  \multicolumn{1}{c|}{65.1} &
  \multicolumn{6}{c|}{} \\ \cline{2-6}
\multicolumn{1}{|c|}{} &
  \multicolumn{1}{l|}{R2D2,frustum,APGeM-1,pycolmap} &
  \multicolumn{1}{c|}{55.97} &
  \multicolumn{1}{c|}{45.3} &
  \multicolumn{1}{c|}{57.6} &
  \multicolumn{1}{c|}{65.0} &
  \multicolumn{6}{c|}{} \\ \cline{2-6}
\multicolumn{1}{|c|}{\multirow{-14}{*}{Baidu-mall}} &
  \multicolumn{1}{l|}{R2D2,frustum,APGeM-1,ransaclib} &
  \multicolumn{1}{c|}{56.17} &
  \multicolumn{1}{c|}{45.6} &
  \multicolumn{1}{c|}{57.8} &
  \multicolumn{1}{c|}{65.1} &
  \multicolumn{6}{c|}{\multirow{-14}{*}{\begin{tabular}[c]{@{}c@{}}Accuracy thresholds\\      Baidu-mall\\      high: 0.25m, 2°\\      mid: 0.5m, 5°\\      low: 5m, 10°\end{tabular}}} \\ \hline
\end{tabular}